\documentclass{elsarticle}

\usepackage{lineno,hyperref}
\usepackage[usenames,dvipsnames,svgnames,table]{xcolor}
\usepackage{color}
\usepackage{hyperref}
\hypersetup{
     colorlinks   = true,
     citecolor    = gray
}
\usepackage{amsmath} 
\usepackage{gensymb}
\usepackage{amssymb} 
\usepackage{multirow}
\usepackage{epstopdf}
\usepackage{gensymb}
\modulolinenumbers[5]
\journal{Robotics and Autonomous Systems}
\makeatletter
\def\ps@pprintTitle{%
 \let\@oddhead\@empty
 \let\@evenhead\@empty
 \def\@oddfoot{}%
 \let\@evenfoot\@oddfoot}
\makeatother

\begin{document}

\begin{frontmatter}

\title{Probabilistic RGB-D Odometry based on Points, Lines and Planes Under Depth Uncertainty}

\author{ Pedro F. Proen\c{c}a \corref{cor1}}
\cortext[cor1]{Corresponding author}
\ead{p.proenca@surrey.ac.uk}
\author{Yang Gao \corref{cor2}}
\address{Surrey Space Centre, Faculty of Engineering and Physical Sciences, University of Surrey, GU2 7XH Guildford, U.K}

\begin{abstract}
This work proposes a robust visual odometry method for structured environments that combines point features with line and plane segments, extracted through an RGB-D camera. Noisy depth maps are processed by a probabilistic depth fusion framework based on Mixtures of Gaussians to denoise and derive the depth uncertainty, which is then propagated throughout the visual odometry pipeline. Probabilistic 3D plane and line fitting solutions are used to model the uncertainties of the feature parameters and pose is estimated by combining the three types of primitives based on their uncertainties. \par
Performance evaluation on RGB-D sequences collected in this work and two public RGB-D datasets: TUM and ICL-NUIM show the benefit of using the proposed depth fusion framework and combining the three feature-types, particularly in scenes with low-textured surfaces, dynamic objects and missing depth measurements.
\end{abstract}

\end{frontmatter}

\section{Introduction}
Point, line and plane primitives allow a minimalistic, yet comprehensive representation of structured environments, which is more appealing than dense representations \cite{newcombe2011kinectfusion}, in terms of efficiency. While feature points can be insufficient for visual odometry in low textured environments, combining them with planes and line segment features may lead to more robustness to plain planar surfaces \cite{proenca2017planes,yang2017direct,PLSVO}, blur caused by sudden motion \cite{proenca2017planes} and light variations \cite{lu2015robustness}. Therefore, this work proposes a feature-based odometry method that combines points, lines and planes for visual odometry by relying on an RGB-D camera to capture densely the scene geometry and texture. \par

\begin{figure}[tb]
\centering
	\begin{tabular}{@{}c@{}c@{}c@{}}
         \includegraphics[scale=0.12]{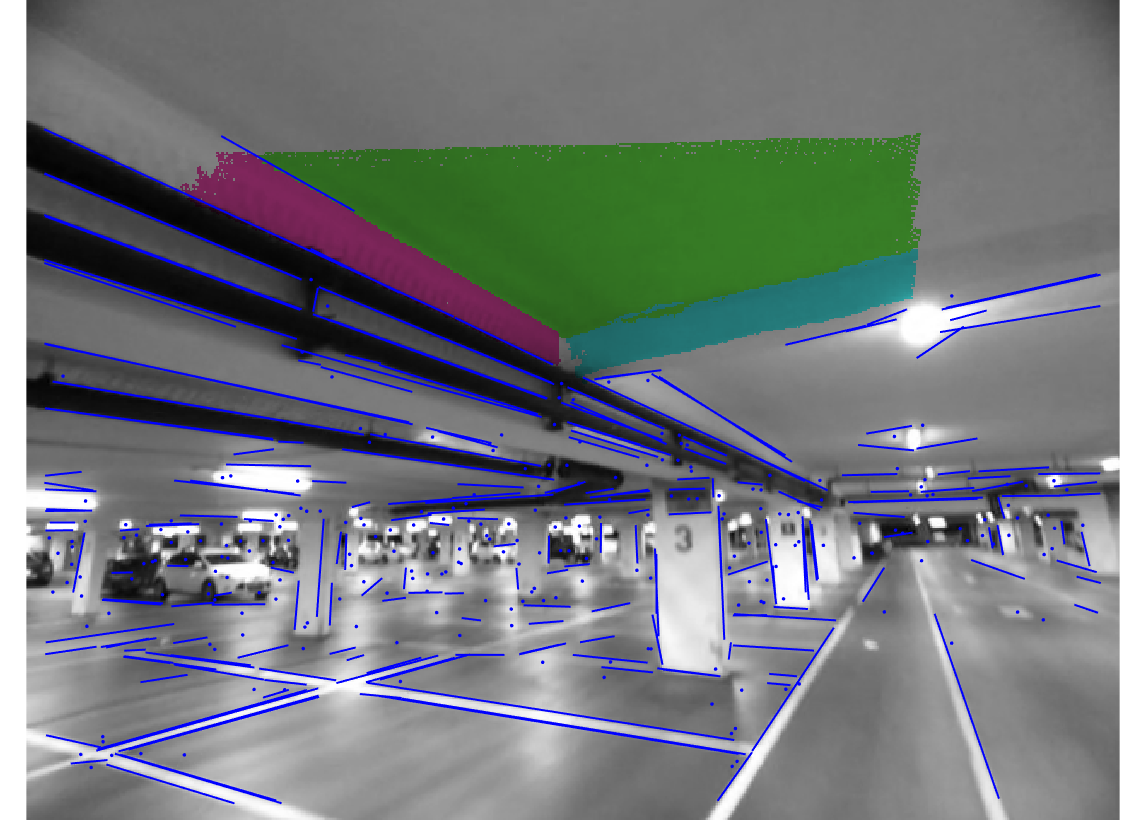} & 
	\includegraphics[scale=0.12]{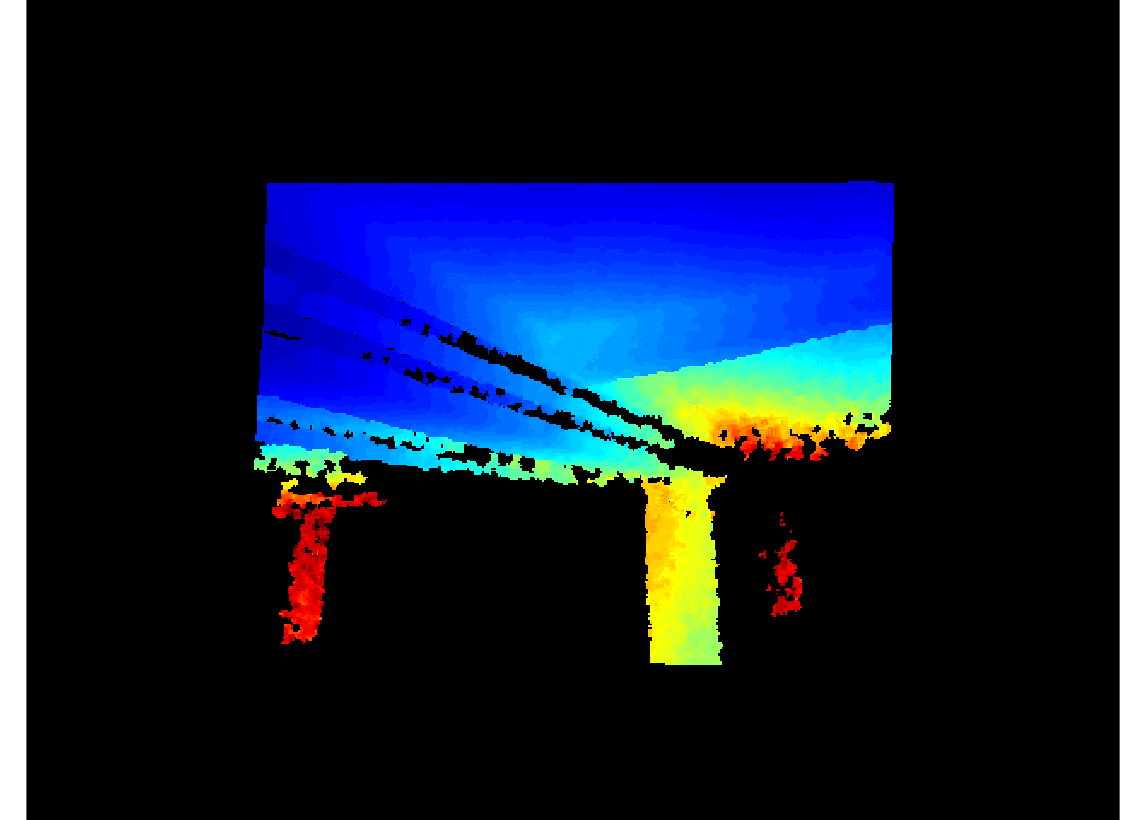} \\ [-7pt]
	{\scriptsize (a)} & {\scriptsize (b)} &\\
	 \includegraphics[scale=0.12]{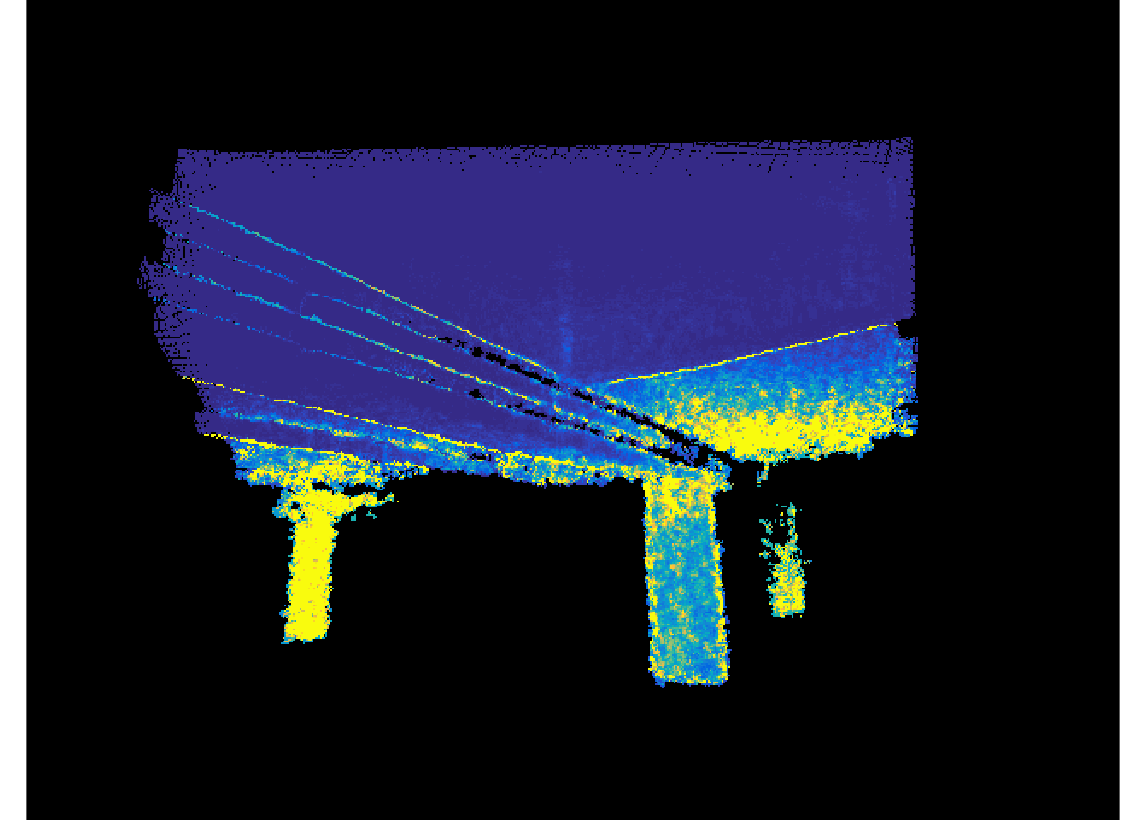} &
	\includegraphics[scale=0.12]{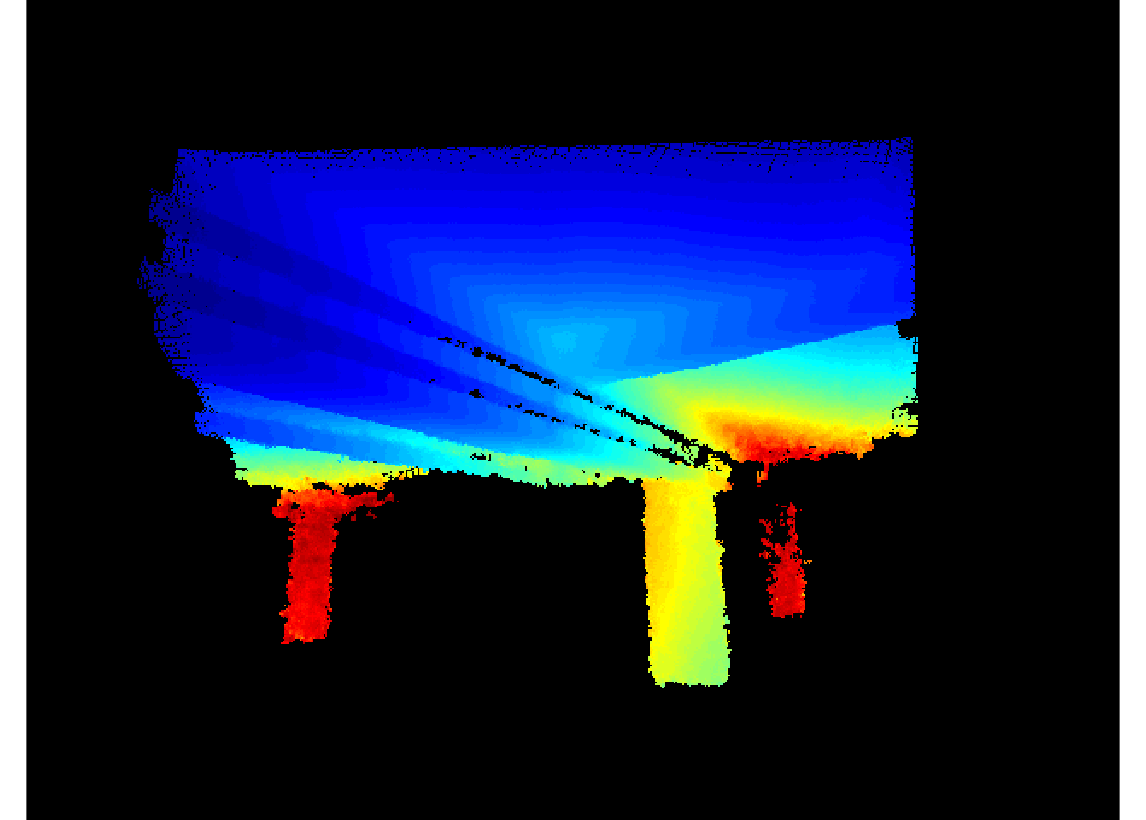}\\ [-7pt]
	{\scriptsize (c)} & {\scriptsize (d)} &\\[-5pt]
	\end{tabular}
	\caption{Raw and processed RGB-D frame by our system, in a challenging environment with low textured areas and missing depth measurements from a structured-light sensor. (a) Detected features overlaid on the intensity image. (b) Raw depth map. (c) Depth uncertainty estimated by the proposed depth filter and (d) the respective fused depth map. To expand the camera FOV, a wide-angle lens was mounted on the RGB camera while the depth fusion allows propagation of depth measurements beyond the narrow FOV of the depth camera.}
	\label{fig0}
\end{figure}

However, the depth measurements captured by active depth sensors is affected by significant error \cite{Khoshelham2012,Sarbolandi2015}, which in turn affects the estimation of 3D feature parameters. In particular, lines tend to be detected on depth discontinuities (see Fig. \ref{fig0}), where noise is more severe. Moreover, features may have missing depth measurements due to the range and field-of-view (FOV) limitations of these cameras. Thus, we propose a depth fusion framework to: (i) denoise the raw depth map, (ii) model the depth uncertainty and (iii) recover temporally missing depth measurements.
Since the effective depth error depends on the scene properties, besides considering the systematic depth sensor error, the proposed framework captures the observed uncertainty by assessing the spatial and temporal distribution of depth measurements. This uncertainty is then propagated throughout the visual odometry pipeline. Specifically, probabilistic 3D line and plane fitting solutions, based on weighted linear least squares, are used to model the uncertainty of these primitives and then pose is estimated by taking into account these uncertainties. The motivation for representing the uncertainties of these primitives, is that their impact on the pose estimation should depend on the precision of their estimated parameters, which depend on the number, uncertainties and distribution of their samples. The key contributions of this paper are the following:
\begin{itemize}
\item A probabilistic depth fusion framework based on Mixture of Gaussians that models depth uncertainty. The code is available as open-source\footnote{\url{https://github.com/pedropro/OMG_Depth_Fusion}}.
\item Extend our recently developed visual odometry method \cite{proenca2017planes} based on points and planes to line segments. 
\item A probabilistic analytical solution to 3D line fitting.
\item Evaluate the system on public and author-collected\footnote{A video of the experiments is available at: \url{https://youtu.be/Y0T2_ghlng0}} RGB-D datasets, and demonstrate the benefit of modelling temporally depth uncertainty and combining points, planes and lines in low textured and dynamic environments.
\end{itemize}
The remainder of this paper is organized as follows. Section \ref{sec:relatedwork} reviews related work focusing on methods that use depth cameras and features (e.g. points, planes and lines). Section \ref{sec:depth_filter} and Section \ref{sec:vis_odometry} describe respectively our depth filter framework and visual odometry method. Our results are reported and discussed in Section \ref{sec:experiments}. Finally, Section \ref{sec:conclusions} concludes and discusses the limitations of the approach.\par

\section{Related work}
\label{sec:relatedwork}
Active depth sensors have been widely adopted in many computer vision tasks, e.g., reconstruction, segmentation, egomotion estimation, object recognition, human pose estimation and scene understanding \cite{han2013enhanced}. However, extensive analysis \cite{Khoshelham2012,smisek20133d,Sarbolandi2015,wasenmuller2016comparison,lateralNoiseNguyen} of these consumer-grade sensors have revealed their limitations. It is now well known that structured-light sensors, used by the first version of Kinect cameras, suffer from severe quantization and consequently the depth error grows quadratically with the distance to the sensor. A theoretical model for this source of error has been derived in \cite{Khoshelham2012}. Later, time-of-flight (ToF) sensors, used by the second version of Kinect cameras, have been empirically analyzed and compared to the first version in \cite{Sarbolandi2015,wasenmuller2016comparison}. Although, the ToF depth error proved to be significantly less affected by the distance to the sensor, ToF sensors suffer from other sources of error: flying pixels arising from depth descontinuities, non-Lambertian surfaces (e.g. black surfaces), multipath interference and the depth error grows with the image distance to the principal point. Besides these sensor-specific issues, depth sensors, in general, have limited range and FOV compared to LIDAR sensors, as shown in Fig. \ref{fig0}. \par
KinectFusion \cite{newcombe2011kinectfusion} was the first work to reconstruct dense models from the noisy and incomplete depth maps captured by these sensors. This system uses raw depth maps to update a global volumetric model based on cumulative moving average updates of voxel states, which are represented as Truncated Signed Distance Functions (TSDF), while pose is estimated by using Iterative Closest Point (ICP) algorithm. To perform ICP, the model is raycasted and the depth maps are first denoised by a bilateral filter \cite{tomasi1998bf}. Since then, several works \cite{lateralNoiseNguyen,selectivedepthfusion,WasenmuellerInspection} have extended KinectFusion to achieve better quality reconstructions. In \cite{lateralNoiseNguyen}, a depth noise model that takes into account both the sensor lateral and axial noise, was empirically derived and incorporated into the KinectFusion pipeline. Specifically, the depth uncertainty was used to weight the ICP and the voxel TSDF updates. Due to the GPU memory requirements and voxel discretization of these volumetric methods, a selective point-based fusion method was instead proposed in \cite{selectivedepthfusion} to reconstruct denoised 3D models in dynamic environments. More recently, temporal depth map fusion has been used to denoise depth maps either by using the median \cite{WasenmuellerInspection} or the moving average \cite{kpaslam}. However, in these works, depth uncertainty is neither modelled nor explicitly used for fusion. \par
The dense RGB-D odometry method, termed DVO \cite{DVO2013}, which is based on the minimization of the photometric and geometric error, in \cite{DVO2}, has also been improved, in \cite{gutierrez2016dense,wasenmuller2016dna}, by considering the depth error. \cite{gutierrez2016dense} proposed using the inverse depth to parameterize the geometric error, whereas \cite{wasenmuller2016dna} proposed using the image derivatives to weight the residuals of the minimization problem. \par
In feature-based SLAM methods, to address the low depth resolution of structured light cameras, specific sensor depth uncertainty models \cite{Khoshelham2012,smisek20133d} were adopted for point and line odometry \cite{PLVO} and for our recently proposed point and plane odometry \cite{proenca2017planes}. Moreover, \cite{dryanovski2013fast} proposed using a Mixture of Gaussians convolution to assess the uncertainty of a single depth map. Such framework represents the uncertainty around the object edges better than the sensor error models, proposed in \cite{Khoshelham2012,lateralNoiseNguyen}, thus our proposed depth filter builds on this framework. The resulting uncertainty was further used, in \cite{dryanovski2013fast}, to update a sparse model of feature points, through Kalman filter correction equations. An experimental study comparing dense vs. feature point based VO and ICP variations was performed in \cite{fang2014experimental} and showed no clear winner since their relative performance depends on the particular environment characteristics. \par
Beside features points, line primitives are becoming increasingly popular in monocular \cite{yang2017direct,PLSVO,PumarolaICRA2017} and RGB-D Odometry \cite{lu2015robustness,PLVO,SPLODE2017pro}. A non-linear 3D line fitting was proposed in \cite{lu2015robustness} to fit depth measurement samples and their uncertainties, however it is not efficient to cast each line fitting as an iterative problem, considering the typical high number of detected 2D lines (e.g. 100). This problem was more recently simplified in \cite{yang2017direct}, where an analytical method was devised for monocular odometry by exploiting the fact that a 3D line is projected as a plane. While, this solution is more appealing than the previous one, pixel samples from a 2D line may not lie all on the same plane, e.g., the pixels crossed by an oblique line. Moreover, one may wish to sample depth measurements from the 2D line neighbourhood, due to a lack of measurements, or fit a 3D line to measurements obtained from multiple line observations. Therefore, the analytical 3D line fitting solution, proposed here, is more general, as it supports all these cases. Basic 3D line fitting based on PCA was employed in \cite{SPLODE2017pro} but this neglects the depth uncertainties. \par 
Plane primitives have also been widely exploited by SLAM methods: A planar method was proposed in \cite{trevor2012planar} to use data from both a 2D LIDAR and a depth camera as these complement each other in terms of FOV and operating range. Points and planes were initially combined by a SLAM system, in \cite{PointsAndPlanes_2013}, to avoid the geometric degeneracy of planes. The system used a RANSAC framework for mixed 3D registration of both point-to-point and plane-to-plane matches by sampling any triplets formed by these matches. To cope with missing depth measurements, this framework was later extended in \cite{ataer2016pinpoint} to include also 2D-to-3D point matches as a triplet hypothesis and 2D-to-2D matches for checking the hypothesis consensus. To reduce the computational cost of plane extraction, plane tracking was proposed in \cite{ataer2013tracking}, however faster plane extraction algorithms have been recently developed \cite{feng2014fast,holz2011real}. Alternatively, in \cite{li2016novel}, feature points have been enhanced with planar patches, for a small overhead, to improve feature matching and increase the constraints imposed by feature points on pose estimation, but nevertheless this approach does not exploit featureless planar patches. On the contrary, in \cite{ma2016cpa}, the Direct SLAM method \cite{DVO2} was combined with global plane model tracking through an EM framework to reduce drift. Later, a more efficient alternative was developed in \cite{kpaslam}, without using GPU. The uncertainty in plane extraction was analyzed thoroughly in \cite{pathak2010uncertainty} by comparing direct and iterative plane fitting methods, in terms of accuracy and speed. In \cite{proenca2017planes}, we proposed an RGB-D Odometry method for points and planes that modelled and propagated the depth uncertainty throughout the system pipeline. Here, we extend this approach to lines and propose a better depth model.

\section{Probabilistic Depth Filter}
\label{sec:depth_filter}

\begin{figure}[tb]
\centering
	\includegraphics[scale=0.5]{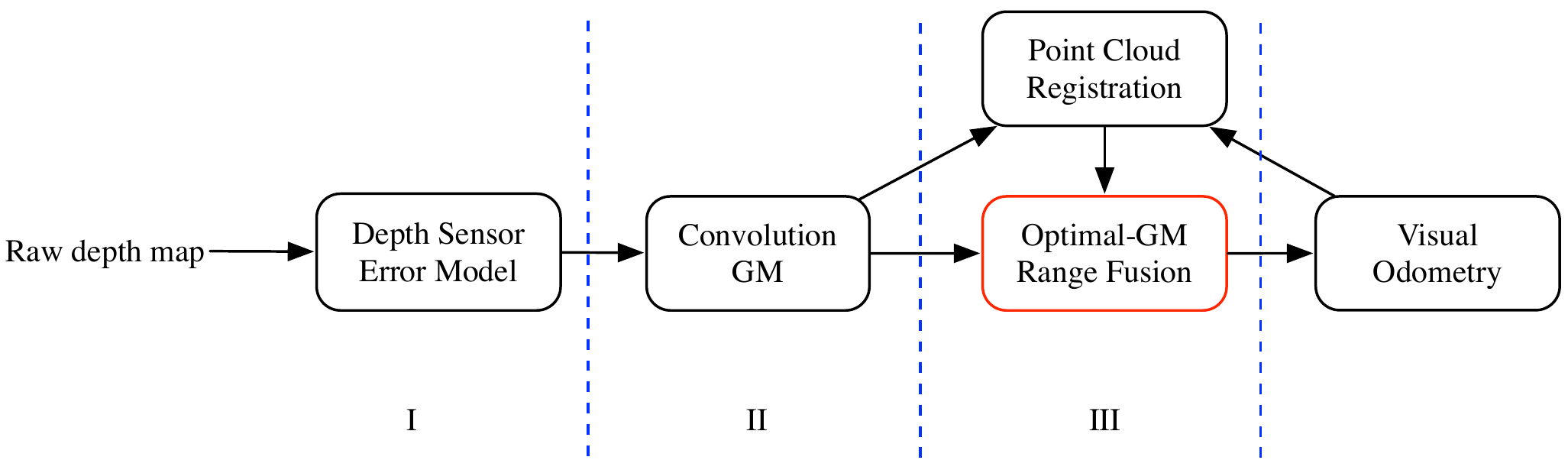}
	\vspace{-1em}
	\caption{Overview of the proposed probabilistic depth filter}
	\label{fig1}
\end{figure}

The proposed depth filter, outlined in Fig. \ref{fig1}, can be split into three stages: (i) Given the raw depth map of the current frame, depth uncertainty is assessed according to a specific sensor model, (ii) Based on this depth uncertainty, the raw depth map is convolved with the Gaussian Mixture (GM) kernel proposed in \cite{dryanovski2013fast}, to capture the uncertainty within the pixels neighbourhood, (iii) The depth estimates and uncertainties resulting from this GM convolution are then combined with estimates from past frames by using our proposed Optimal-GM fusion method. In order to do so, depth estimates from a sliding window of frames are maintained and updated as 3D measurements by a point cloud registration module. A detailed explanation of these modules is given in the following subsections.

\subsection{Depth Sensor Error Model}

In our experimental work, we have used structured-light depth sensors based on active stereo, which suffer inherently from disparity quantization, therefore, we adopted the theoretical error model of \cite{Khoshelham2012}, which accounts for the propagation of disparity random error $\sigma_d$ in Kinect structured-light sensors, so that the depth uncertainty is given by: $\sigma_z = \sigma_d(\frac{m}{fb})Z^2$ where $f$ is the focal length, $b$ is the baseline between the IR projector and camera and $m$ is a normalizing parameter. Setting $\sigma_d=0.5$, as in \cite{Khoshelham2012}, yields the following expression:
\begin{equation}
\label{eq:depth_sensor error_model}
\sigma_z = 1.425\times 10^{-6}z^2  \quad [mm]
\end{equation}
which fits well the planar residuals in \cite{Khoshelham2012} and is consistent with the axial noise in \cite{lateralNoiseNguyen}. However, this simple expression does not comprehend many other sources of depth error, e.g., lateral noise \cite{lateralNoiseNguyen}, ambient background light and temperature drift \cite{Sarbolandi2015}. Although, a more comprehensive model could be developed, it is extremely difficult to model the actual depth error, since this depends also on the properties of the observed object surfaces. Thus, we refrain from doing so and instead look at the spatial and temporal distribution of depth samples, through the next consecutive modules.

\subsection{Convolution of Gaussian Mixtures}
\label{sec:GMDC}
To address the lateral error, Dryanovski et al. \cite{dryanovski2013fast} proposed to quantify the uncertainty of depth pixels based on the depth values of their image neighbourhoods through a GM formulation. Let the probability density function of depth in a $3\times3$ local window, centered at pixel $p$,  be given by the following $N$ mixture of Gaussians:
\begin{equation}
\label{eq:GMD_eq}
f(z) = \frac{1}{S} \sum_{i=1}^{N}w_i\mathcal{N}(z_i, \sigma_{z_i}^2)
\end{equation}
where each Gaussian corresponds to a pixel of the local window with a variance given by the depth sensor error model, $S$ is a normalizing constant and the weights $w_i$ are assigned to the local window according to the following kernel:
\begin{equation}
\label{eq:kernel}
W = \begin{bmatrix} 
1&  2  & 1\\[-5pt]
2&   4  &  2\\[-5pt]
1 &  2  & 1
\end{bmatrix} 
\end{equation}
Then, the new estimated depth for $p$ takes the value of the mean of (2):
\begin{equation}
\label{eq:Z_GMD}
\bar{z}= \frac{1}{S} \sum_{i=1}^{N}w_i z_i
\end{equation}
and the GM uncertainty can be found by expressing the variance in terms of moments:
\begin{equation}
\label{eq:GMD_Var}
\text{Var}(f(z)) = \frac{1}{S} \sum_{i=1}^{N}w_i(z^2_i + \sigma^2_{z_i}) - \bar{z}^2
\end{equation}

To account for pixels with missing depth values, we simply represent their depth and uncertainty as 0 and flag them with an indicator function $y$, such that the normalizing constant is given by:

\begin{equation}
\label{eq:normalizing constant}
S = \sum_{i=1}^{N}w_iy_i
\end{equation}

Effectively, the resulting variance allows assigning high uncertainty to outliers (e.g. flying pixels) and depth discontinuity locations. One motivation for the latter, is that the 2D coordinates of features detected on the RGB images are also subject to error and moreover RGB images may not be perfectly aligned with the depth map due to errors in the extrinsic calibration and temporal synchronization, consequently image features corresponding to foreground may be associated to background. This information should be taken into account during both the pose estimation and the temporal fusion to reduce the impact of wrong depth associations.

\begin{figure}[tb]
\centering
	\includegraphics[scale=0.5]{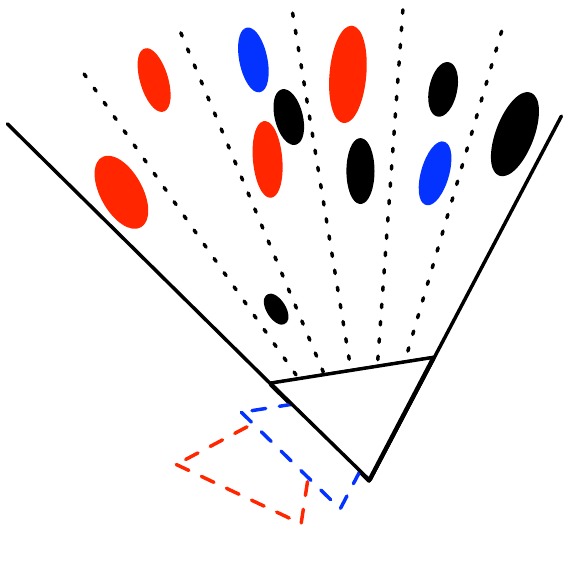}
	\caption{Illustration of the problem of fusing point measurements with their uncertainties from three frames}
	\label{fig2}
\end{figure}

\subsection{Optimal Gaussian Mixture for Temporal Fusion}

Given the 3D measurements of past frames, which are transformed to the current frame by the point cloud registration, these are projected to the image grid, as illustrated in Fig. \ref{fig2}, such that each pixel will have a set of points with their respective uncertainties, estimated by the GM convolutions. Due to the transformation between frames, ideally the $3\times3$ covariances of the points should be rotated as well, however this is computationally expensive, thus we work on the range space which is invariant to camera rotation unlike the depth. Specifically, the depth uncertainties given by the GM convolutions are propagated to range uncertainties and then stored, as follows:

\begin{equation}
\label{eq:depth2range} 
\sigma^2_r = \frac{\sigma^2_z}{ \cos^2\alpha}
\end{equation}
where $\alpha$ is the angle of incidence of the projection ray on the image plane and a matrix of cosines, for all pixels, can be pre-computed according to the camera intrinsic parameters. Once the points are converted to range as well, the previous GM framework can be applied to obtain a new range estimate for each pixel, although in this case we use the range uncertainties to weight the GM, such that the resulting range is:

\begin{equation}
\label{eq:Z_GMD}
\bar{r}= \frac{1}{\sum_{i=1}^{M}\sigma^{-2}_{r_i}} \sum_{i=1}^{M}\frac{r_i}{\sigma^2_{r_i}}
\end{equation}
for $M$ projected points. This expression corresponds in fact to the Maximum Likelihood Estimator (MLE) \cite{efron1978MLE}, but more notably because this framework does not make independence assumptions, the uncertainty given by the expression (\ref{eq:GMD_Var}) takes into account both the intra-group and inter-group variances, unlike the least squares formulation. Thus, pixels with inconsistent range measurements will have high uncertainty and the gross errors, e.g., flying pixels, that were \textit{a-priori} modelled by the GM convolutions will be penalized during the range fusion by the weighting function in (\ref{eq:Z_GMD}).\par
\begin{figure}[tb]
\centering
	\begin{tabular}{@{}c@{ }c@{ }c@{ }c@{ }c@{}}
	\multirow{5}{*}{\includegraphics[scale=0.16]{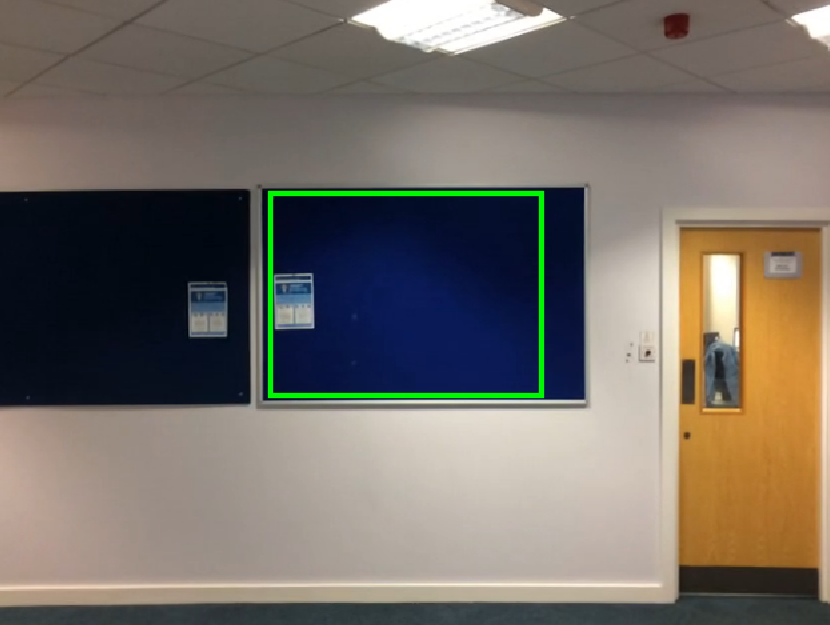}} &  &  &  &\multirow{5}{*}{\includegraphics[scale=0.16]{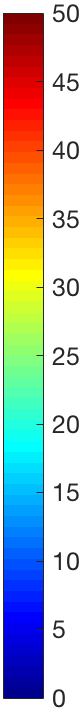}}\\[-18pt] &
	\includegraphics[scale=0.11]{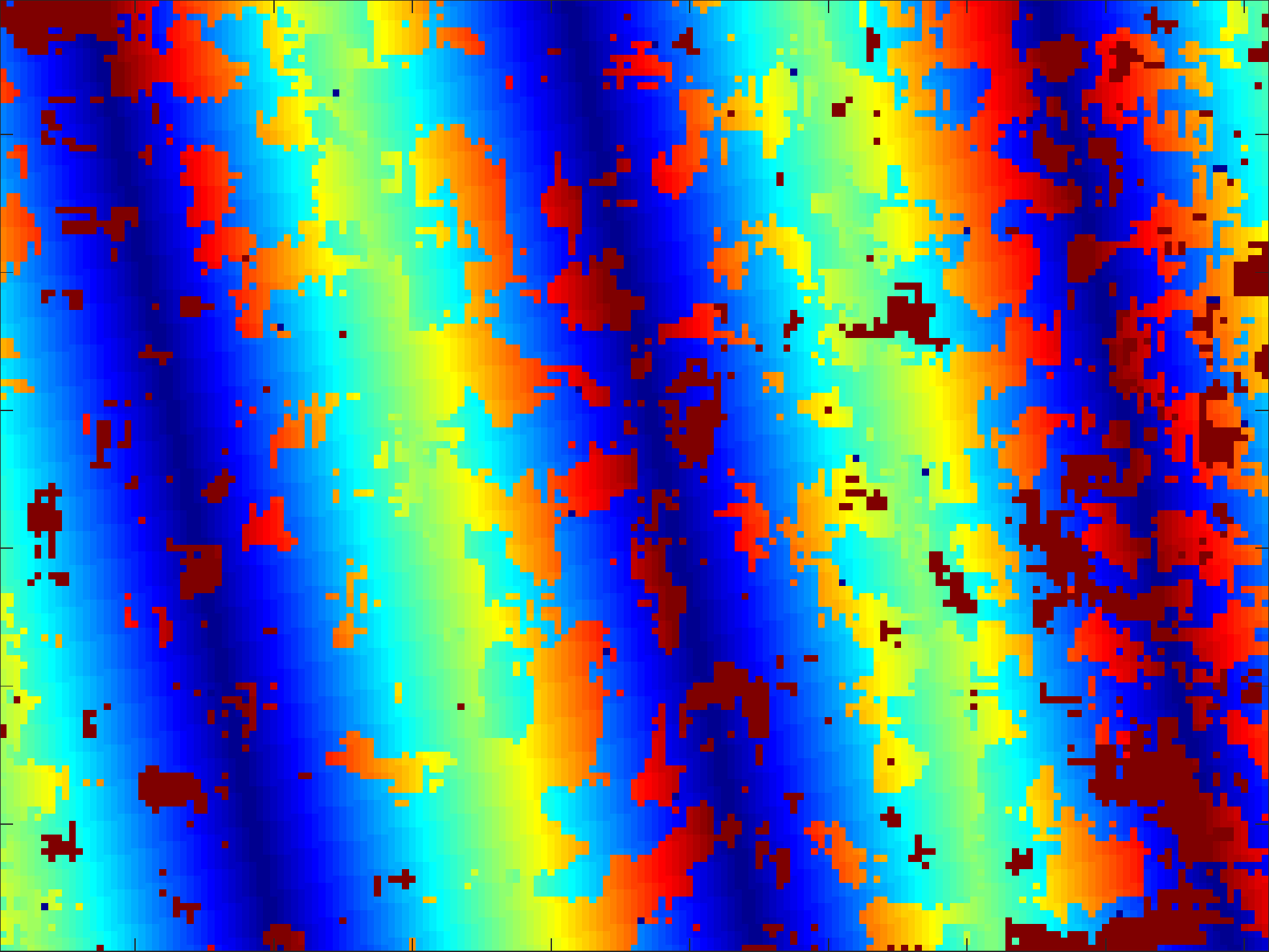} & 
	\includegraphics[scale=0.11]{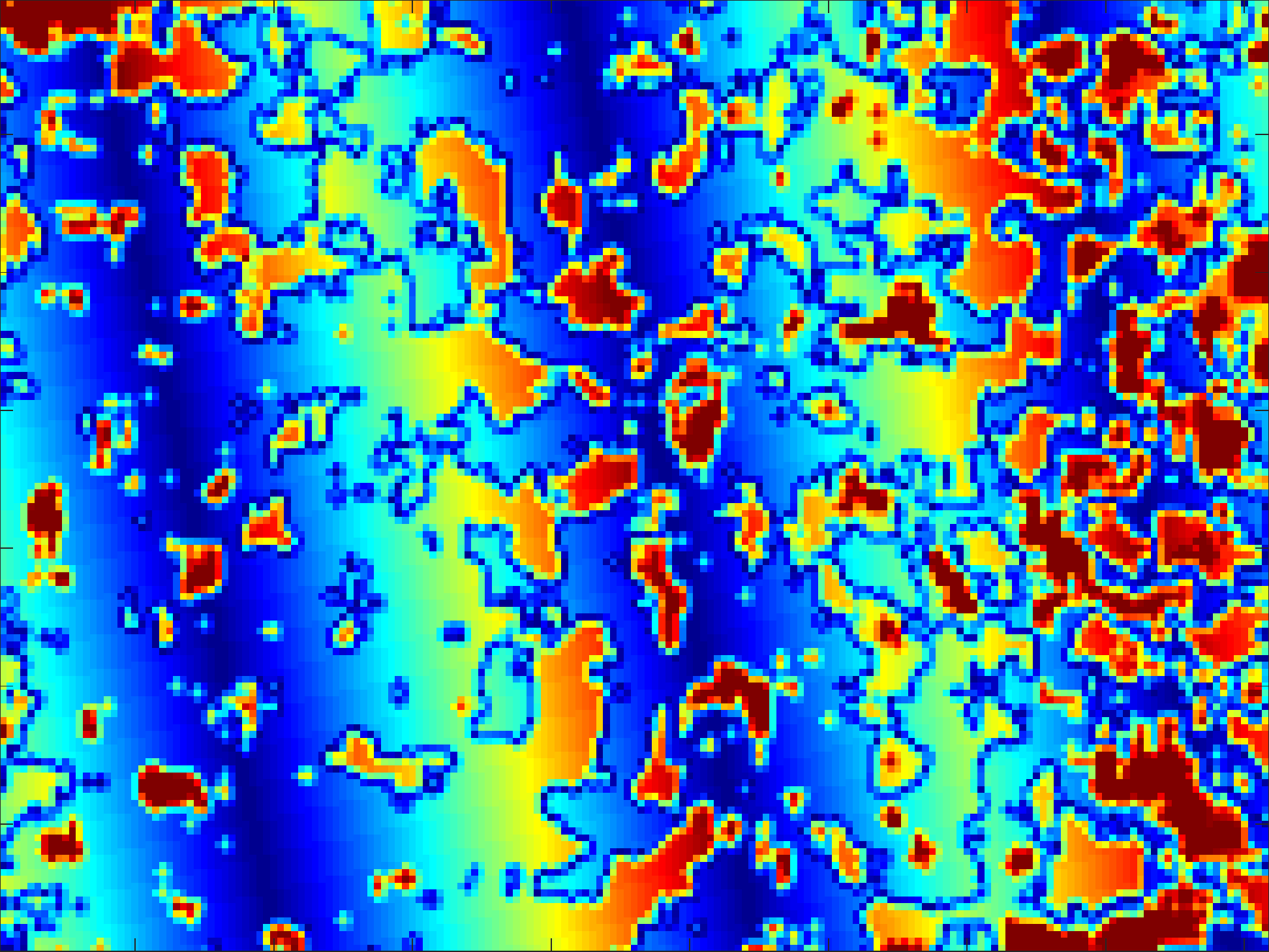} &  
	\includegraphics[scale=0.11]{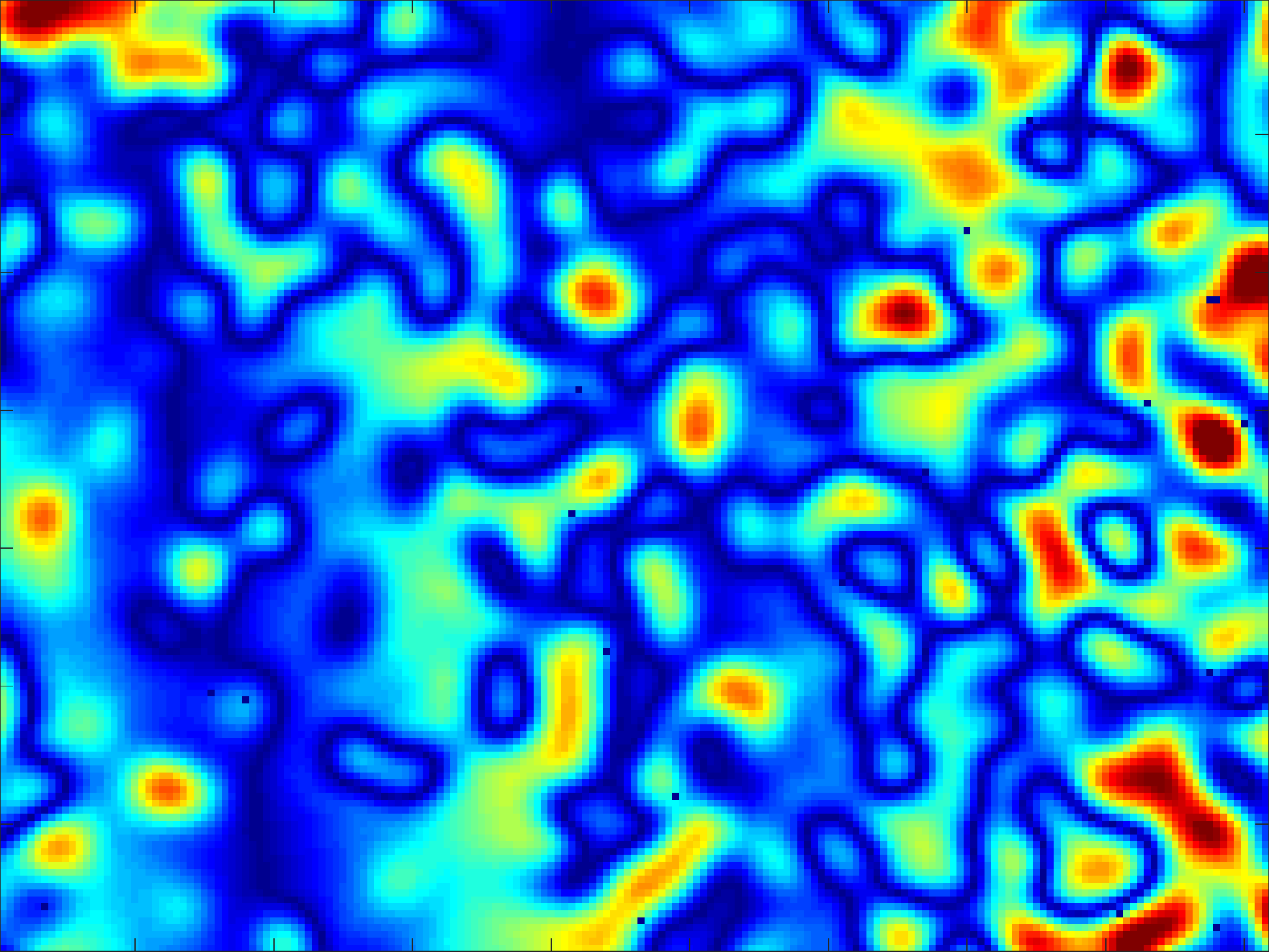} & \\[-7pt]
	& {\scriptsize Raw: 28 mm} & {\scriptsize C-GM: 24 mm}  & {\scriptsize BF: 17 mm } &\\[-2pt]
         &  \includegraphics[scale=0.11]{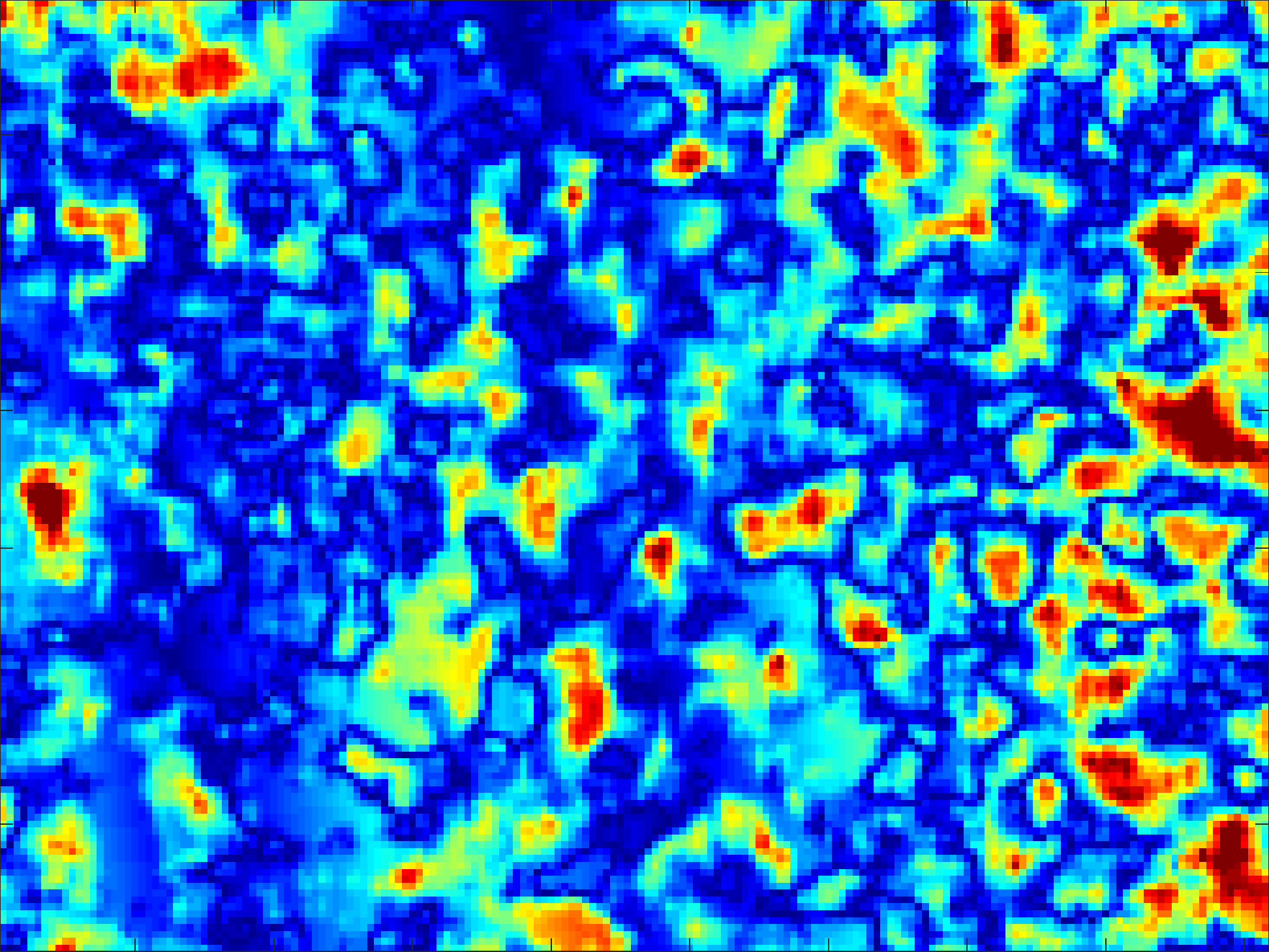} & 
         \includegraphics[scale=0.11]{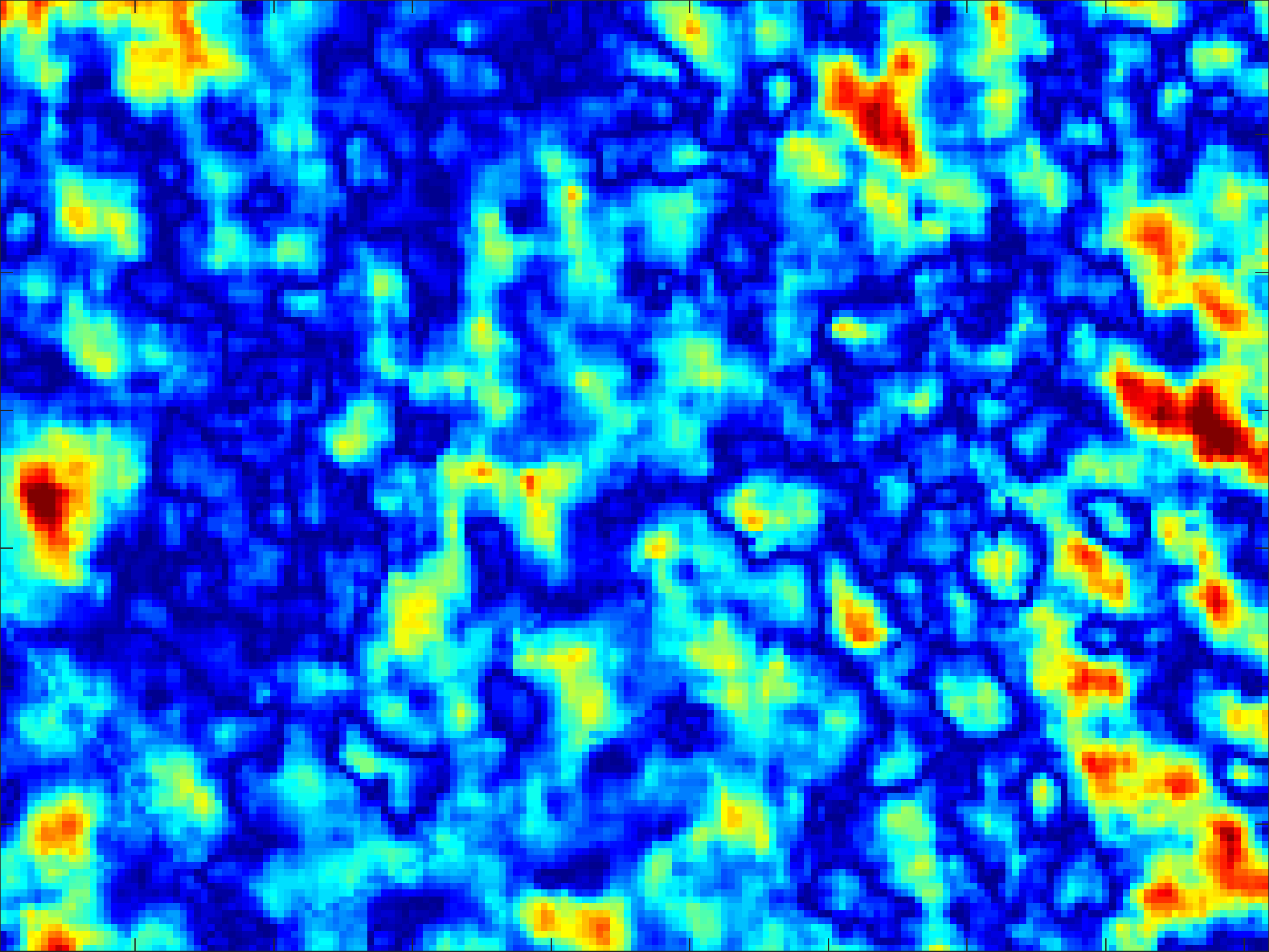} & 
         \includegraphics[scale=0.11]{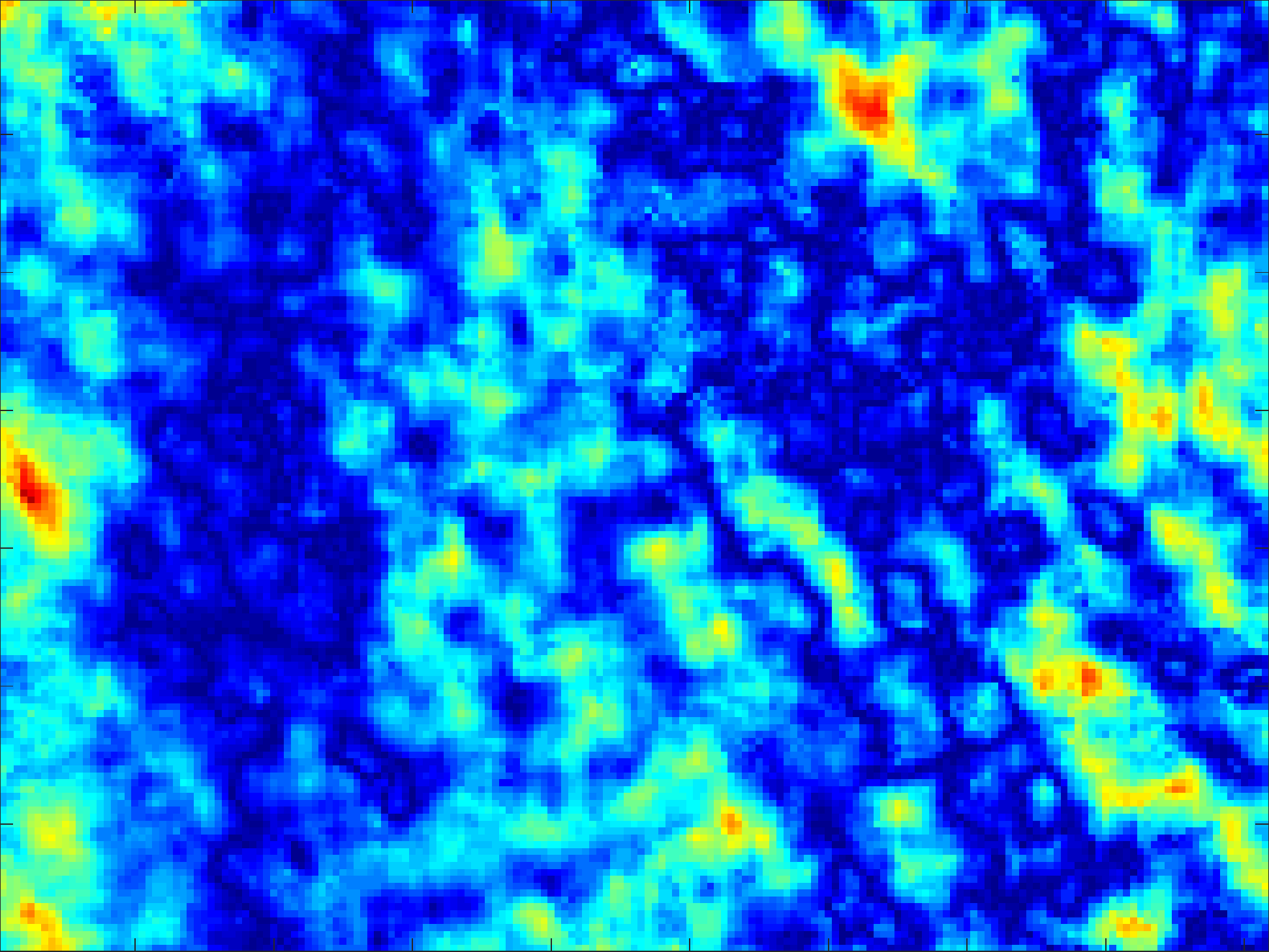} &\\[-7pt]
         &{\scriptsize O-GM$_5$: 18 mm} & {\scriptsize O-GM$_{10}$: 15 mm} & {\scriptsize O-GM$_{20}$: 13 mm}
	\end{tabular}
	\caption{Planar fitting residuals for three depth filtering methods: C-GM is the method described in Section \ref{sec:GMDC}; BF is the widely used bilateral filter \cite{tomasi1998bf} (e.g. used in KinectFusion \cite{newcombe2011kinectfusion}) and O-GM$_n$ is our temporal fusion method with measurements from $n$ frames. Maps of point-to-plane distances and the respective RMSE are depicted for the region highlighted on the left image. O-GM$_n$ and BF are able to remove two types of error, which are revealed in the raw depth map: random errors and quantization errors (appearing as stripes).}
	\label{fig3}
\end{figure}
\begin{figure}
	\centering
	\begin{tabular}{@{}c@{ }c@{ }c@{}}
		\includegraphics[scale=0.256]{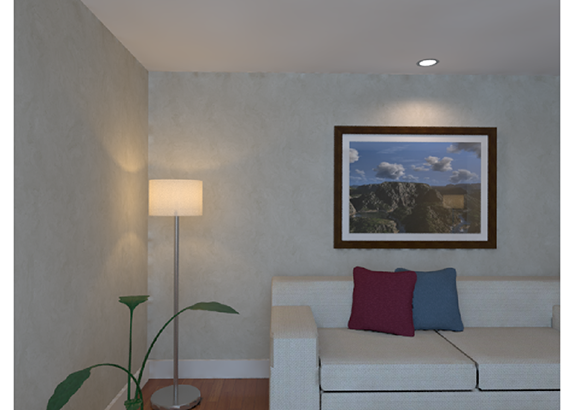} & 
		\includegraphics[scale=0.25]{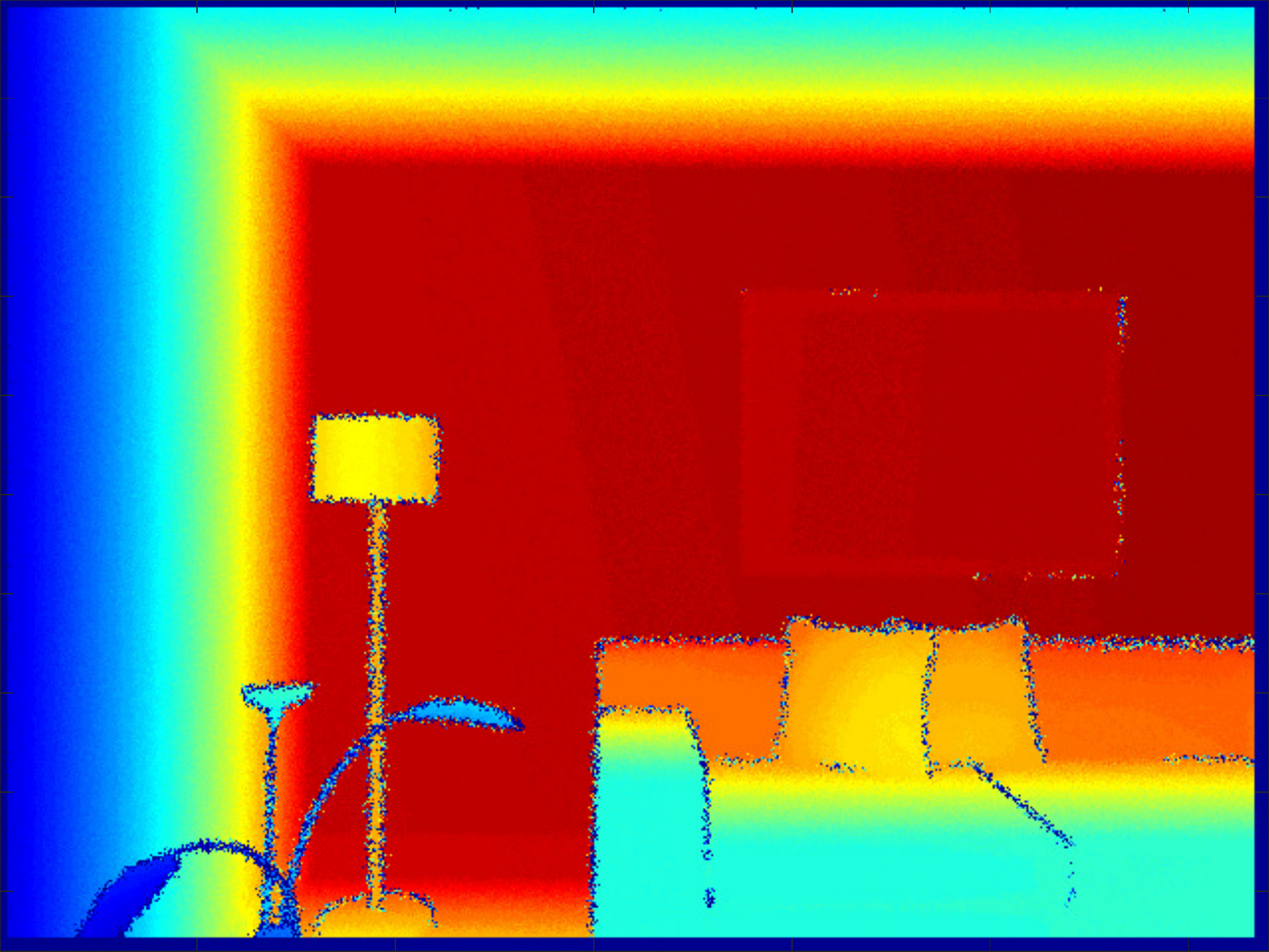} &
		\includegraphics[scale=0.155]{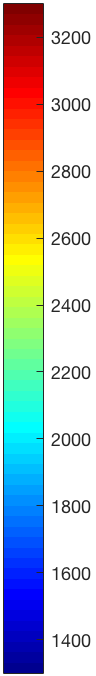} \\[-7pt]
		{\scriptsize RGB with simulated noise} & {\scriptsize Raw depth map with simulated noise}\\[-5pt]
		\includegraphics[scale=0.25]{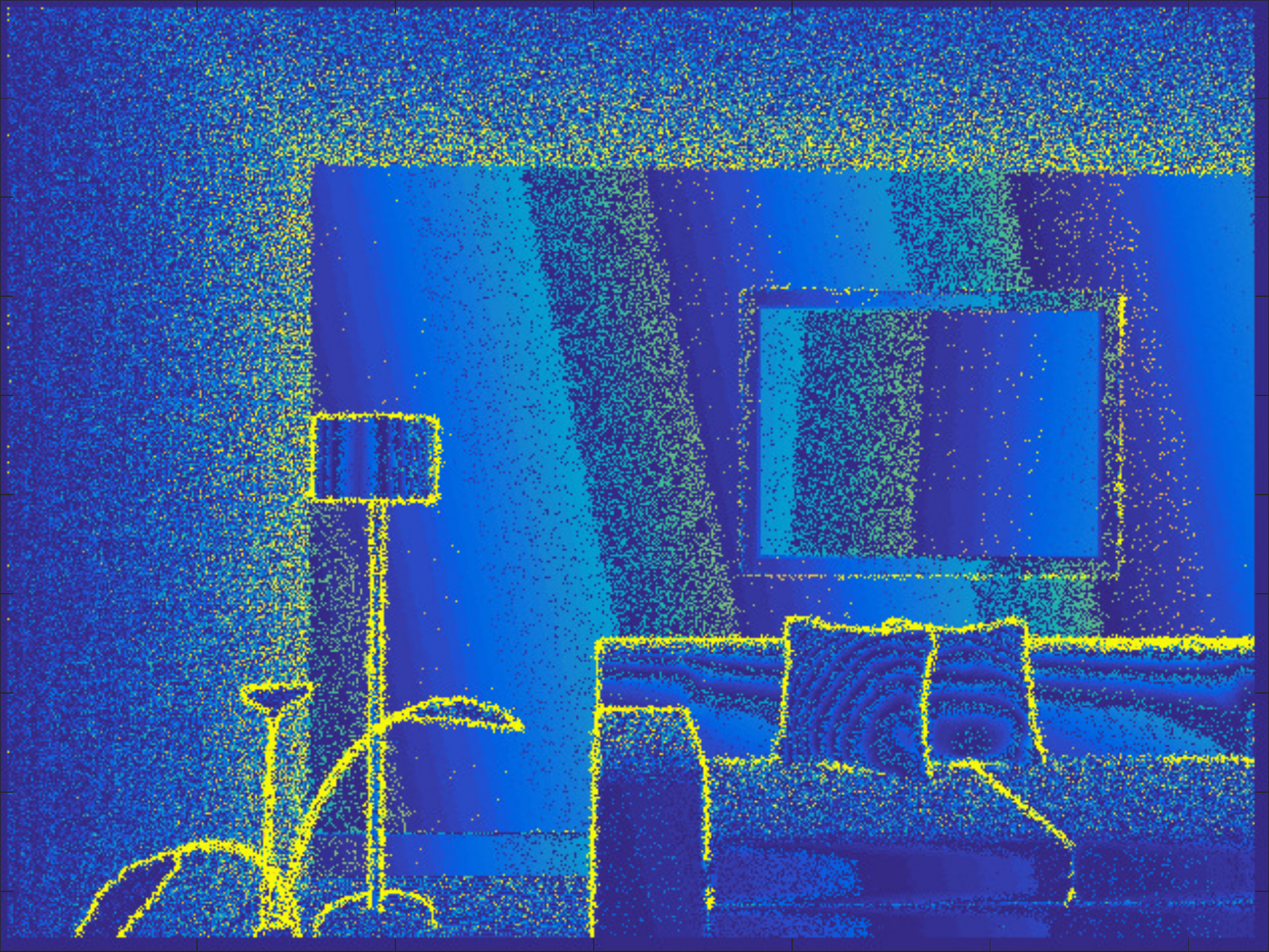} &
		\includegraphics[scale=0.25]{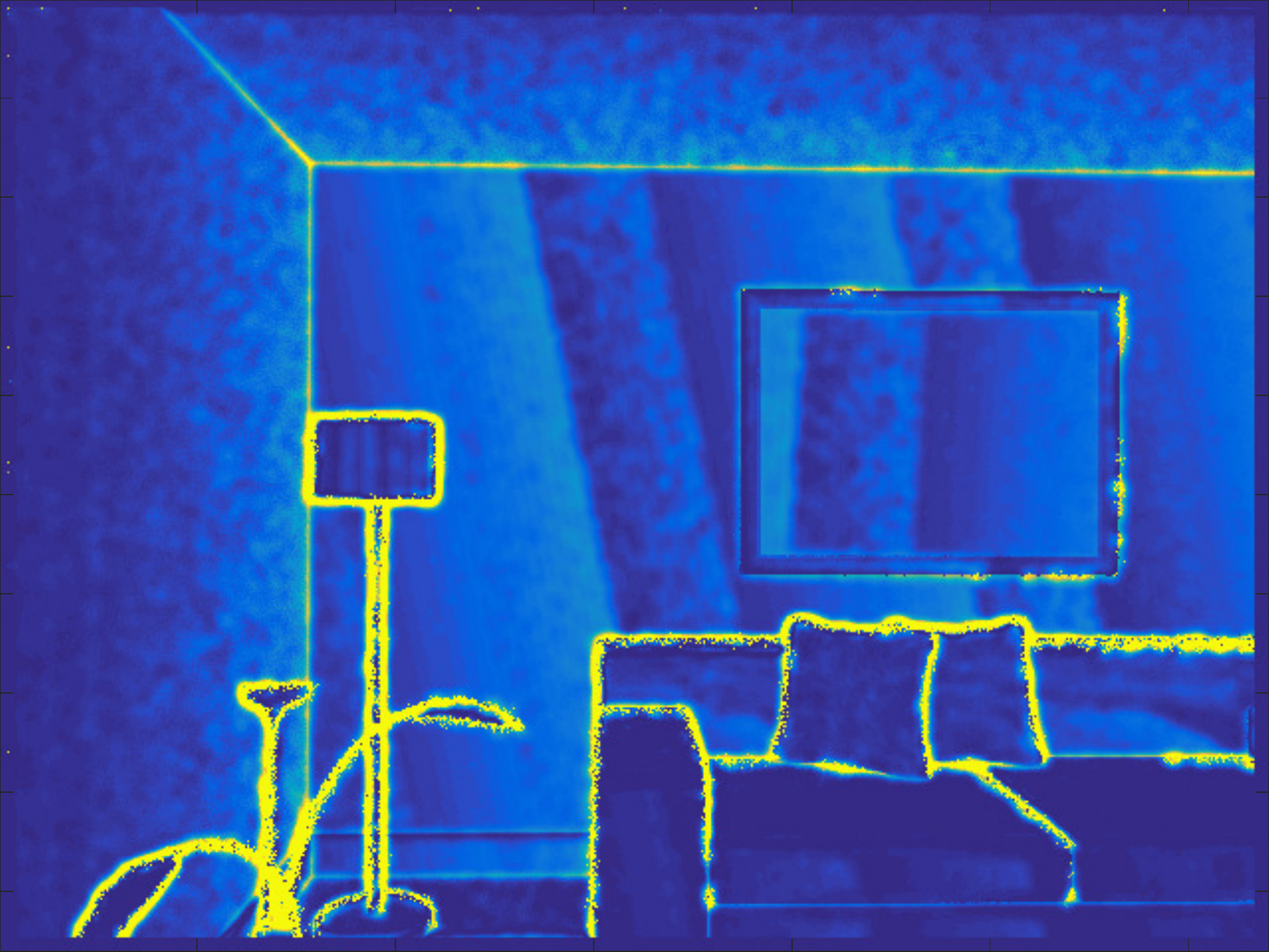} & \
		\includegraphics[scale=0.152]{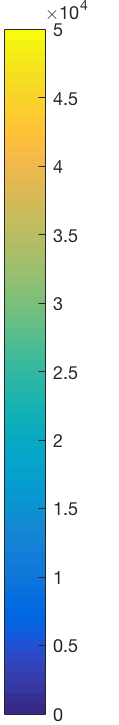} \\[-7pt]
		{\scriptsize Raw depth error: 186 mm} & {\scriptsize BF error: 175 mm}\\[-5pt]
		\includegraphics[scale=0.25]{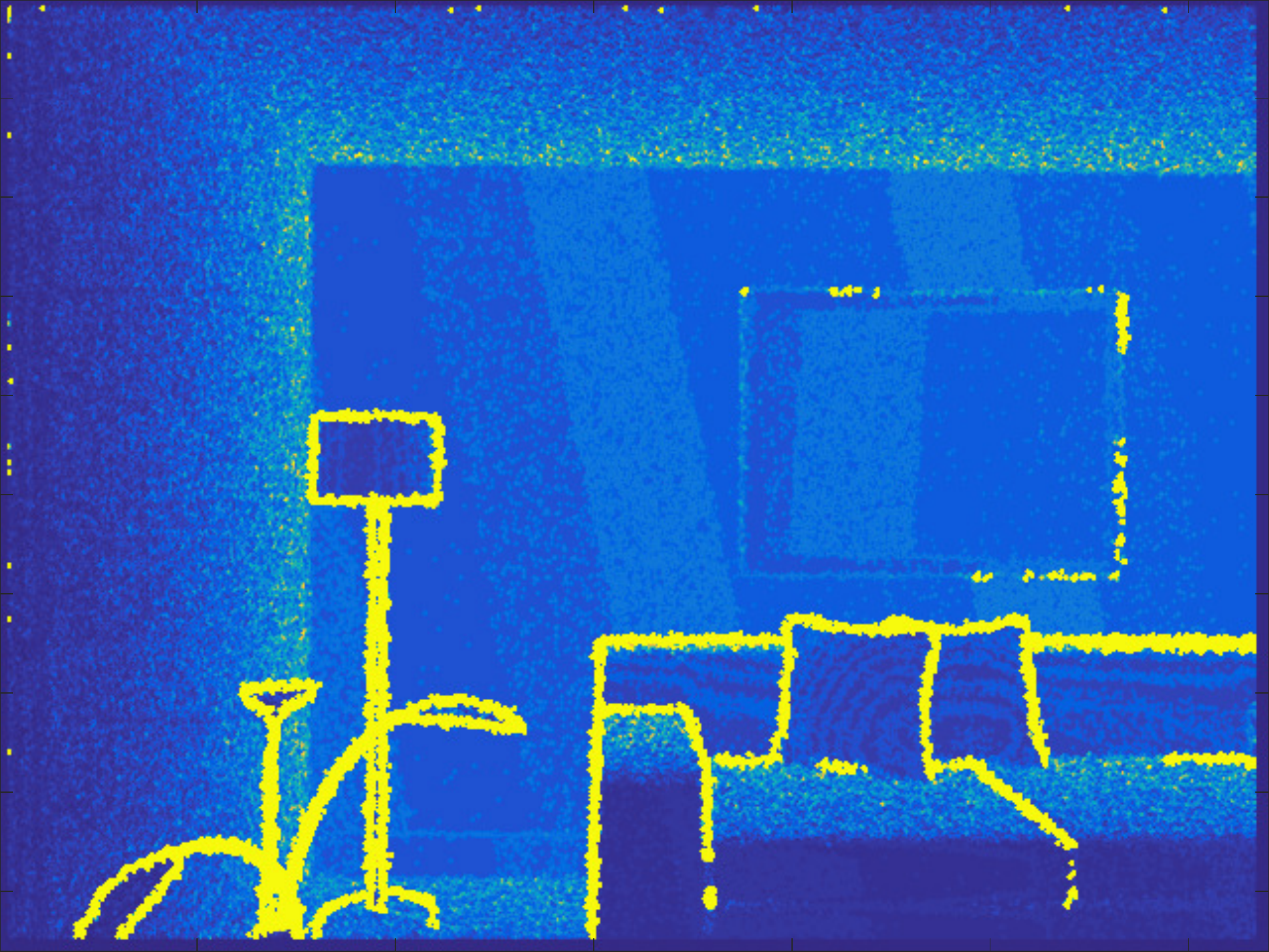} &
		\includegraphics[scale=0.25]{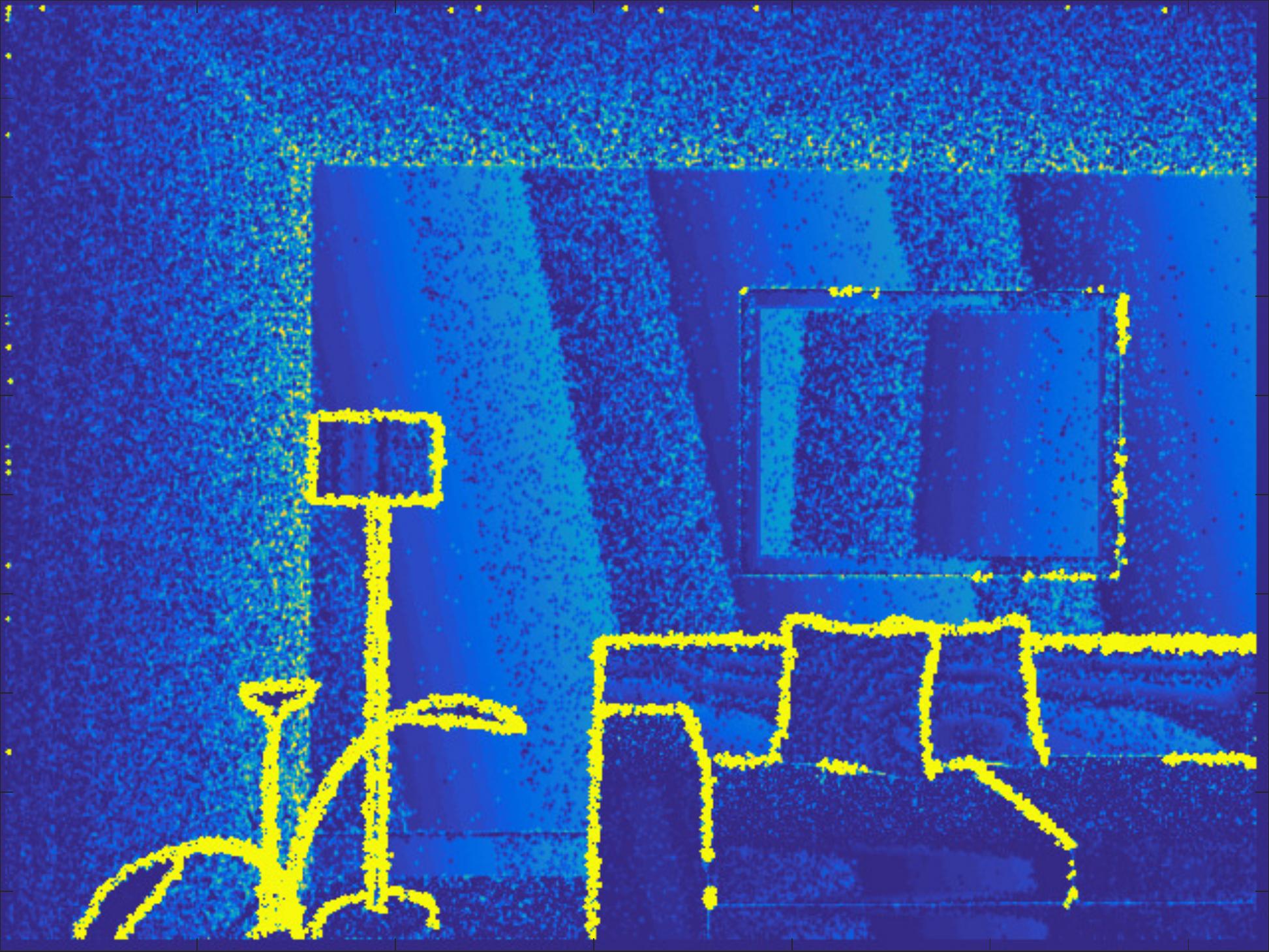} & \
		\includegraphics[scale=0.152]{figs/dynamic/uncertainty_scale.png} \\[-7pt]
		{\scriptsize C-GM uncertainty} & {\scriptsize C-GM error: 131 mm}\\ [-5pt]
		\includegraphics[scale=0.25]{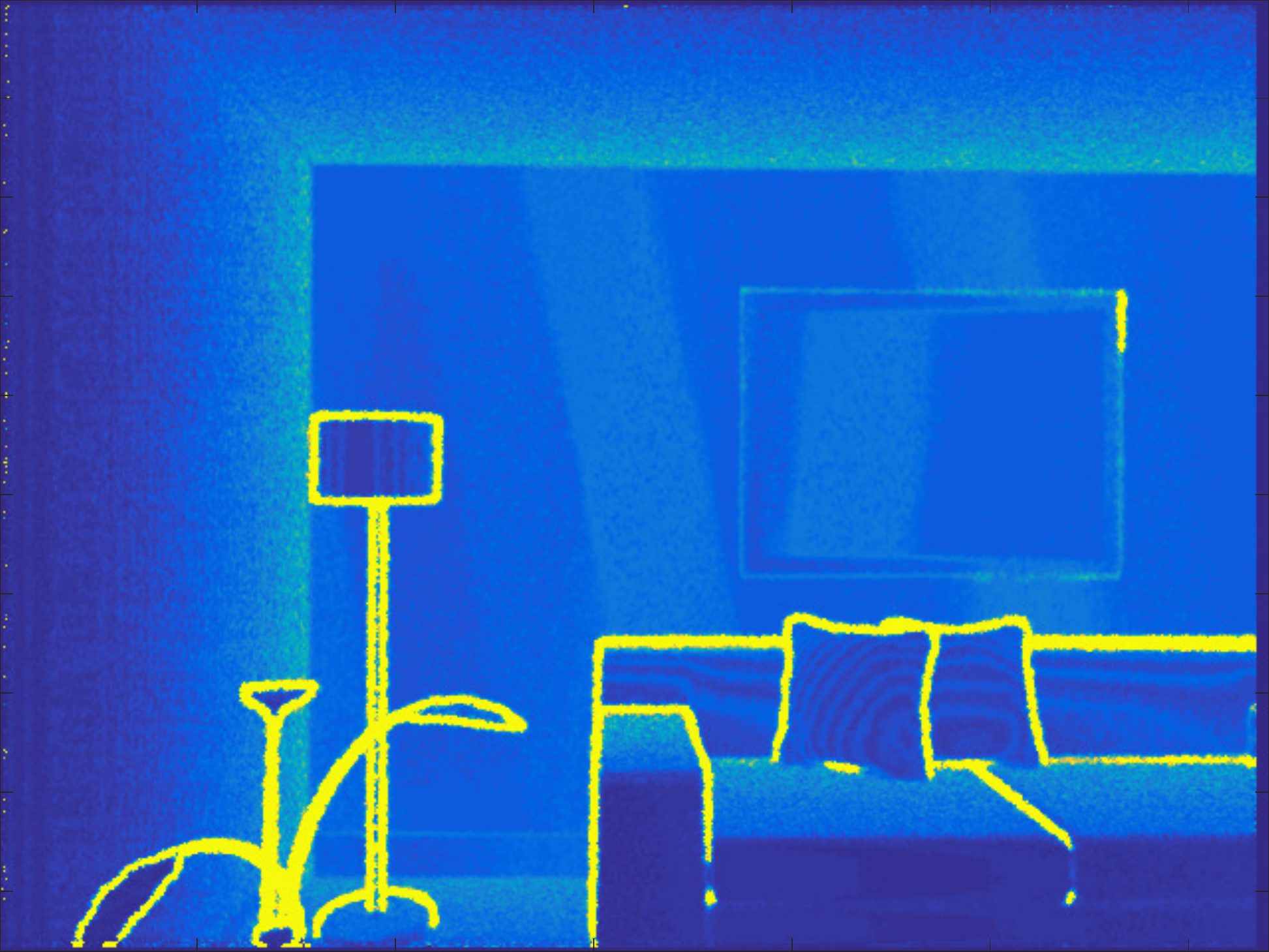} &
		\includegraphics[scale=0.25]{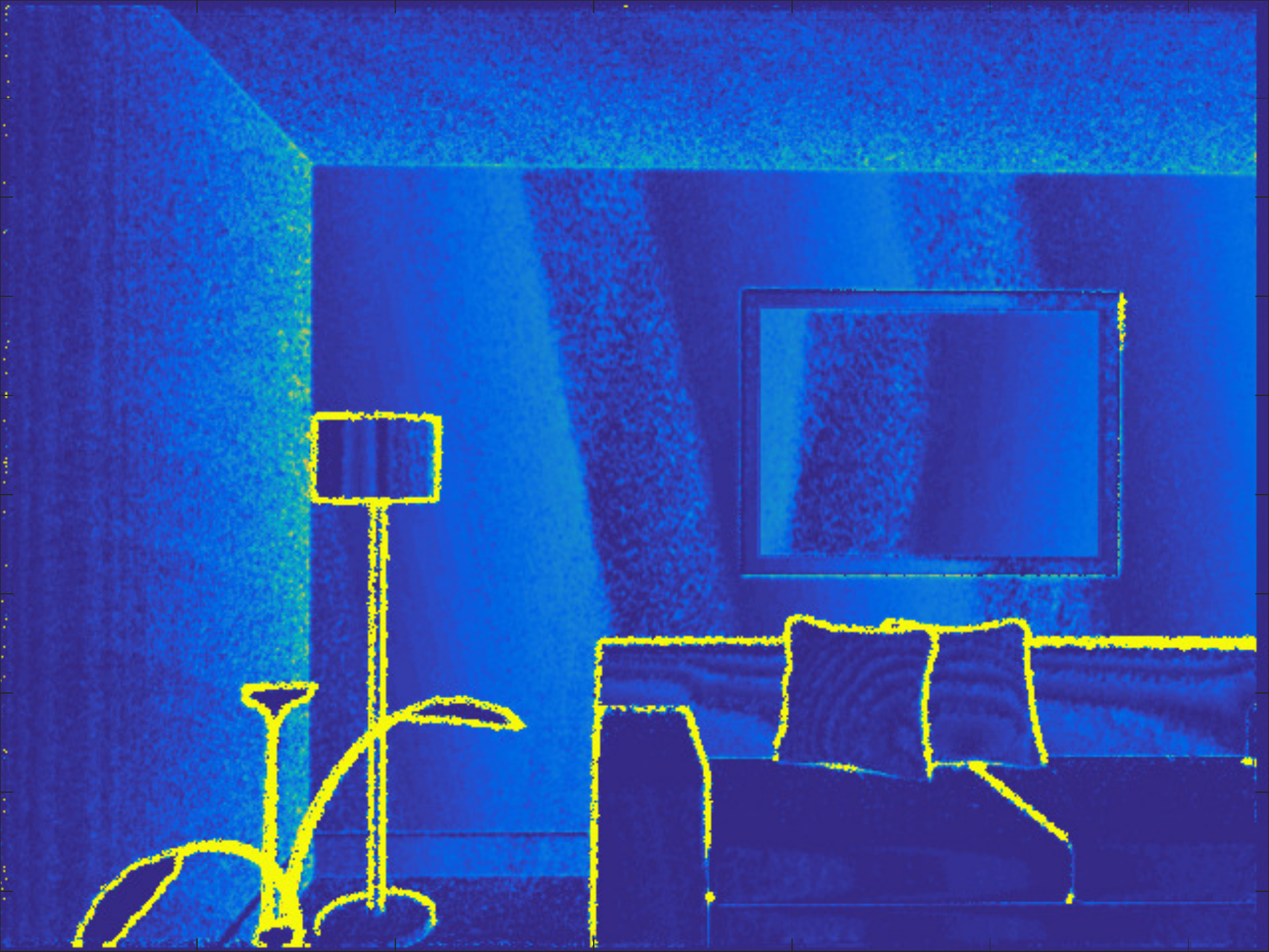} & \
		\includegraphics[scale=0.152]{figs/dynamic/uncertainty_scale.png} \\[-7pt]
		{\scriptsize  O-GM$_{10}$ uncertainty} & {\scriptsize O-GM$_{10}$ error: 96 mm}\\[-5pt]
	\end{tabular}
	\caption{Depth error and uncertainty on one frame from the synthetic ICL-NUIM dataset \cite{iclnuim}, with simulated noise, along with the respective RMSE for each depth filter. Depth error is measured by checking the available noiseless version of the depth maps. Despite the ability of the bilateral filter (BF) to preserve edges, errors are still introduced on the room edges. Notice how some of the errors on the picture frame and on the sofa are removed after applying the O-GM fusion.}
	\label{fig5}
\end{figure}

The fused range image and uncertainty can then be converted back to depth using: $z = r\cos\alpha$ and (\ref{eq:depth2range}). As can be seen in Fig. \ref{fig3} and \ref{fig5}, this framework removes significant noise from the raw depth maps. Additionally, one can see in Fig. \ref{fig5} that the depth uncertainty represents well the respective depth error in a synthetic dataset, though our sensor model differs from the one used to generate the depth noise in \cite{iclnuim}.

\begin{figure}[tb]
\centering
	\begin{tabular}{@{}c@{ }c@{ }c@{}}
         \includegraphics[scale=0.25]{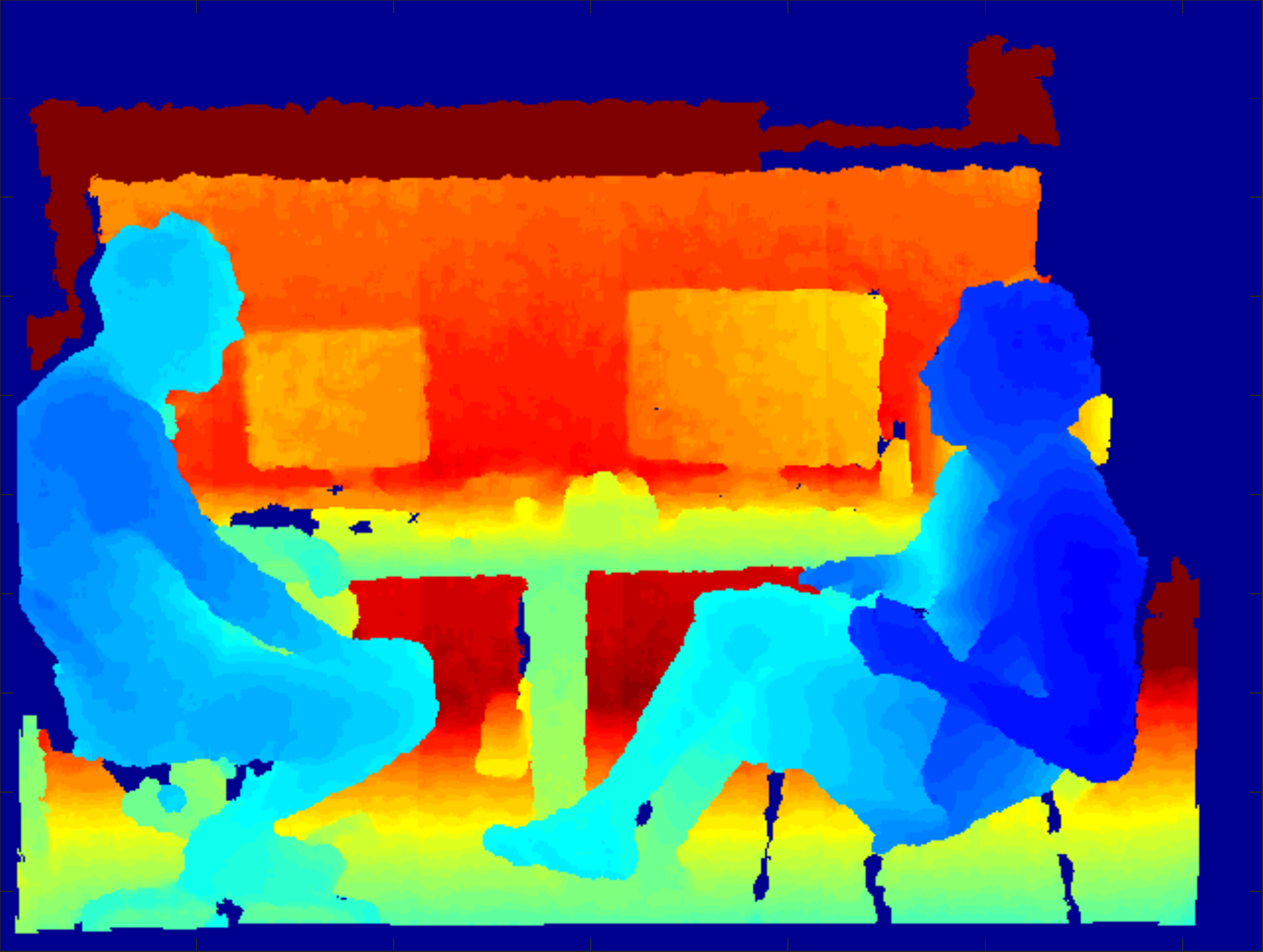} & 
	\includegraphics[scale=0.25]{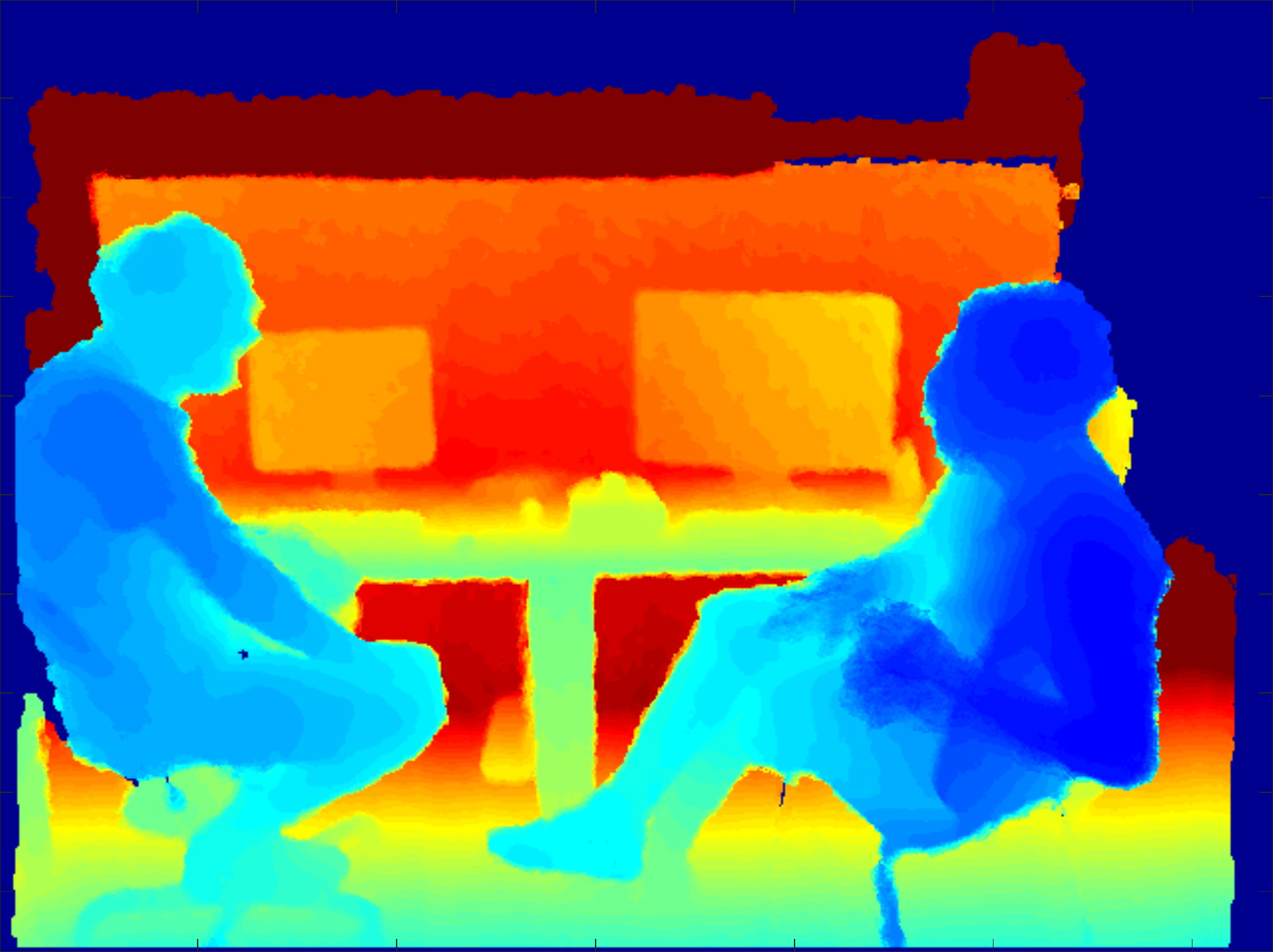} &
	 \includegraphics[scale=0.155]{figs/dynamic/scale.png} \\[-7pt]
	  {\scriptsize Raw depth map} & {\scriptsize Fused depth map} &\\[-5pt]
	 \includegraphics[scale=0.25]{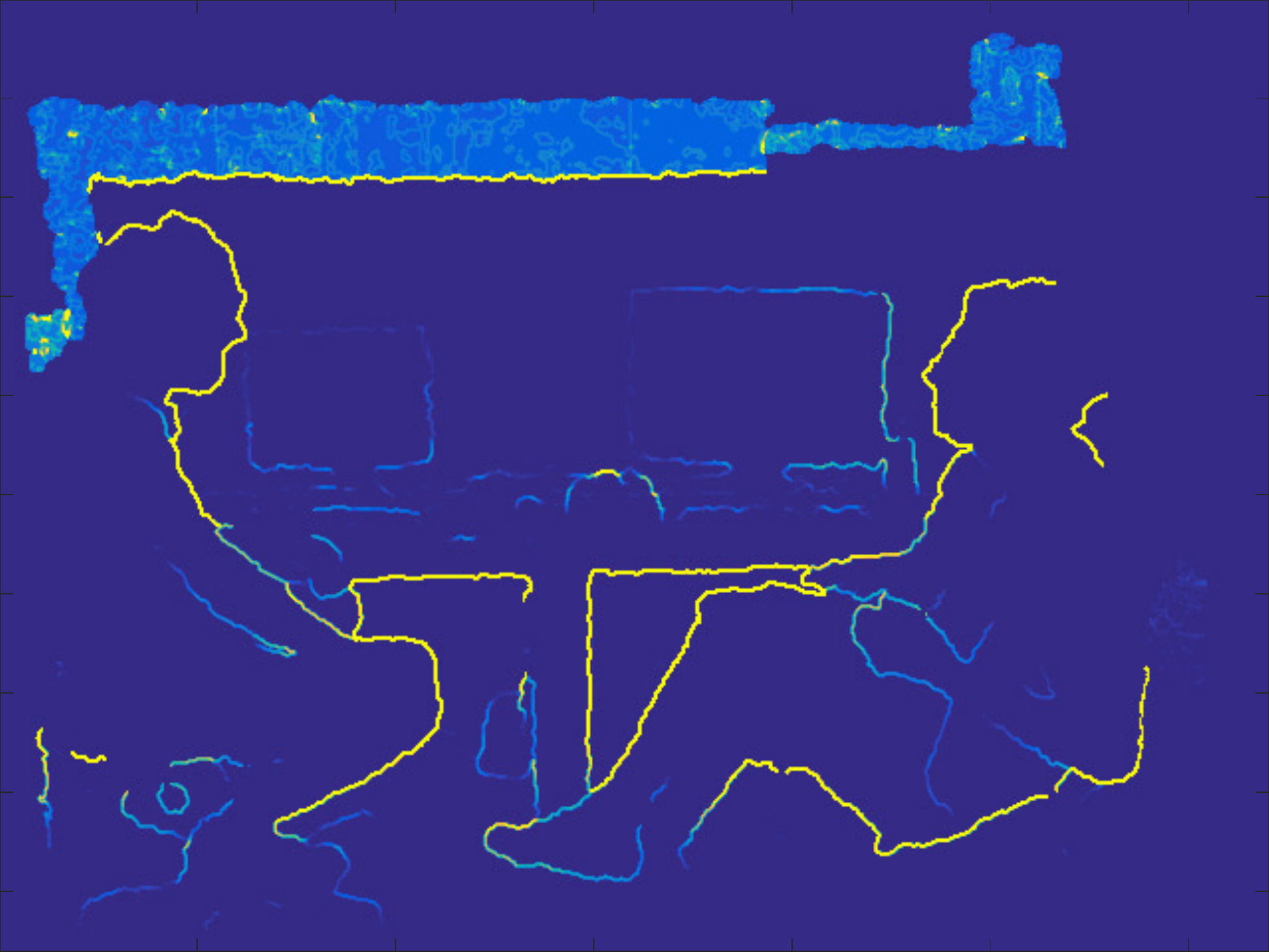} &
	\includegraphics[scale=0.25]{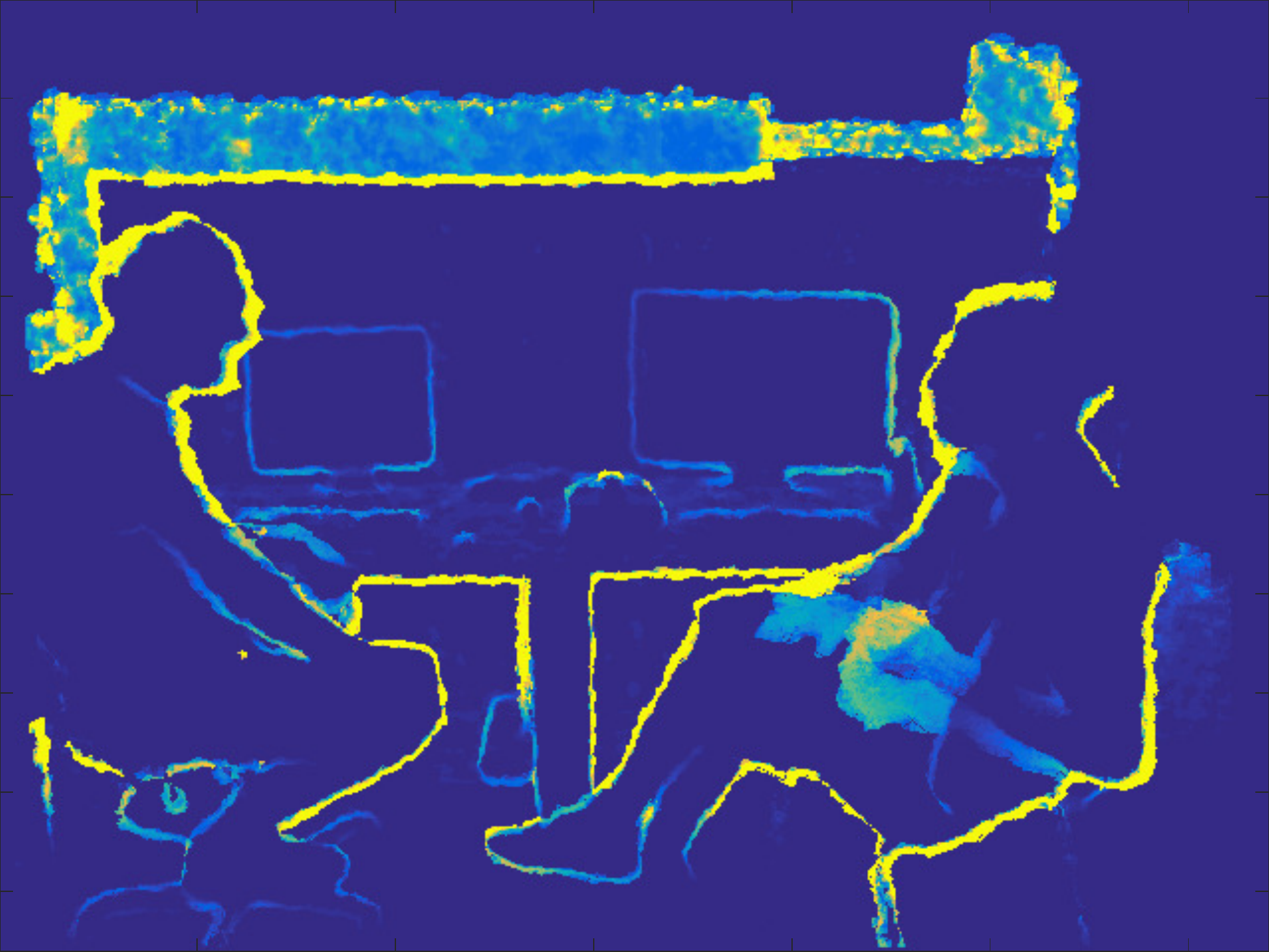} &
	\ \includegraphics[scale=0.15]{figs/dynamic/uncertainty_scale.png}  \\[-7pt]
	 {\scriptsize C-GM uncertainty} & {\scriptsize O-GM$_{10}$ uncertainty} &\\
	\end{tabular}
	\caption{Depth fusion and uncertainty in a dynamic scene, captured in \cite{tumdataset12iros}, where two people talk and gesticulate. Although depth fusion degrades the depth map around the human contours, these errors are captured by the uncertainty model (notice the hands), which has the benefit, for pose estimation, of assigning lower weight to any features detected in these regions.}
	\label{fig4}
\end{figure}

\subsection{Depth Fusion Constraints}
\label{sec:Point_cloud_registration}

Temporal fusion assumes that the transformations (i.e. stereo poses) between the frames of the sliding window are sufficiently good to fuse measurements from the same point in space. Therefore, besides using the pose estimated by the visual odometry to bring the registered point cloud to the current frame, the uncertainty of the transformations is monitored using the method described in Section \ref{sec:poseuncertainty}. If the uncertainty of a transformation exceeds a given threshold, measurements from the respective frame are removed from the registered point cloud. As a result, the length of the sliding window of frames is dynamic.\par
Furthermore, the temporal fusion is not intended for long durations and wide baselines due to: memory and computation time requirements, dynamic objects and occlusions. Parallax, due to camera translation, causes incorrect fusion of measurements from occluded background with more recent foreground measurements. To reject measurements from occlusions, we enforce a consistency constraint to old depth measurements, i.e., during the range fusion, point are projected from newest to oldest, if a pixel receives at least $k=5$ points, more points are only accepted if their ranges are within the margin: $\bar{r} \pm 3\sigma_{r}$, where $\bar{r}$ and $\sigma_{r}$ correspond to the current pixel state of range and uncertainty. \par
Fig. \ref{fig4} shows the behaviour of the filter in a dynamic environment. Although dynamic objects could be addressed by segmentation, as in \cite{selectivedepthfusion,VOSFdynamic}, our temporal fusion framework already assigns high uncertainty to the depth values that are affected by the motion of moving objects, which implicitly will downweight the features arising from the moving objects. Our results, in Section \ref{sec:experiments}, support this idea.

\section{RGB-D Odometry based on Points, Planes and Lines}
\label{sec:vis_odometry}

The proposed visual odometry method, outlined in Fig. \ref{fig6}, starts by detecting points, lines and planes from the current RGB-D frame. While 2D points and lines, along with their feature descriptors, are extracted from the intensity of the RGB channel, planes are extracted from an organized point cloud back-projected from the depth map, after applying the first and second stages of the depth filter (i.e. the depth sensor error model and the GM convolution) in order to obtain the 3D point uncertainties, which are then used by a weighted least squares plane fitting. The extracted primitives are then matched against the ones extracted from the previous frame. Resulting 3D-to-2D point and line matches and 3D-to-3D plane matches are subsequently used jointly to estimate the frame-to-frame pose, according to their uncertainties. Once the pose is estimated, a depth fused map is obtained, using the third stage of the depth filter, described in Section \ref{sec:depth_filter}. Given this new depth map, the 3D coordinates of the current point and line features are finally obtained for the next frame-to-frame pose estimation through backprojection and a weighted 3D line fitting method, which also takes into account the depth uncertainty. Furthermore, plane detection and fitting is repeated to obtain presumably better plane estimates. These modules are described in further detail below.

\begin{figure}[h]
\centering
	\includegraphics[scale=0.31]{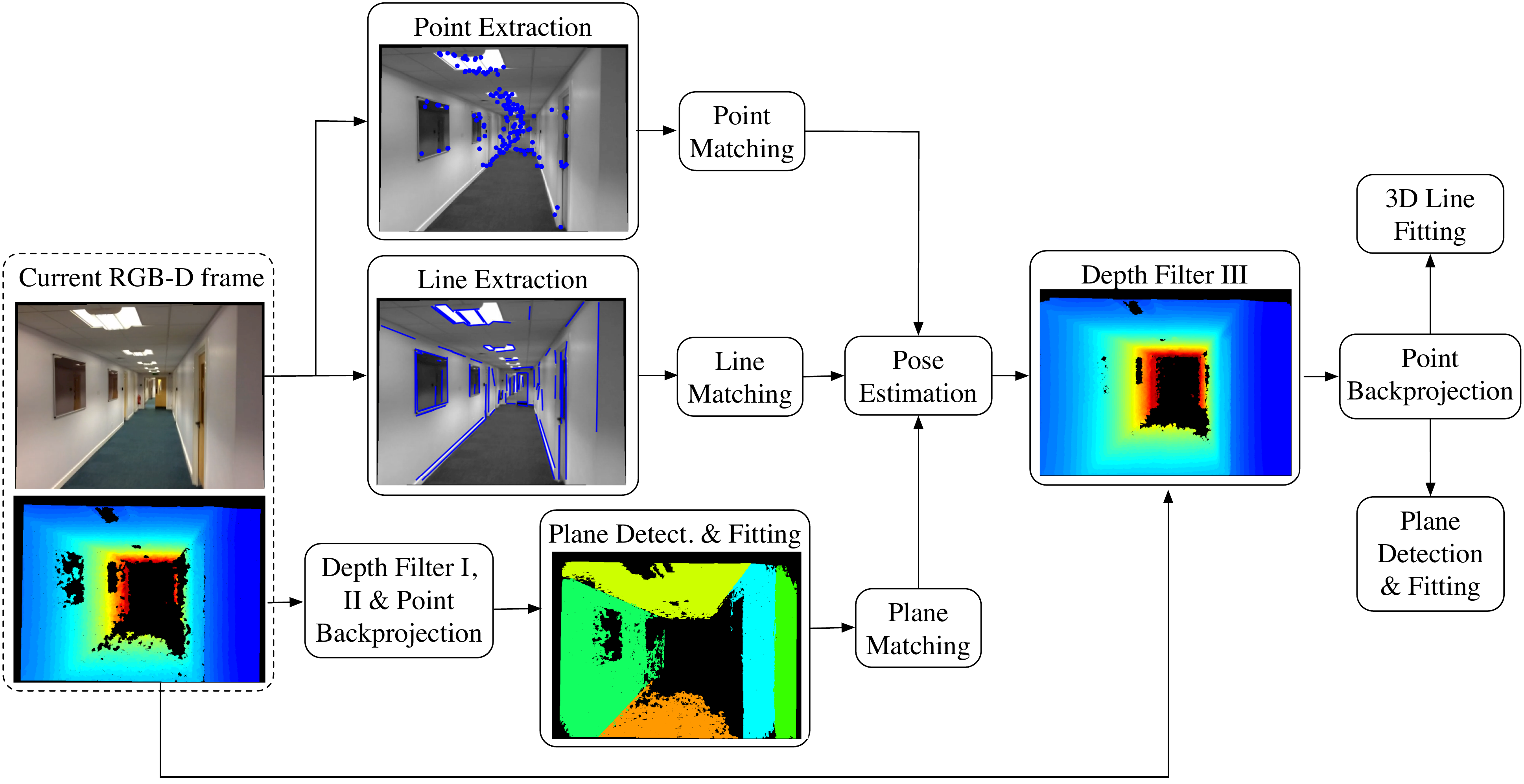}\\[-5pt]
	\caption{Visual odometry system overview}
	\label{fig6}
\end{figure}

\subsection{Extraction of Points, Lines and Planes}
Image points are detected by relying on SURF features, whereas for lines, the LSD \cite{LSD} method is used to detected line segment endpoints and then binarized LBD \cite{LBD} descriptors (implemented in OpenCV) are extracted from the respective lines. For plane extraction, once the depth map is backprojected, we make use of the method proposed in \cite{feng2014fast}, which processes efficiently organized point clouds in real-time, the result is a segmented point cloud (as depicted in Fig. \ref{fig6}). For each point cloud segment, a plane model is fit and its uncertainty is derived using the method described in Section \ref{sec:plane_fitting}.

\subsection{Point Backprojection}
Assuming that the depth image is mapped to the RGB reference frame, given the extrinsic calibration, the 3D coordinates $P=\begin{bmatrix}X, Y, Z\end{bmatrix}^\top$corresponding to a pixel $p=\{u,v\}$ on either depth or RGB image can be obtained through backprojection:
\begin{equation}
\label{eq:backprojection}
P = Z\begin{bmatrix}(u - c_x)/f_x \\[-5pt] (v - c_y)/f_y \\[-5pt] 1 \end{bmatrix}
\end{equation}
where $Z$ is the value of the depth pixel, and $\{f_x,f_y\}$ and $\{c_x,c_y\}$ are respectively the focal length and principal point of the RGB camera. The uncertainty of $P$ can be obtained by the first order error propagation: 
\begin{equation}
\label{eq:backprojection_uncertainty}
\Sigma_P = J_P \begin{bmatrix}\Sigma_p & 0 \\[-5pt] 0 & \sigma_Z^2\end{bmatrix}J_P^\top
\end{equation}
where $J_P$ is the Jacobian of (\ref{eq:backprojection}) with respect to $p$ and $Z$, $\sigma_Z^2$ is the uncertainty of the depth value given by the depth filter and $\Sigma_p$ is a $2\times2$ identity matrix times the pixel coordinate uncertainty $\sigma^2_p$, which accounts for the pixel quantization error. Let this error be modelled by a uniform PDF of length equal to 1 pixel, then its variance is $\sigma^2_p=1/12$.

\subsection{WLS Plane Fitting}
\label{sec:plane_fitting}
For plane fitting, we employ our recently proposed weighted least squares method \cite{proenca2017planes}. For the sake of completeness, we describe here the method and then propose a modification to derive a more accurate plane uncertainty. \par
It is efficient to express planes as infinite planes in the Hessian normal form: $\theta =\{N_x,N_y,N_z,d\}$. However, such representation is overparameterized, thus the estimation of these parameters by unconstrained linear least squares is degenerate. This issue has been solved in \cite{pathak2010uncertainty} by using constrained optimization and in \cite{weingarten2004probabilistic} by using a minimal plane parameterization. Similarly to \cite{weingarten2004probabilistic}, we use a minimal plane representation: $\theta_m = \begin{bmatrix}N_x,N_y,N_z\end{bmatrix}/d$, as an intermediate parameterization. Since, a plane with $d=0$ implies detecting a plane that passes through the camera center (i.e. projected as a line), it is safe to use this parameterization. The new parameters are then estimated by minimizing the point-to-plane distances through the following weighted least-squares problem:
\begin{equation}
\label{eq:lsq_fitting}
E = \sum_{i=1}^{n} \frac{w_i(\theta_mP_i +1)^2}{2}
\end{equation}
where the scaling weights were chosen to be the inverse of the point depth uncertainties: $w_i = \sigma_{Z_i}^{-2}$, which represent well the point-to-plane distance uncertainties when the detected plane is approximately parallel to the image plane. By setting the derivative of (\ref{eq:lsq_fitting}), with respect to $\theta_m$, to zero, we arrive at the solution of the form: $\theta_m^\top = A^{-1}b$, where $A = \sum_{i=1}^{n}w_iP_iP^\top_i$ and $b$ = $-\sum_{i=1}^{n}w_iP_i$. \par
Following the Fisher observed information \cite{efron1978MLE}, the covariance of $\theta_m$ is given by the inverse Hessian matrix of $E$, i.e., $\Sigma_{\theta_m} = H^{-1}$ where $H$ is simply $A$. However, the residuals are scaled by an heuristic choice of weights, and as a result $E$ is a just a scaled approximation of the negative log-likelihood function. This fact was neglected in \cite{proenca2017planes}, and as a result the uncertainty was overestimated. Therefore the weights need to be first updated with the actual uncertainty of the plane residuals $\Sigma_{r_i}$. These can be found at the solution $\theta_m$ by propagating the point uncertainties $\Sigma_{P_i}$ as follows:  $\Sigma_{r_i} = \theta_m\Sigma_{P_i}\theta^\top_m$. The uncertainty of $\theta_m$ is then derived from the updated matrix $A$.
Finally, the Hessian normal form can be recovered by:
\begin{equation}
\label{eq:plane_params}
\theta = \frac{\begin{bmatrix} \theta_m & 1\end{bmatrix}}{\|\theta_m\|}
\end{equation}
and the respective uncertainty is obtained via first order error propagation: $\Sigma_{\theta}  = J_{\theta}  \Sigma_{\theta_m} J_{\theta}^\top$, where $J_{\theta}$ is the Jacobian of (\ref{eq:plane_params}).

\subsection{WLS Line Fitting}

The proposed solution to 3D line fitting is illustrated in Fig. \ref{fig6}. First, as in \cite{PLVO}, depth pixels are sampled uniformly across the 2D line segments (maximum 100 pixels per line). The pixels with available depth are backprojected to 3D points and then these are processed by a RANSAC loop based on 3D point-to-line Euclidean distances to remove outliers. We believe that, in this work, the Euclidean distance is more adequate to remove outliers than the Mahalanobis metric proposed in \cite{PLVO}, due to the high uncertainties given by the GM convolution at depth discontinuities (see Fig. \ref{fig4}). The final consensus set of 3D points is denoted as $P =\{P_1,...,P_n\}$.

\begin{figure}[t]
	\centering
	\includegraphics[scale=0.75]{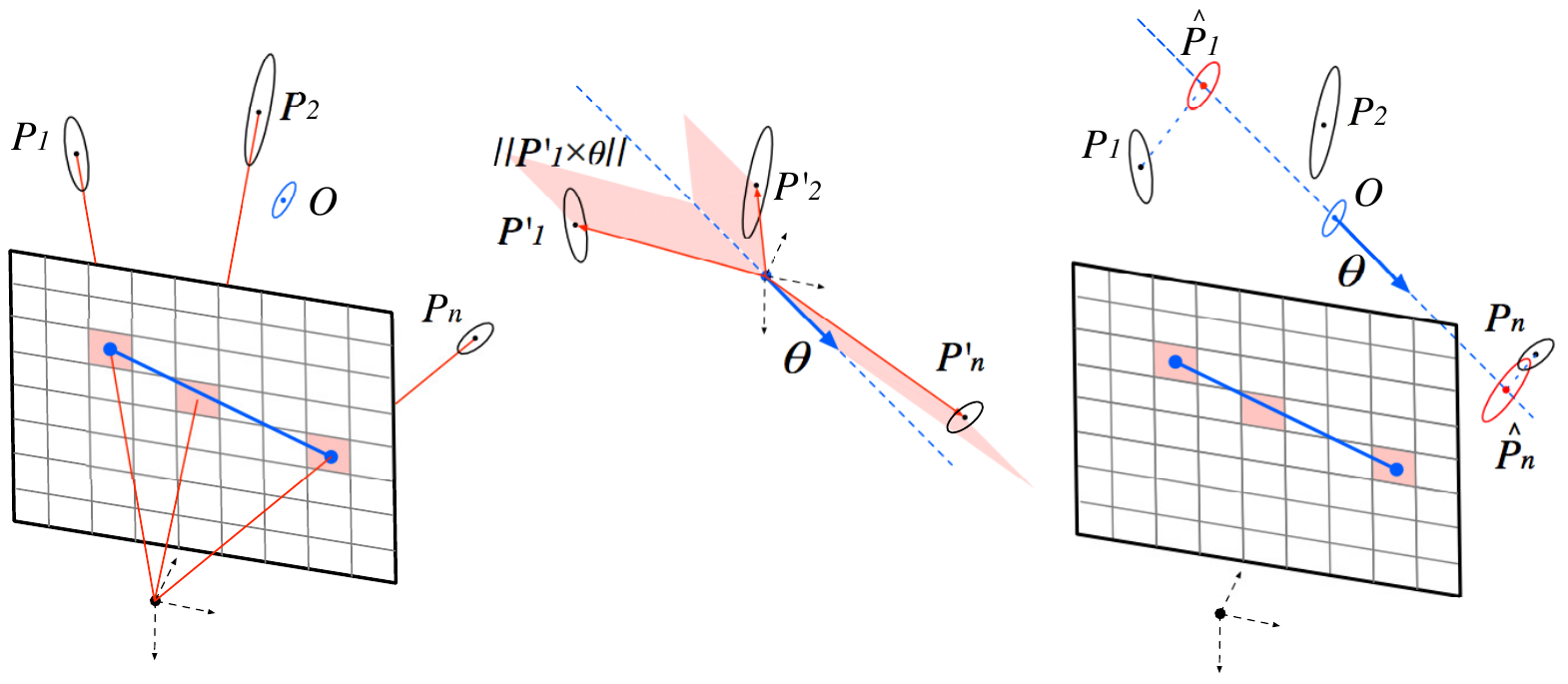}
	\\[-5pt]
	 \hspace{-30pt} (a) \hspace{100pt} (b) \hspace{90pt} (c)
	\caption{Illustration of the solution proposed to 3D line fitting. (a) After sampling and backprojecting depth pixels across the 2D line, the center of mass $O$ is determined. (b) Points are translated so $O$ is in the origin, and the vector $\theta$ is estimated by minimizing the sum of the shaded areas, which correspond to the cross product norms. (c) Line endpoints are selected by projecting the extreme points $P_1$ and $P_n$ on $\theta$.}
	\label{fig6}
\end{figure}

The problem of fitting a 3D line to 3D points can be solved non-iteratively by casting it as 2D vector estimation problem.
The key idea is to exploit the fact that the optimal line passes in the center of mass, denoted as $O$, by estimating a line pinpointed at $O$, as depicted in Fig. \ref{fig6}. The centroid $O$ corresponds to the MLE: $(\sum_{i=1}^{n}W_i)^{-1}\sum_{i=1}^{n}W_iP_i$ where $W_i=\Sigma^{-1}_{P_i}$ and the MLE variance is $(\sum_{i=1}^{n}W_i)^{-1}$. But for efficiency, we instead approximate $O$ as the mean of the $n$ points weighted by the inverse of their depth variances, in order to avoid inverting the $n$ covariance matrices, required by the MLE. This approximation implies a covariance isotropic assumption.
Therefore, $O = \sum_{i=1}^{n}w_iP_i$ where $w_i$ is a normalized weight: $\sigma_{Z_i}^{-2}/\sum_{j=1}^{n}\sigma_{Z_j}^{-2}$ and the respective covariance is $\Sigma_O = \sum_{i=1}^{n}w^2_i\Sigma_{P_i}$. \par
The point samples are then translated by subtracting $O$: $P'_i = P_i-O$ in order to estimate a vector $\theta$ by minimizing the magnitudes of the cross products between $\theta$ and the vectors $\overrightarrow{OP_i}$ through the following weighted least squares cost function:
\begin{equation}
\label{eq:line_fitting_cost_fx}
E = \sum_{i=1}^{n}\frac{w_i\|  P'_i\times\theta \|^2}{2}
\end{equation}
Once again, the chosen weights are $w_i = \sigma_{Z_i}^{-2}$. Since the 3D vector $\theta$ is overparameterized, we reduce it to 2D by fixing one of its dimensions $\theta^{(k)}$ at 1. This dimension cannot be chosen arbitrary, as the optimal vector may have zero entries. Thus, we select the dimension where the range of samples is the highest. Given the resulting 2D parameterization $\theta_m$ and by setting the partial derivatives of (\ref{eq:line_fitting_cost_fx}) equal to zero, we arrive at a solution of the form $\theta_m = A^{-1}b$ with three possible results for $A$ and b depending on the fixed dimension: \par
\vspace{2mm}
\noindent
$(\theta^{(1)}=1)$
\begin{equation}
\label{eq:line_fitting_1}
A =  \begin{bmatrix}  
\sum_{i=1}^{n}w_i(X^2_i+Z^2_i) &  -\sum_{i=1}^{n}w_iY_iZ_i\\
-\sum_{i=1}^{n}w_iY_iZ_i &   \sum_{i=1}^{n}w_i(X^2_i+Y^2_i)
\end{bmatrix}, \
b = \begin{bmatrix}  
\sum_{i=1}^{n}w_iX_iY_i\\
\sum_{i=1}^{n}w_iX_iZ_i 
\end{bmatrix} 
\end{equation}
\\
$(\theta^{(2)}=1)$
%\vspace{-2em}
%\end{equation*}
\begin{equation}
\label{eq:line_fitting_2}
A =  \begin{bmatrix}  
\sum_{i=1}^{n}w_i(Y^2_i+Z^2_i) &  -\sum_{i=1}^{n}w_iX_iZ_i\\
-\sum_{i=1}^{n}w_iX_iZ_i &   \sum_{i=1}^{n}w_i(X^2_i+Y^2_i)
\end{bmatrix}, \
b = \begin{bmatrix}  
\sum_{i=1}^{n}w_iX_iY_i\\
\sum_{i=1}^{n}w_iY_iZ_i 
\end{bmatrix} 
\end{equation}
\\
$(\theta^{(3)}=1)$
%\vspace{-2em}
%\end{equation*}
\begin{equation}
\label{eq:line_fitting_2}
A =  \begin{bmatrix}  
\sum_{i=1}^{n}w_i(Y^2_i+Z^2_i) &  -\sum_{i=1}^{n}w_iX_iY_i\\
-\sum_{i=1}^{n}w_iX_iY_i &   \sum_{i=1}^{n}w_i(X^2_i+Z^2_i)
\end{bmatrix}, \
b = \begin{bmatrix}  
\sum_{i=1}^{n}w_iX_iZ_i\\
\sum_{i=1}^{n}w_iY_iZ_i 
\end{bmatrix} 
\end{equation}
where, here, $P'_i = \{X_i,Y_i,Z_i\}$ for readability. To obtain $\Sigma_{\theta_m}$, as explained in the last section, the weights need to be rectified with the inverse of the uncertainties of the residuals $r_i =\|  P'_i\times\theta \|$ through first order error propagation: $\Sigma_{r_i} = J_{r_i}\Sigma_{P_i}J_{r_i}^\top$. When $r_i$ is exactly zero, $J_{r_i}$ is indeterminate, thus we add a small perturbation to $P'_i$ to avoid such case. Once A is rectified, $\Sigma_{\theta}$ is found by restructuring $\Sigma_{\theta_m} = A^{-1}$ as a 3$\times$3 matrix, where the entries corresponding to the fixed dimension are 0. \par
Finally, estimated line endpoints $\{\hat{P_1},\hat{P_n}\}$ can be sampled through interpolation as follows:
\begin{equation}
\label{eq:line_interpolation}
\hat{P_i} = O + \lambda_i \theta
\end{equation}
where $\lambda_i$, denoting the interpolation factor for each endpoint, is obtained by projecting the measured line endpoint onto the estimated line: $\lambda_i = \theta^\top P'_i/\|\theta\|^2$. The endpoint uncertainties are then given by propagating $\Sigma_{\theta}$ and $\Sigma_{O}$ through (\ref{eq:line_interpolation}).

\subsection{Matching Points, Lines and Planes}

For 2D points and lines, feature correspondences are established between successive frames by matching their descriptors using a $k$-NN search and then select the strongest match per query that satisfies an image geometric distance constraint: the image coordinates of point matches must be within a certain Euclidean distance and 2D line matches must have a similar slope angle and distance to origin (i.e. image top-left corner), according to their line Hessian normal parameterization. \par
Planes are matched between successive frames using the approach proposed in \cite{proenca2017planes}, as follows: First, 1-to-N candidate matches are obtained by enforcing the following constraints:
\begin{itemize}
\item Projection overlap: The projections of two planes, defined as the image segments covered by the inliers of the planes, must have an overlap of at least 50\% the number of plane inliers of the smallest plane. This can be checked efficiently by using bitmask operations after checking the geometric constraint.
\item Geometric constraint: Given the Hessian plane equations of two planes: $\{N,d\}$ and $\{N',d'\}$, the angle between the plane normals: $\arccos(N \cdot N')$ must be less than 10$^\circ$ and the distance: $\lvert d-d'\rvert$  must be less than 10 cm.
\end{itemize}

\begin{figure}
\centering
	\includegraphics[scale=0.6]{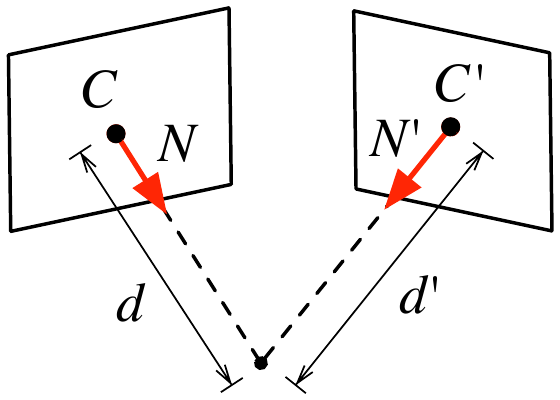}
	\caption{Geometry of two planes and their representation as points: $\{C,C'\}$}
	\label{fig7}
\end{figure}

To select the best plane match between the plane candidates, we select the plane candidate that yields  the minimum plane-to-plane distance, a concept introduced in \cite{proenca2017planes}, described as follows: Let $\{N',d'\}$ and $\{N,d\}$ be the equations of two planes then the distance between the two planes is expressed by:
\begin{equation}
\label{eq:plane_dist}
\lVert C - C' \rVert = \lVert d'N' - dN \rVert
\end{equation}
where $C$ and $C'$ represent points on the planes, as shown in Fig. \ref{fig7}.

\subsection{Pose Estimation}

Pose, defined as the 3D rigid body transformation: $\{R,t \mid R \in SO(3), t \in \mathbb{R}^3\}$, is estimated by jointly minimizing the point and line 3D-to-2D reprojection errors and the 3D plane-to-plane distances. While, pose could be alternatively estimated, as shown in \cite{proenca2017planes} and \cite{PLVO}, by minimizing the 3D Mahalanobis distance between point matches, the unidirectional reprojection error tolerates missing depth values and it could be less affected by a depth error that is incorrectly modelled by the uncertainty. \par
Given a 3D-to-2D point match $\{P,p'\}$, the reprojection error is expressed in the vector form as follows:\begin{equation}
\label{eq:pt_err}
\widetilde{p}  = (p' - \pi(RP +t\,))^\top
\end{equation}
whereas the residual of a 3D-to-2D correspondence of line segments is expressed as the point-to-line distance between a 2D line: $l'$ and the projection of the corresponding 3D line endpoints $\{P_1,P_2\}$: 
\begin{equation}
\label{eq:line_err}
\widetilde{l}  = l'\begin{bmatrix} \pi(RP_1+t\,) & \pi(RP_2+t\,) \\[-5pt] 1 & 1\end{bmatrix}
\end{equation}
For two plane matches: $\{N,d\}$ and $\{N',d'\}$, we make use of the plane-to-plane distance, defined in (\ref{eq:plane_dist}), such that, the residual can be derived, in the vector form, as: 
\begin{equation}
\label{eq:plane_r}
\widetilde{C} = N'R(N't+d') - dN
\end{equation}
Given a set  of point matches: $S_1$, a set of line matches: $S_2$ and a set of plane matches: $S_3$, we minimize the following joint cost function, by using a Levenberg-Marquart algorithm:
\begin{equation}
\label{eq:objective_fx}
\begin{aligned}
E = \sum_{i=1}^{S_1}\tilde{p_i}^2w(\tilde{p_i}) + \sum_{i=1}^{S_2}\tilde{l_i}^2w(\tilde{l_i}) + \alpha\sum_{i=1}^{S_3}\widetilde{C_i}^2w(\widetilde{C_i})
\end{aligned}
\end{equation}
where $w(r_i)$ is a function that computes a vector of weights for a given residual $r_i$ based on its uncertainty $\Sigma_{r_i}$. Since $\Sigma_{r_i}$ depends on the pose parameters, the weights are recomputed in an iteratively re-weighted least-squares fashion. The residual uncertainties are derived through first order error propagation of (\ref{eq:pt_err}, \ref{eq:line_err} and \ref{eq:plane_r}) given the uncertainties of the respective extracted primitives (i.e. 3D points, line endpoints and plane equations).
Then, $w(r_i)$ returns simply the inverse of the diagonal entries of $\Sigma_{r_i}$. Although, this means that the covariances between dimensions are neglected for points and planes, this allows maintaining the residuals as vectors in the least squares problem, which we have found to improve the convergence, and it does not require inverting the covariance matrices.
\par
Despite the residual weighting, we found necessary to use a fixed scaling factor $\alpha$ to tune the impact of the plane residuals on the pose optimization. In this work, we have used the trade-off $\alpha=0.02$, based on coarse tuning. Furthermore, an M-estimator with Tukey weights is used to further reweight the point and line residuals in order to down-weight the impact of outliers, whereas plane matching outliers are already addressed by the plane matching method and plane matches are typically too few to rely on statistics.

\subsubsection{Pose Uncertainty}
\label{sec:poseuncertainty}
Assessing the uncertainty of the estimated pose allows: (i) to detect degenerate feature configurations and reject pose estimates under such configurations and (ii) to avoid inaccurate depth fusion due to pose errors. The derivation of the pose uncertainty is described below. \par
During the pose optimization, the rotation is parameterized locally as a three-dimensional representation of an unit quaternion: $\{q_1,q_2,q_3\}$ such that
$q_4 = \sqrt{1-q_1^2-q_2^2-q_3^2}$. Let the pose parameters be: $\xi =\{t_x,t_y,t_z,q_1,q_2,q_3\}$, then its uncertainty can be approximated by back-propagation \cite{hartley2003multiple}:
\begin{equation}
\label{eq:pose_var}
\Sigma_{\xi}=(J_rWJ_r^\top)^{-1}
\end{equation}
where $J_r$ is the stacked Jacobian matrix of the residuals with respect to the pose parameters and $W$ is a diagonal matrix that contains the weights assigned to the residuals. To validate the pose estimate, we simply check if the largest eigenvalue of the matrix block corresponding to the translation vector is larger than a given threshold, if so, the optimized pose is ignored and a decaying velocity model is used instead. To further assess the pose drift between frame $1$ and $k+1$: $\Sigma^{(k+1|1)}_{\xi}$, the transformation uncertainty can be propagated using the EKF state covariance propagation:
\begin{equation}
\label{eq:pose_prop_var}
\Sigma^{(k+1|1)}_{\xi} = F\Sigma^{(k|1)}_{\xi}F^\top + GQG^\top
\end{equation}
where $Q$, known as the process noise covariance, is given by (\ref{eq:pose_var}), and $F$ and $G$ are the Jacobian matrices of the first 6 columns of the following state transition equation, with respect to $\xi^{(k|1)}$ and  $\xi^{(k+1|k)}$, respectively:
\begin{equation}
\label{eq:pose_comp}
f = \begin{bmatrix} t^{(k+1|1)} \\ q^{(k+1|1)} \end{bmatrix}  =\begin{bmatrix} R^{(k+1|k)} t^{(k|1)} + t^{(k+1|k)} \\ q^{(k+1|k)} \otimes q^{(k|1)} \end{bmatrix} 
\end{equation}
where $\otimes$ denotes the quaternion product. As mentioned in Section \ref{sec:Point_cloud_registration}, this framework is used to validate the propagation of depth measurements from the sliding window of frames. Specifically, the uncertainties of the transformations between the current frame and the frames where the depth measurements were taken are continuously updated using (\ref{eq:pose_prop_var}) and validated based on the largest eigenvalue criterion (described above).

\section{Experiments and Results}
\label{sec:experiments}

In order to evaluate the proposed dataset, we tested our method on various sequences from two public RGB-D datasets: the TUM benchmark \cite{tumdataset12iros} and the synthetic ICL-NUIM \cite{iclnuim} benchmark. The results of this evaluation are reported and discussed in the next section. Additionally, we have captured four RGB-D sequences in structured environments using the setup shown in Fig. \ref{fig9} and evaluated the trajectory estimated by our method, in Section \ref{sec:author_dataset}. Throughout the experiments, we fixed the maximum length of the depth fusion sliding window at 10 frames. Timing results are reported and discussed in Section \ref{sec:processing_time}.

\subsection{Public datasets}

\begin{figure}[!h]
	\centering
	\begin{tabular}{@{\hspace{-10pt}}c@{\hspace{-10pt}}c@{\hspace{-10pt}}c@{\hspace{-10pt}}}
		{\tiny fr1/desk} & {\tiny  fr1/360}& {\tiny fr2/360\_hemisphere}\\[-4pt]
		\includegraphics[scale=0.3]{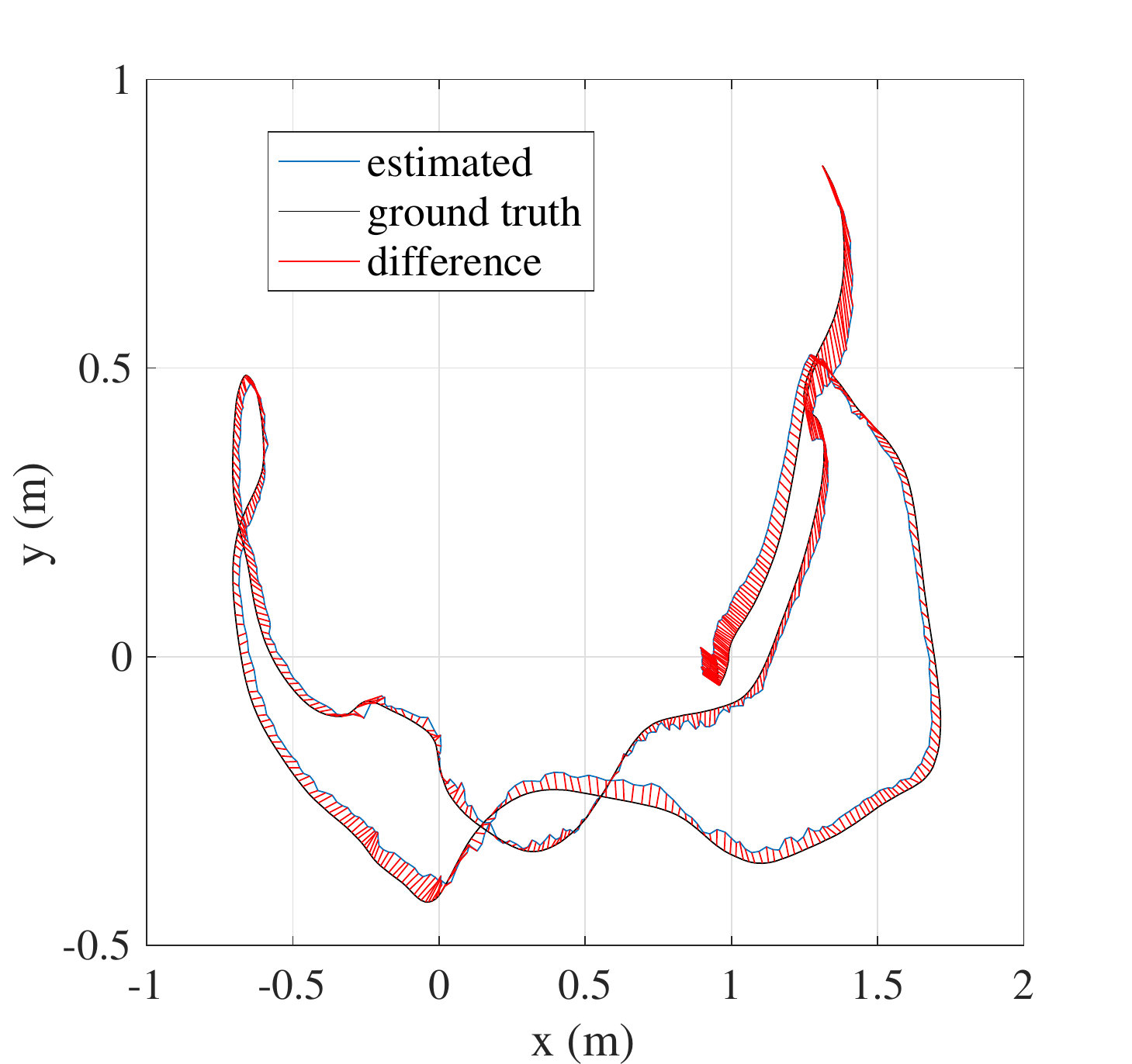} & 
		\includegraphics[scale=0.3]{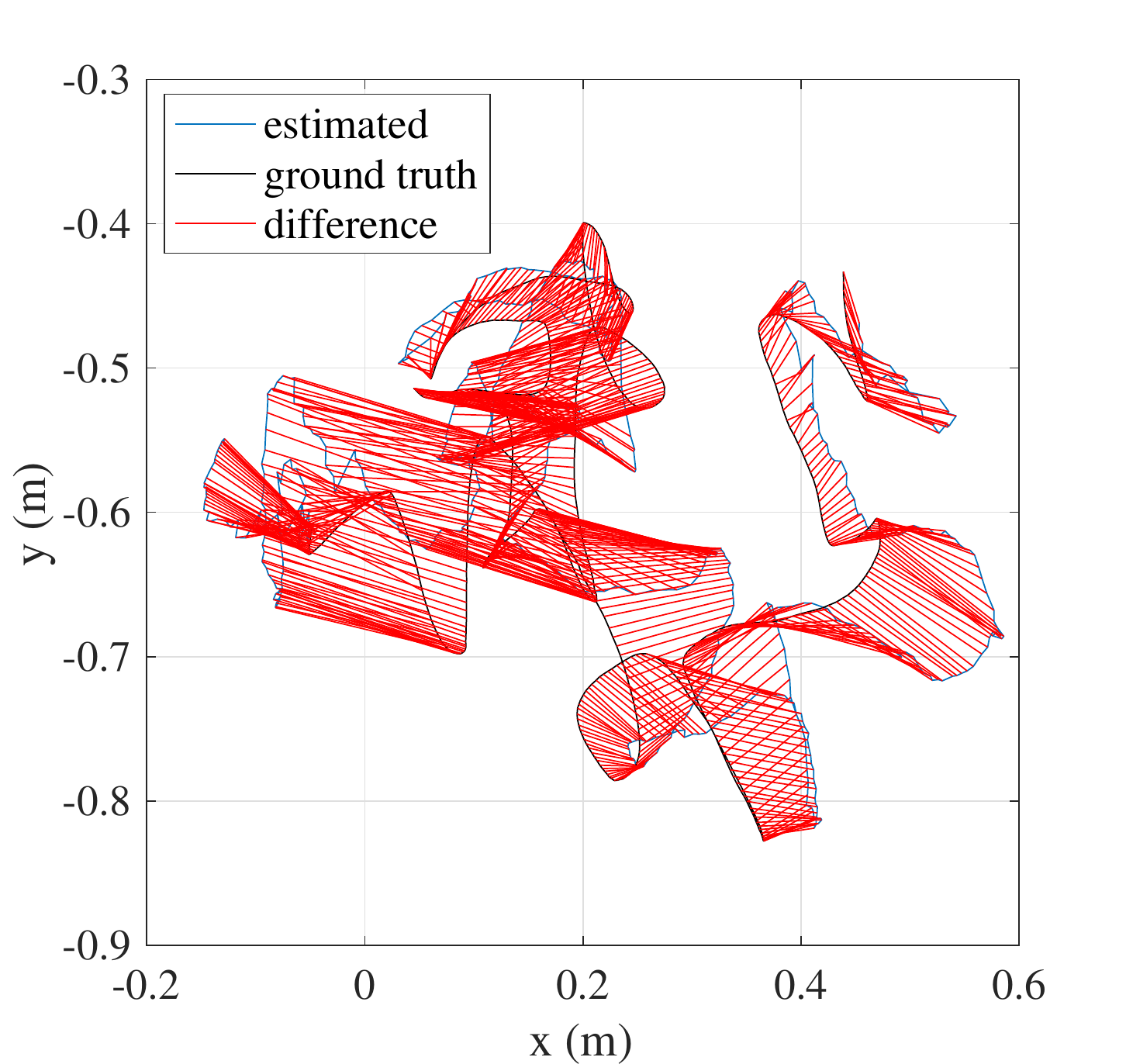} &
		\includegraphics[scale=0.3]{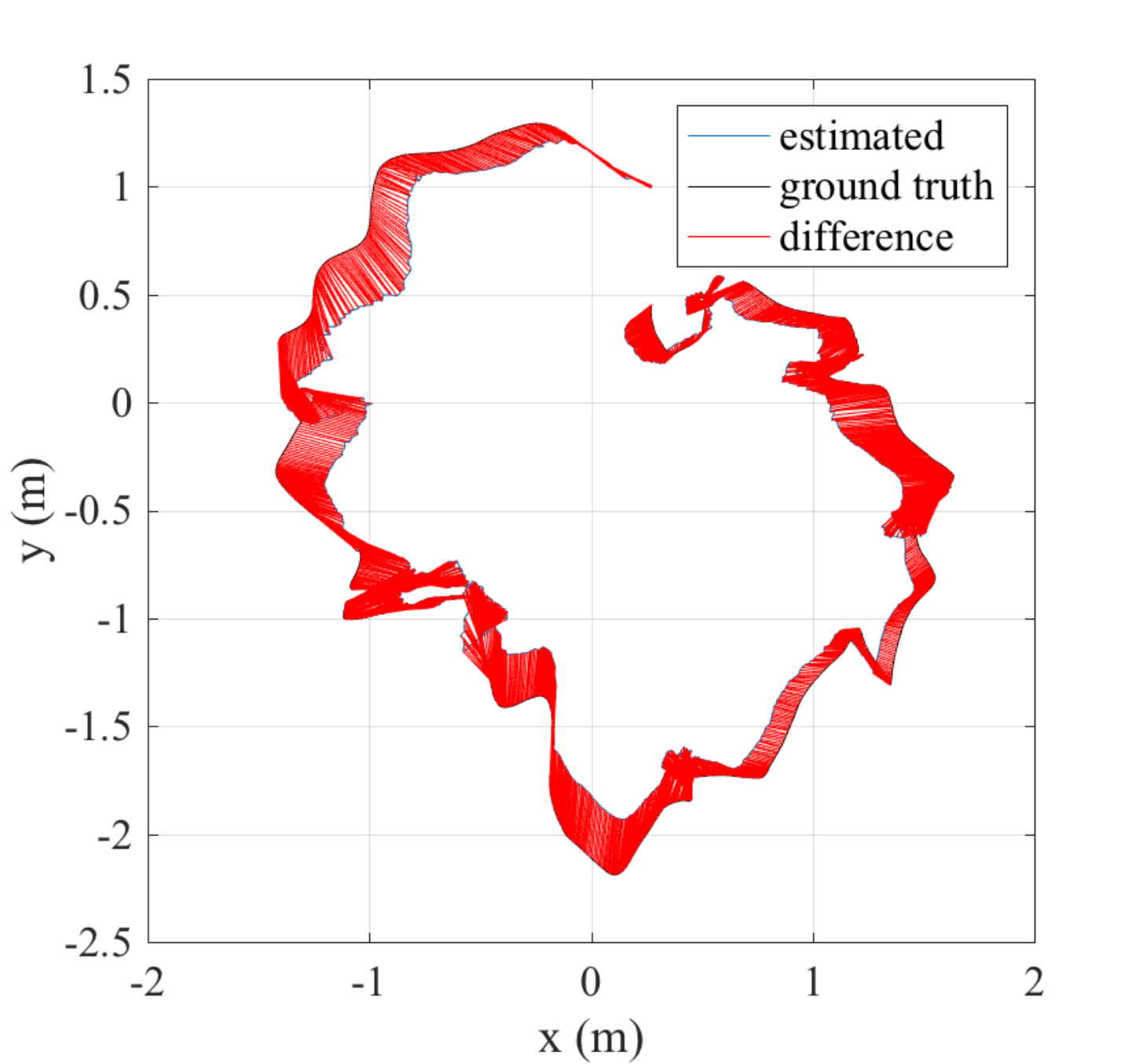} \\[-2pt]
		{\tiny fr3/struct\_no\_text\_far} & {\tiny fr3/cabinet}& {\tiny kt0 (lr)}\\[-4pt]
		\includegraphics[scale=0.3]{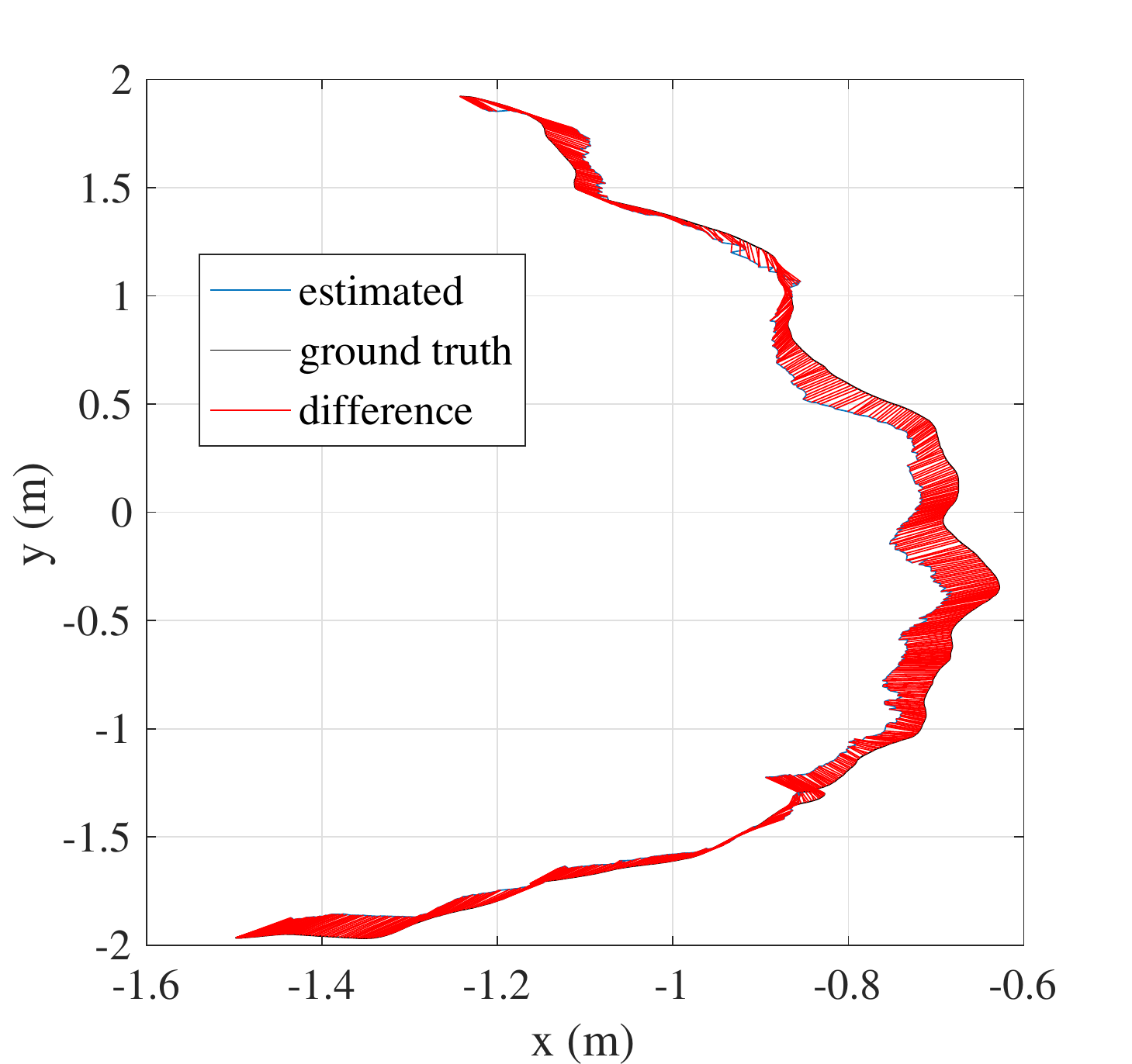} & 
		\includegraphics[scale=0.3]{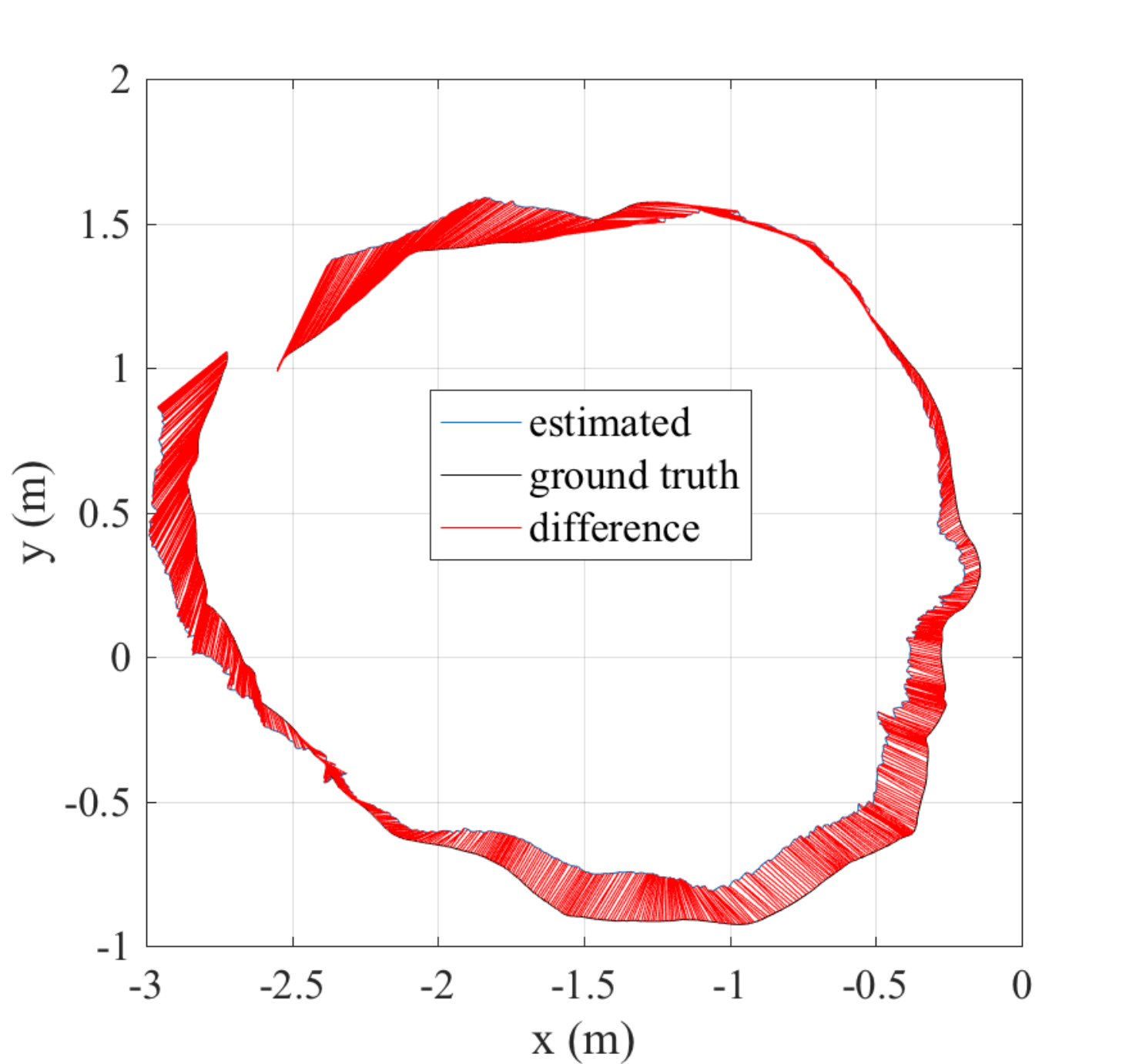} &
		\includegraphics[scale=0.3]{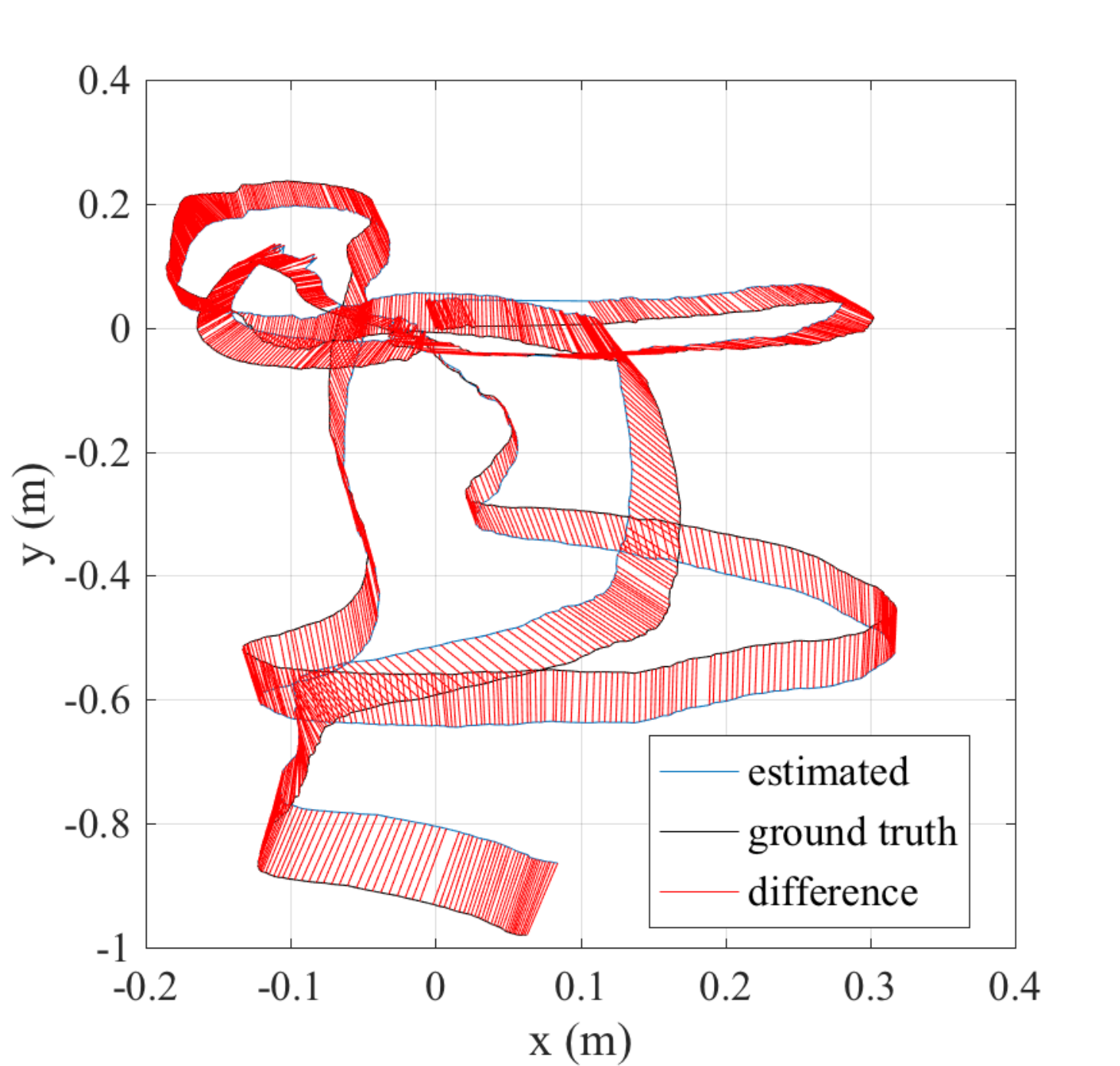} \\[-2pt]
		{\tiny kt0 (lr) w/ noise} & {\tiny  kt0 (or)}& {\tiny kt0 (or) w/ noise}\\[-4pt]
		\includegraphics[scale=0.3]{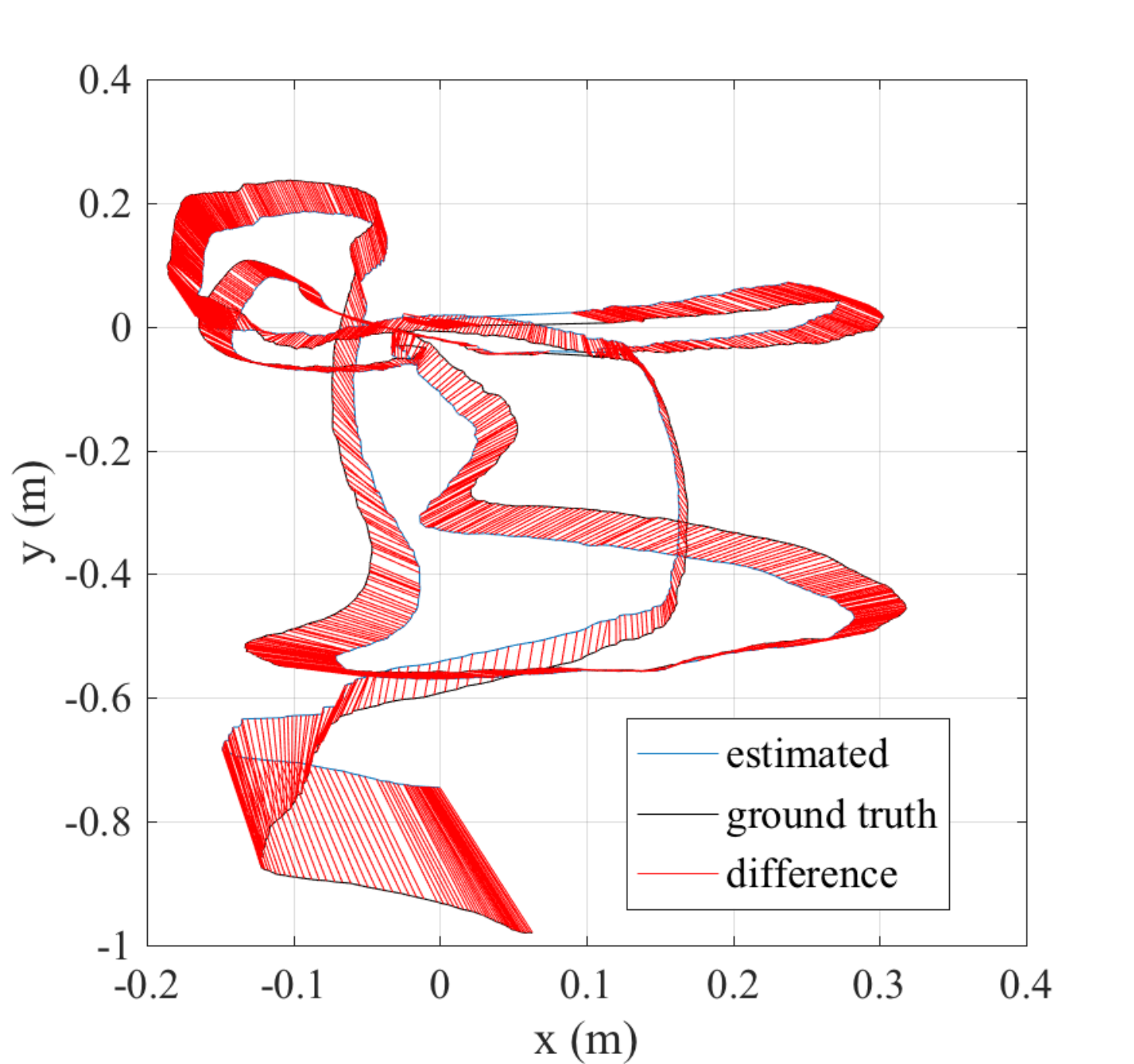} & 
		\includegraphics[scale=0.3]{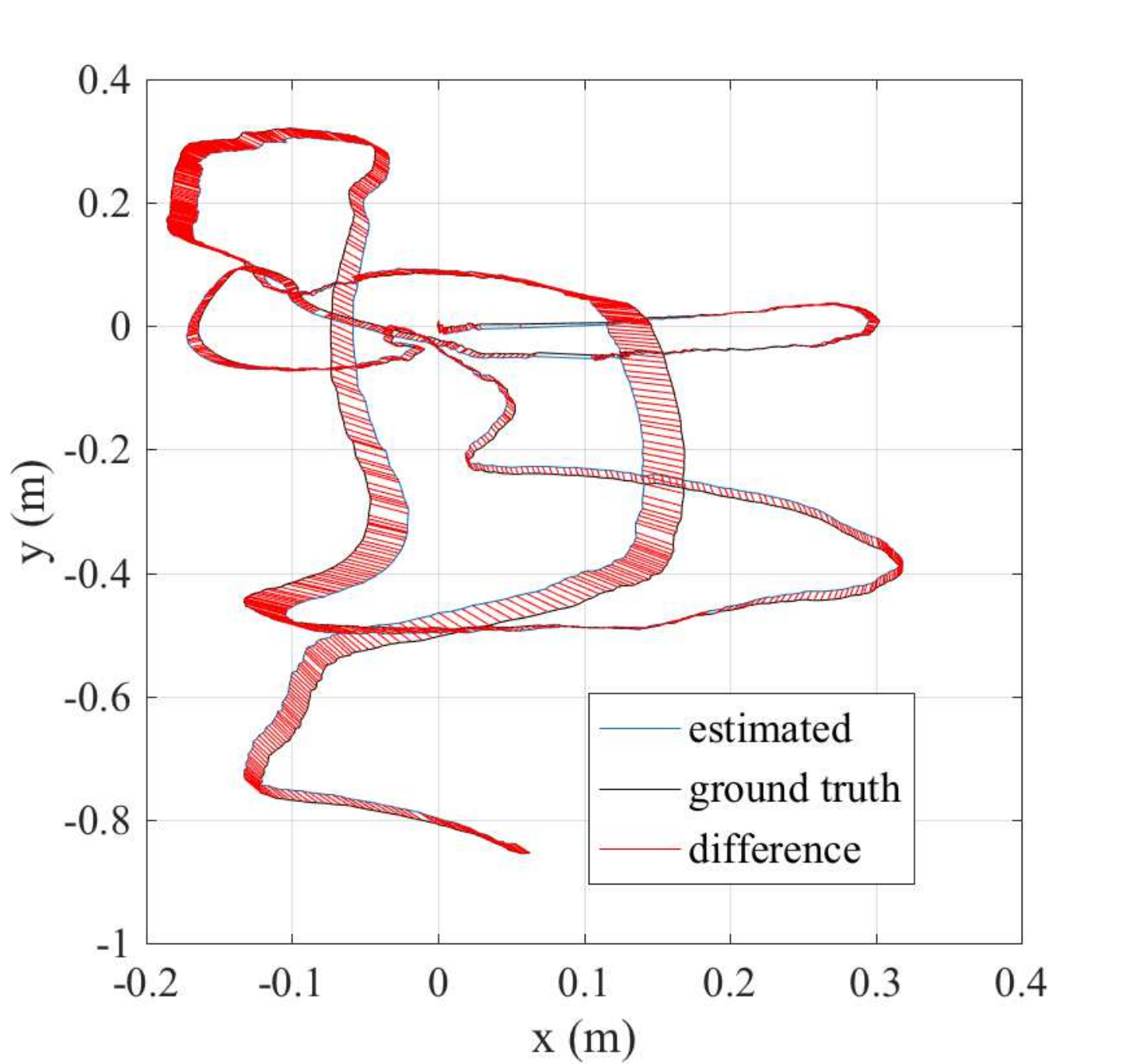} &
		\includegraphics[scale=0.3]{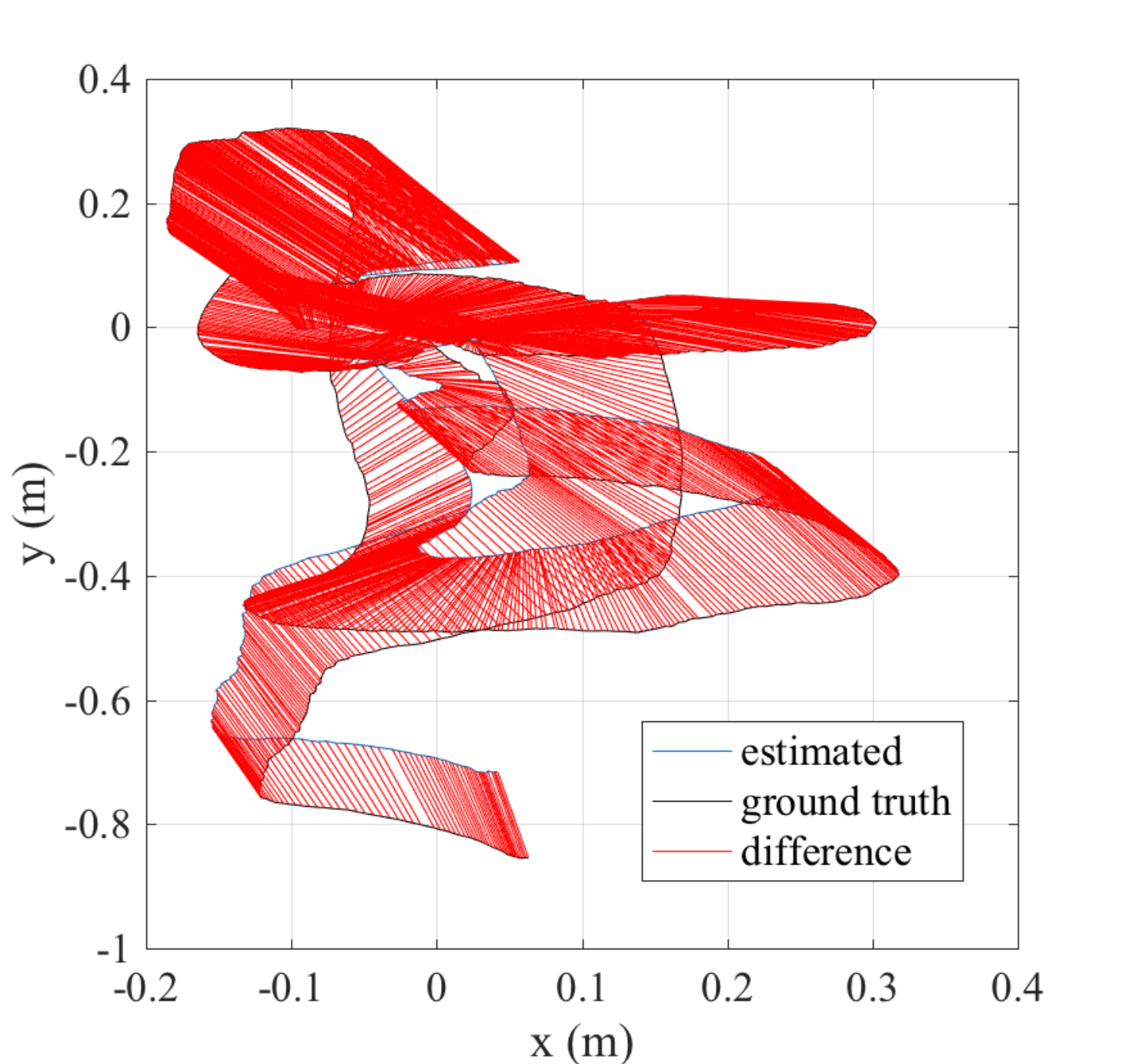} \\[-8pt]
	\end{tabular}
	\caption{Trajectories estimated by the proposed method versus the groundtruth}
	\label{fig8}
\end{figure}

For the sake of diversity, the following sequences were selected from the TUM dataset for evaluation: \textit{fr1/desk} and \textit{fr1/360} captured from a textured office; \textit{fr3/struct\_no\_text\_far} and \textit{fr3/cabinet} collected from low textured and structured scenes; \textit{fr3/walking\_static} captured from a dynamic environment with people walking; and \textit{fr2/360\_hemisphere} captured in a warehouse. The ICL-NUIM dataset contains two versions of sequences rendered from realistic models of an office and a living room, one without any noise and another with simulated RGB and depth noise, as show in Fig. \ref{fig5}. Two sequences were selected respectively: \textit{kt0 (lr)} from the living room and \textit{kt0 (or)} from the office. While the RGB noise does not seem significant, the depth noise introduced around the object boundaries, seen in Fig. \ref{fig5}, is significantly worse than the observed depth noise of real structured-light cameras (see Fig. \ref{fig11}). Furthermore, the depth map boundaries are corrupted with dense noise, thus we removed all depth values within a margin of 5 pixels. Both the relative pose error (RPE) per second and the absolute trajectory error (ATE) are reported, as RMSEs, for these sequences and the estimated trajectories are compared against the ground-truth in Fig. \ref{fig8}. \par

\par
{\renewcommand{\arraystretch}{0.9}
\begin{table}[h]
\centering
\scriptsize{
\begin{tabular}{|l|c|c|c|c|}
\hline
Features & \multicolumn{1}{c|}{Points} & \multicolumn{1}{c|}{\begin{tabular}[c]{@{}c@{}}Points \&\\  Lines\end{tabular}} &   \multicolumn{1}{c|}{\begin{tabular}[c]{@{}c@{}}Points \&\\  Planes\end{tabular}} &
All\\ \hline
fr1/desk & \begin{tabular}[c]{@{}c@{}}34 mm\\ 2.4 deg\end{tabular} & \begin{tabular}[c]{@{}c@{}}30 mm\\ 2.2 deg\end{tabular} & \begin{tabular}[c]{@{}c@{}}28 mm\\ 1.9 deg\end{tabular} &
\begin{tabular}[c]{@{}c@{}}23 mm\\ 1.7 deg\end{tabular}\\ \hline
fr1/360 & \begin{tabular}[c]{@{}c@{}}88 mm\\ 4.4 deg\end{tabular} & \begin{tabular}[c]{@{}c@{}}69 mm \\ 3.1 deg\end{tabular} & \begin{tabular}[c]{@{}c@{}}76 mm \\ 3.5 deg\end{tabular} &
\begin{tabular}[c]{@{}c@{}}64 mm \\ 2.7 deg\end{tabular} \\ \hline
\begin{tabular}[c]{@{}l@{}}fr3/struct\_no\_text\_far\end{tabular} & Fail &  \begin{tabular}[c]{@{}c@{}}32 mm \\ 0.9 deg\end{tabular} & \begin{tabular}[c]{@{}c@{}}28 mm \\ 0.9 deg\end{tabular} &
\begin{tabular}[c]{@{}c@{}}19 mm \\ 0.7 deg\end{tabular}\\ \hline
\begin{tabular}[c]{@{}l@{}}fr3/cabinet\end{tabular} & \begin{tabular}[c]{@{}c@{}}112 mm \\ 4.5 deg\end{tabular} & \begin{tabular}[c]{@{}c@{}}70 mm\\ 2.8 deg\end{tabular} & \begin{tabular}[c]{@{}c@{}}40 mm\\ 1.8 deg\end{tabular} & 
\begin{tabular}[c]{@{}c@{}}39 mm\\ 1.8 deg\end{tabular} \\ \hline
fr3/walking\_static & \begin{tabular}[c]{@{}c@{}}87 mm\\ 1.1 deg\end{tabular} & \begin{tabular}[c]{@{}c@{}}69 mm\\ 1.0 deg\end{tabular} & \begin{tabular}[c]{@{}c@{}}86 mm\\ 1.1 deg\end{tabular} & \begin{tabular}[c]{@{}c@{}}68 mm\\ 0.7 deg\end{tabular} \\ \hline
fr2/360\_hemisphere & \begin{tabular}[c]{@{}c@{}}78 mm \\ 1.4 deg\end{tabular} & \begin{tabular}[c]{@{}c@{}}72 mm\\ 1.1 deg\end{tabular} & \begin{tabular}[c]{@{}c@{}}79 mm\\ 1.7 deg\end{tabular} & 
\begin{tabular}[c]{@{}c@{}}69 mm\\ 1.1 deg\end{tabular} \\  \hline
kt0 (lr) &  \begin{tabular}[c]{@{}c@{}}8 mm\\ 0.6 deg\end{tabular} & \begin{tabular}[c]{@{}c@{}}9 mm\\ 0.6 deg\end{tabular} & \begin{tabular}[c]{@{}c@{}}8 mm \\ 0.6 deg\end{tabular} &
\begin{tabular}[c]{@{}c@{}}7 mm \\ 0.5 deg\end{tabular} 
 \\ \hline
kt0 (lr) w/ noise &  \begin{tabular}[c]{@{}c@{}}8 mm\\ 0.6 deg\end{tabular} & \begin{tabular}[c]{@{}c@{}}8 mm\\ 0.7 deg\end{tabular} & \begin{tabular}[c]{@{}c@{}}7 mm \\ 0.5 deg\end{tabular} &
\begin{tabular}[c]{@{}c@{}}6 mm \\ 0.5 deg\end{tabular}  \\ \hline
kt0 (or) & \begin{tabular}[c]{@{}c@{}}9 mm\\ 0.5 deg\end{tabular} & \begin{tabular}[c]{@{}c@{}}7 mm\\ 0.5 deg\end{tabular} & \begin{tabular}[c]{@{}c@{}}7 mm\\ 0.5 deg\end{tabular} &
\begin{tabular}[c]{@{}c@{}}7 mm\\ 0.5 deg\end{tabular} \\ \hline
kt0 (or) w/ noise & \begin{tabular}[c]{@{}c@{}}6 mm\\ 0.5 deg\end{tabular} & \begin{tabular}[c]{@{}c@{}}5 mm\\ 0.5 deg\end{tabular} & \begin{tabular}[c]{@{}c@{}}7 mm\\ 0.5 deg\end{tabular}  &\begin{tabular}[c]{@{}c@{}}6 mm\\ 0.5 deg\end{tabular}  \\ \hline
\end{tabular}}
\caption{RPE on TUM and ICL\_NUIM datasets for different combinations of geometric primitives.}
\label{tab:features_RPE}
\end{table}}

{\renewcommand{\arraystretch}{0.9}
	\begin{table}[!h]
		\centering
		\scriptsize{
			\begin{tabular}{|l|c|c|c|c|c|c|}
				\hline
				Features & \multicolumn{1}{c|}{Points} & \multicolumn{1}{c|}{\begin{tabular}[c]{@{}c@{}}Points \&\\  Lines\end{tabular}} &   \multicolumn{1}{c|}{\begin{tabular}[c]{@{}c@{}}Points \&\\  Planes\end{tabular}} &
				All\\ \hline
				fr1/desk &  64 mm & 53 mm & 50 mm & 40 mm  \\ \hline
				fr1/360 &  116 mm & 109 mm & 91 mm & 91 mm \\ \hline
				fr3/struct\_no\_text\_far & Fail & 80 mm & 63 mm & 54 mm  \\ \hline
				fr3/cabinet & 437 mm & 241 mm & 195 mm & 200 mm \\ \hline
				fr3/walking\_static & 200 mm & 181 mm & 199 mm & 179 mm \\ \hline
				fr2/360\_hemisphere & 237 mm & 238 mm & 350 mm & 203 mm\\ \hline
				kt0 (lr) & 496 mm & 446 mm & 76 mm & 99 mm  \\ \hline
				kt0 (lr) w/ noise & 428 mm & 281 mm & 83 mm & 59 mm \\ \hline
				kt0 (or) & 197 mm & 31 mm & 99 mm & 27 mm \\ \hline
				kt0 (or) w/ noise & 237 mm & 199 mm & 134 mm & 167 mm \\ \hline
		\end{tabular}}
		\caption{ATE on TUM and ICL\_NUIM datasets for different combinations of geometric primitives.}
		\label{tab:features_ATE}
\end{table}}

Table \ref{tab:features_RPE} and \ref{tab:features_ATE} shows how the performance is improved by introducing new feature-types. The performance gain of using the three geometric primitives is consistent and significant, especially in low textured environments and in the \textit{fr1/360}, where several RGB images are blurred due to sudden rotations, causing few detected feature points. \par
Table \ref{tab:depth_model} compares the performance between using the different depth models (i.e. stages) described in Section \ref{sec:depth_filter}. Overall, fusing the depth maps using the Optimal-GM framework decreases significantly the odometry error. Interestingly enough, the performance in the sequence captured in the dynamic environment is significantly improved by using the full depth filter framework, which indicates that modelling temporally the depth uncertainty helps reducing the impact of moving objects. In the ICL-NUIM captures, the introduction of simulated noise increases overall the ATE, however, the virtual camera in \textit{kt0 (lr)} faces a texture-less wall in the middle of the sequence, which causes the pose estimation to fail for a few frames, thus the ATE is affected by the employed velocity model. It is worth noting that although C-GM by itself does not seem advantageous in most sequences, when large amount of noise is present in the ICL-NUIM, we observe the contrary in terms of ATE.

{\renewcommand{\arraystretch}{0.8}
\begin{table}[t]
\centering
\scriptsize{
\begin{tabular}{|l|c|c|c|c|c|c|}
\hline
Error & \multicolumn{3}{c|}{RPE} & \multicolumn{3}{c|}{ATE} \\ \hline
Depth model & \begin{tabular}[c]{@{}c@{}}Sensor \\ model\end{tabular} & C-GM & O-GM & \begin{tabular}[c]{@{}c@{}}Sensor \\ model\end{tabular} & C-GM & O-GM \\ \hline
fr1/desk & \begin{tabular}[c]{@{}c@{}}32 mm\\ 2.3 deg\end{tabular} & \begin{tabular}[c]{@{}c@{}}33 mm\\ 2.4deg\end{tabular} & \begin{tabular}[c]{@{}c@{}}23 mm\\ 1.7 deg\end{tabular} & 60 mm & 65 mm & 40 mm \\ \hline
fr1/360 & \begin{tabular}[c]{@{}c@{}}67 mm\\ 2.8 deg\end{tabular} & \begin{tabular}[c]{@{}c@{}}66 mm \\ 2.9 deg\end{tabular} & \begin{tabular}[c]{@{}c@{}}64 mm \\ 2.7 deg\end{tabular} & 127 mm & 123 mm & 91 mm \\ \hline
\begin{tabular}[c]{@{}l@{}}fr3/struct\_no\_text\_far\end{tabular} & \begin{tabular}[c]{@{}c@{}}27 mm \\ 0.9 deg\end{tabular} & \begin{tabular}[c]{@{}c@{}}29 mm \\ 0.9 deg\end{tabular} & \begin{tabular}[c]{@{}c@{}}19 mm \\ 0.7 deg\end{tabular} & 82 mm & 95 mm & 54 mm \\ \hline
\begin{tabular}[c]{@{}l@{}}fr3/cabinet\end{tabular} &  \begin{tabular}[c]{@{}c@{}}56 mm \\ 2.3 deg\end{tabular} & \begin{tabular}[c]{@{}c@{}}62 mm\\ 2.5 deg\end{tabular} & \begin{tabular}[c]{@{}c@{}}39 mm\\ 1.8 deg\end{tabular} & 239 mm  & 275 mm & 200 mm \\ \hline
fr3/walking\_static & \begin{tabular}[c]{@{}c@{}}149 mm\\ 2.0 deg\end{tabular} &  \begin{tabular}[c]{@{}c@{}}144 mm\\ 1.9 deg\end{tabular} & \begin{tabular}[c]{@{}c@{}}68 mm\\ 0.7 deg\end{tabular} & 417 mm & 407 mm & 179 mm \\ \hline
fr2/360\_hemisphere & \begin{tabular}[c]{@{}c@{}}77 mm\\ 1.1 deg\end{tabular} & \begin{tabular}[c]{@{}c@{}}73 mm\\ 1.1 deg\end{tabular} & \begin{tabular}[c]{@{}c@{}}69 mm\\ 1.1 deg\end{tabular} & 198 mm & 193 mm & 203 mm \\ \hline
kt0 (lr) & \begin{tabular}[c]{@{}c@{}}7 mm \\ 0.5 deg\end{tabular} & \begin{tabular}[c]{@{}c@{}}7 mm \\ 0.5 deg\end{tabular} & \begin{tabular}[c]{@{}c@{}}7 mm \\ 0.5 deg\end{tabular} & 198 mm & 97 mm & 99 mm \\ \hline
kt0 (lr) w/ noise & \begin{tabular}[c]{@{}c@{}}6 mm \\ 0.6 deg\end{tabular} & \begin{tabular}[c]{@{}c@{}}6 mm \\ 0.5 deg\end{tabular} & \begin{tabular}[c]{@{}c@{}}6 mm \\ 0.5 deg\end{tabular} & 303 mm & 113 mm & 59 mm \\ \hline
kt0 (or) & \begin{tabular}[c]{@{}c@{}}7 mm \\ 0.5 deg\end{tabular} & \begin{tabular}[c]{@{}c@{}}7 mm \\ 0.5 deg\end{tabular} & \begin{tabular}[c]{@{}c@{}}7 mm\\ 0.5 deg\end{tabular} & 25 mm & 36 mm & 27 mm \\ \hline
kt0 (or) w/ noise & \begin{tabular}[c]{@{}c@{}}5 mm \\ 0.5 deg\end{tabular} & \begin{tabular}[c]{@{}c@{}}6 mm \\ 0.5 deg\end{tabular} & \begin{tabular}[c]{@{}c@{}}6 mm\\ 0.5 deg\end{tabular} & 359 mm & 220 mm & 167 mm \\ \hline
\end{tabular}}
\caption{RMSE on TUM and ICL\_NUIM datasets for different depth uncertainty models. As described in Section \ref{sec:depth_filter}, the sensor model corresponds to the first stage of the proposed filter method, the C-GM uses the two first stages and the O-GM uses the full depth fusion method.}
\label{tab:depth_model}
\end{table}}

\begin{table}[b]
	\centering
	\scriptsize{
		\begin{tabular}{|l|c|c|c|c|}
			\hline
			& \multicolumn{2}{c|}{Ours} & \multicolumn{2}{c|}{State-of-the-art (VO)} \\ \hline
			Error & RPE & ATE & RPE & ATE \\ \hline
			fr1/desk & 23 mm & 40 mm & 25 mm \cite{gutierrez2016dense} & 32 mm \cite{gutierrez2016dense} \\ \hline
			fr1/360 & 64 mm & 91 mm & 73 mm \cite{proenca2017planes} & - \\ \hline
			fr3/struct\_no\_text & 19 mm & 54 mm & 43 mm \cite{yang2017direct} & 19 mm \cite{wangedge} \\ \hline
			fr3/cabinet & 39 mm & 200 mm & 80 mm \cite{proenca2017planes} & 268 mm \cite{wangedge} \\ \hline
			fr3/walking\_static & 68 mm & 179 mm & 111 mm \cite{VOSFdynamic}& - \\ \hline
			fr2/360\_hemisphere & 69 mm & 203 mm & 66 mm \cite{SPLODE2017pro} & - \\ \hline
	\end{tabular}}
	\caption{Comparison of visual odometry methods on TUM dataset.}
	\label{tab:soa_tum}
\end{table}

Our method is compared to state-of-the-art visual odometry methods in Tables \ref{tab:soa_tum} and \ref{tab:soa_icl_nuim} for each respective dataset. For the sake of fairness, our comparison does not include full SLAM systems that perform map optimization or loop closure detection. In terms of RPE, our method achieves state-of-the-art results in the TUM dataset. \par

\begin{table}[t]
	\centering
	\scriptsize{
		\begin{tabular}{|l|c|c|c|c|}
			\hline
			& Ours & DVO & FOVIS & \cite{gutierrez2016dense} \\ \hline
			kt0 (lr)  & 99 mm & 114 mm & 1931 mm & 10 mm \\ \hline
			kt0 (lr) w/ noise & 59 mm & 291 mm & 2051 mm & 6 mm \\ \hline
			kt0 (or) & 27 mm & 398 mm & 3396 mm & 4 mm \\ \hline
			kt0 (or) w/ noise & 167 mm & 335 mm & 3296 mm & 15 mm \\ \hline
	\end{tabular}}
	\caption{Comparison of absolute trajectory errors obtained by several visual odometry methods on ICL-NUIM dataset, according to the results published in \cite{iclnuim} and \cite{gutierrez2016dense}. For \cite{gutierrez2016dense}, we report the results for the best overall visibility ratio threshold used in the keyframe selection.}
	\label{tab:soa_icl_nuim}
\end{table}

Notably, we outperform the point and line odometry methods \cite{yang2017direct,PLVO}, and the recently proposed \cite{VOSFdynamic}, which addresses explicitly dynamic environments. In the ICL-NUIM dataset, our method is compared against: the DVO \cite{DVO2013}, which minimizes densely the photometric error; the feature point-based FOVIS \cite{FOVIS}; and the method proposed in \cite{gutierrez2016dense}, which is essentially an improved version of the DVO that minimizes also the geometric error by adopting an inverse depth parameterization and performs keyframe-to-frame alignment. Although, our method outperforms DVO and FOVIS, it performs significantly worse than the remarkable performance published in \cite{gutierrez2016dense}. Nevertheless, the frame-to-frame version of this method, shown in \cite{gutierrez2016dense}, compares well with our results, suggesting that this observed discrepancy is partially due the use of a keyframe-to-frame strategy, which is known to be a good way to avoid the accumulation of pose errors.

\subsection{Author-collected dataset}
\label{sec:author_dataset}

\begin{figure}
	\centering
	\includegraphics[scale=0.08]{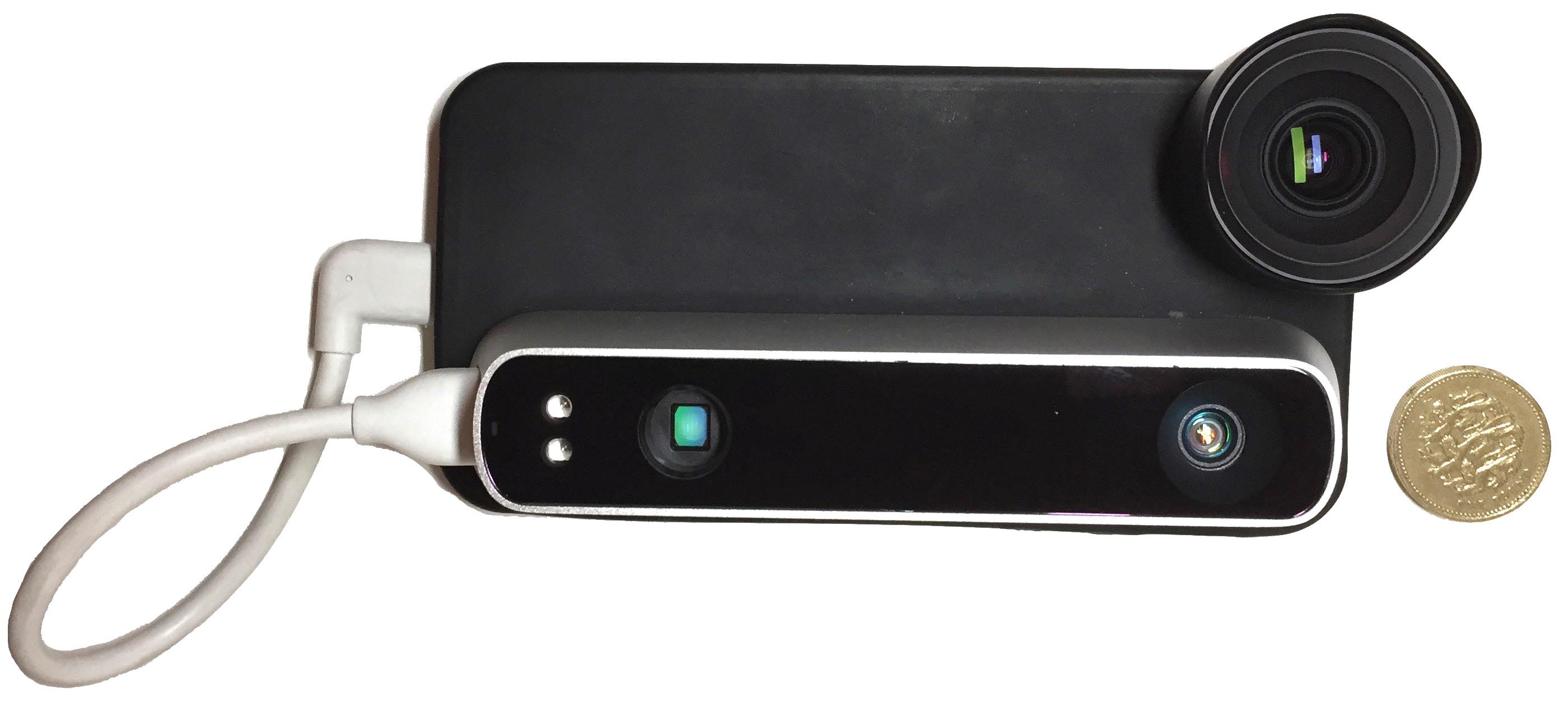}
	\caption{Mobile RGB-D capture setup used in this work. An Occipital Structure sensor is used to collect depth while RGB is collected by a wide-angle lens (Moment wide lens) mounted on an iPhone camera. The effective angle-of-views are respectively $85\degree \times70\degree$ and $58\degree\times45\degree$ for the color and the infrared camera and the baseline between them is around  38 mm. Both cameras operate at 30 fps with VGA resolution.}
	\label{fig9}
\end{figure}

Here, we evaluate the visual odometry method on four closed-loop trajectories, recorded, whilst walking, by the hand-held RGB-D setup shown in Fig. \ref{fig9}.  The employed depth sensor is also based on structured-light and follows, along with Kinect 1, the same design as Primesense cameras \cite{zanuttigh2016time} with an equal projector-camera baseline of 75 mm, thus we used the same depth sensor noise model.  As shown in Fig. \ref{fig11}, all sequences were collected in structured environments, where low textured surfaces are predominant. To avoid degenerate scene configurations due to the lack of textures and depth information, a wide-angle lens was mounted on the color camera to expand the FOV and the depth fusion was used to recover depth values outside the FOV of the depth camera. 

The drawback of expanding the FOV is that the pixel resolution is reduced and consequently mapping the depth measurements to the color image downsamples significantly the depth map, since the FOV of the RGB camera is more than the double of the depth camera FOV. As this results in the projection multiple depth measurements to a pixel, analogous to temporal depth fusion, we used, here, the O-GM method in place of the C-GM, at the second stage, to obtain simultaneously the aligned depth image and its uncertainty. \par
{\renewcommand{\arraystretch}{0.8}
\begin{table}[h]
	\centering
	\scriptsize{
		\begin{tabular}{|l|c|c|c|c|}
			\hline
			Sequence (distance) & Points & Points \& Lines & Points \& Planes & All \\ \hline
			lab (51 m) &  \begin{tabular}[c]{@{}c@{}}2.3 m \\ 20 deg\end{tabular} &  \begin{tabular}[c]{@{}c@{}}1.1 m \\  12 deg\end{tabular} & \begin{tabular}[c]{@{}c@{}}2.4 m \\  19 deg \end{tabular} & \begin{tabular}[c]{@{}c@{}}1.2 m \\  12 deg\end{tabular}\\ \hline
			parking garage 1 (56 m)  & \begin{tabular}[c]{@{}c@{}}1.6 m \\  17 deg\end{tabular} & \begin{tabular}[c]{@{}c@{}}1.9 m \\  23 deg\end{tabular} & \begin{tabular}[c]{@{}c@{}}4.2 m \\  14 deg\end{tabular} &  \begin{tabular}[c]{@{}c@{}}1.3 m \\  18 deg\end{tabular} \\ \hline
			parking garage 2 (53 m)  & \begin{tabular}[c]{@{}c@{}}7.9 m \\  51 deg\end{tabular} & \begin{tabular}[c]{@{}c@{}}1.7 m \\  10 deg\end{tabular} &  \begin{tabular}[c]{@{}c@{}}3.0 m \\  18 deg\end{tabular} & \begin{tabular}[c]{@{}c@{}}0.8 m \\  7 deg\end{tabular} \\ \hline
			corridor (110 m)  &  \begin{tabular}[c]{@{}c@{}}23 m \\  47 deg\end{tabular} & \begin{tabular}[c]{@{}c@{}}5.7 m \\  26 deg\end{tabular} &  \begin{tabular}[c]{@{}c@{}}13.8 m \\  37 deg\end{tabular}& \begin{tabular}[c]{@{}c@{}}3.6 m \\  25 deg\end{tabular} \\ \hline
	\end{tabular}}
	\caption{Final trajectory errors for the RGB-D sequences collected in this work.}
	\label{tab:wideangledataset}
\end{table}}
\begin{figure}[!h]
	\centering
	\begin{tabular}{@{}c@{}c@{}}
		\includegraphics[scale=0.4]{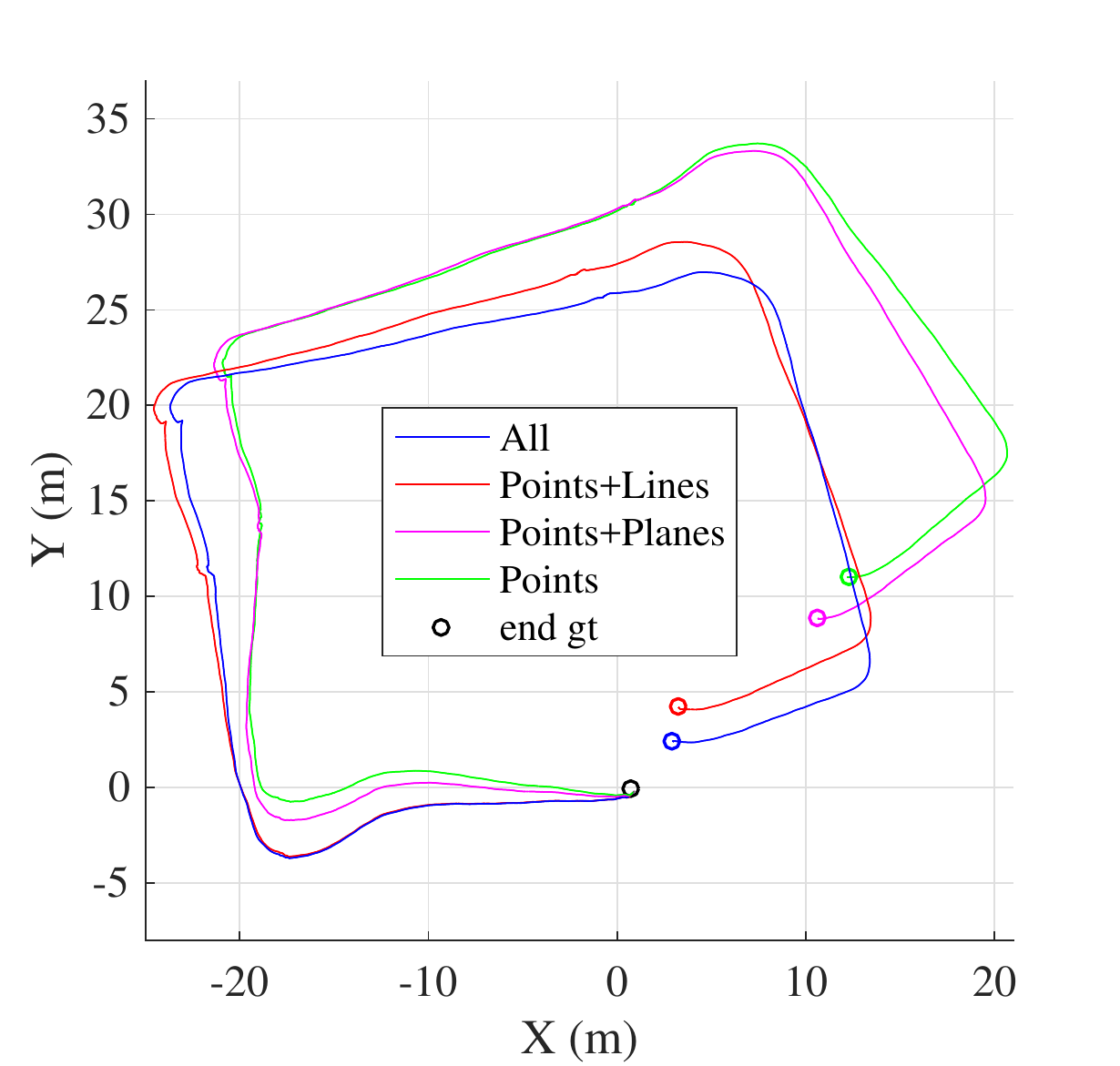} & 
		\includegraphics[scale=0.4]{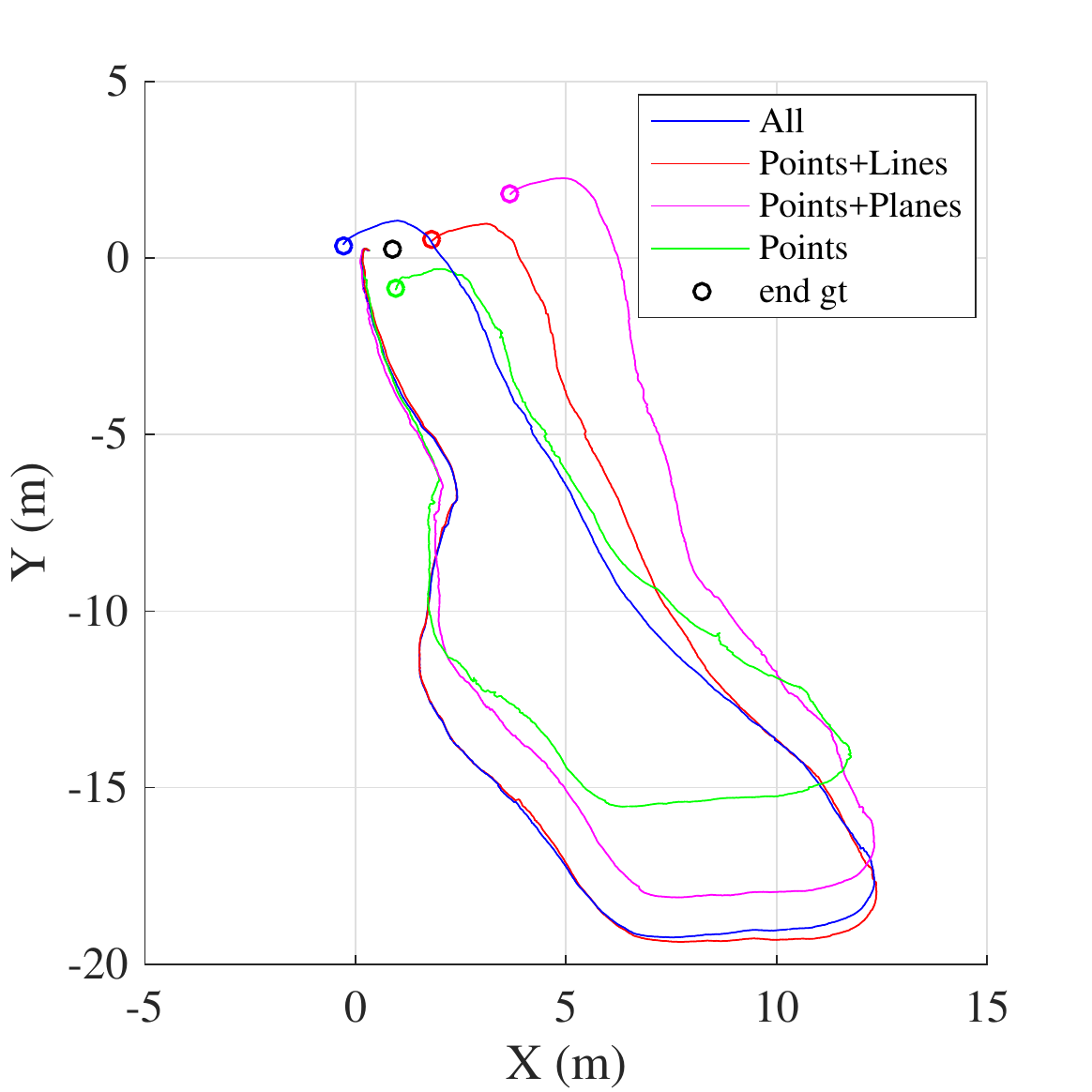} \\[-6pt]
		\includegraphics[scale=0.4]{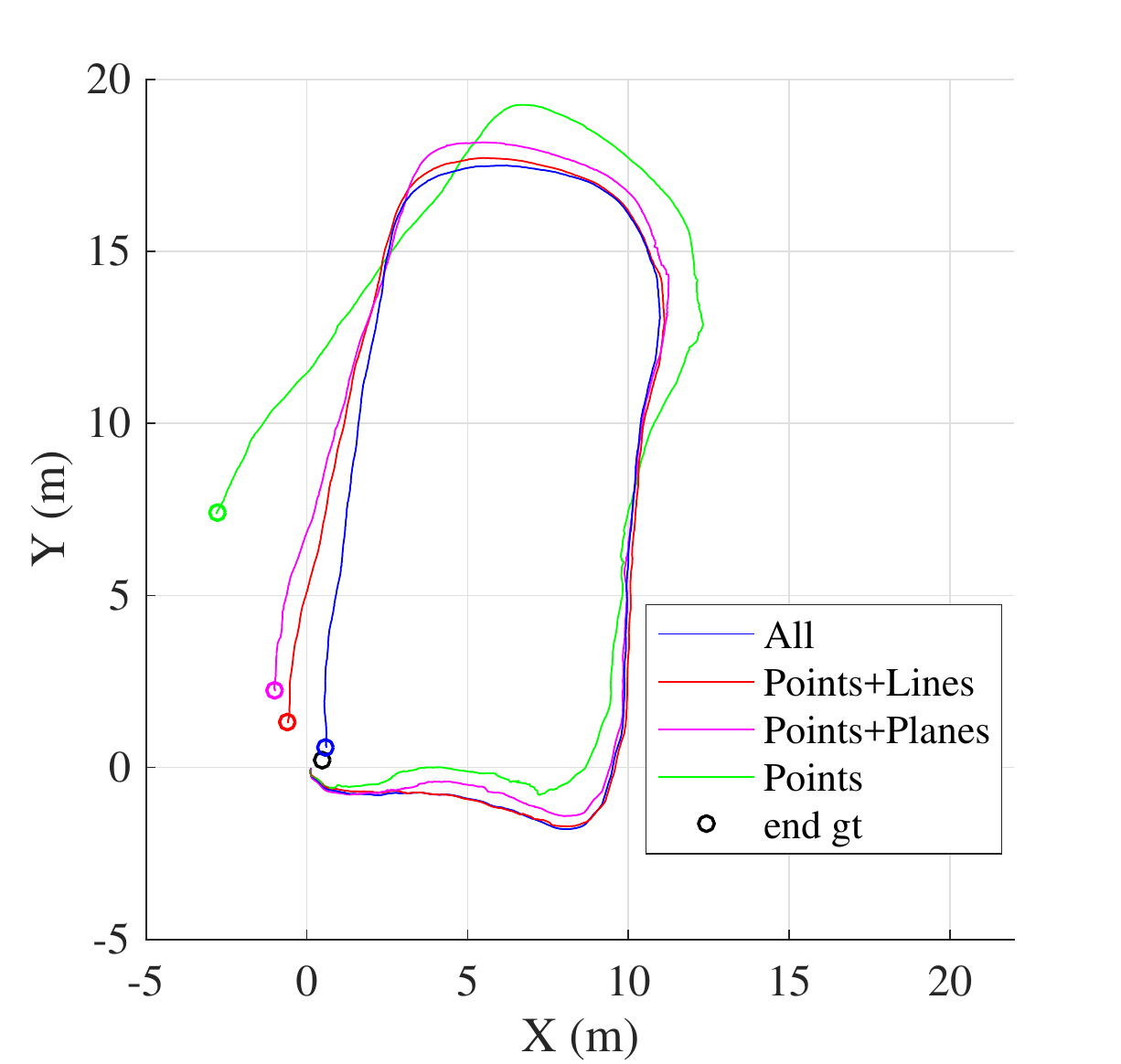} &
		\includegraphics[scale=0.4]{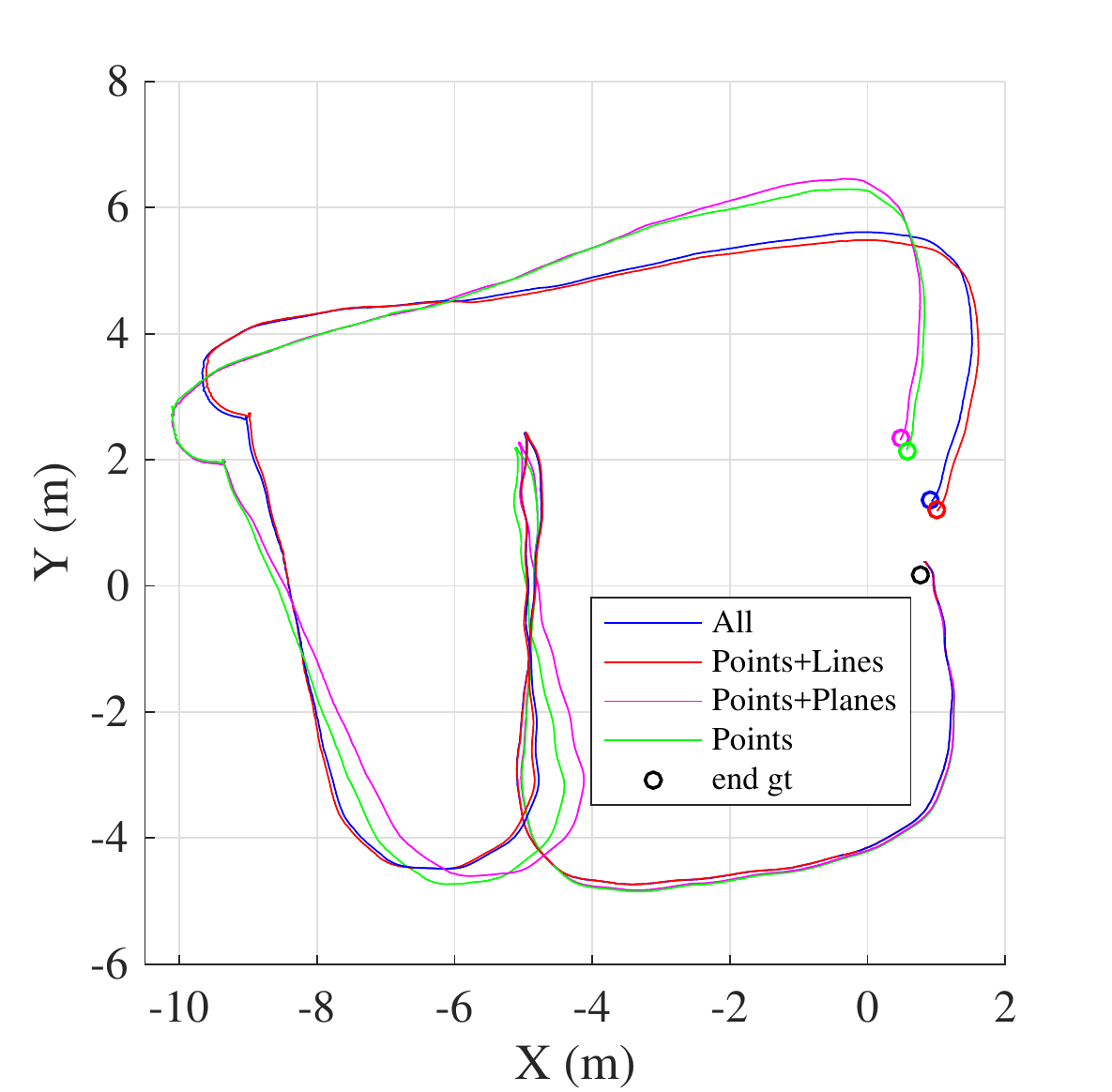} \\[-5pt]
	\end{tabular}
	\caption{Top view of the trajectories estimated by the proposed method for different feature combinations versus the final position groundtruth (end gt). Clockwise from top left: corridor, parking garage 1, lab and parking garage 2.}
	\label{fig10}
\end{figure}
Table \ref{tab:wideangledataset} reports the final trajectory error for different feature-type combinations, while Fig. \ref{fig10} shows the respective estimated trajectories. The translational and angular error were measured in these closed-loop trajectories by using a marker \cite{artoolkitplus} and expressing the angular error through the angle-axis representation. The combination of points, lines and planes shows consistently better results than the other versions, whereas the point odometry yields poor results and loses tracking for several frames. Although the final error in the \textit{parking garage 1} indicates that using just points is better than combining points and lines, the estimated trajectory, shown in Fig. \ref{fig10}, is coarser when using just points. Additionally, we observed that both versions fail tracking in this sequence, thus switch temporally the pose estimation to the velocity model, contrary to the full combination.

\begin{figure}
	\centering
	\includegraphics[scale=0.34]{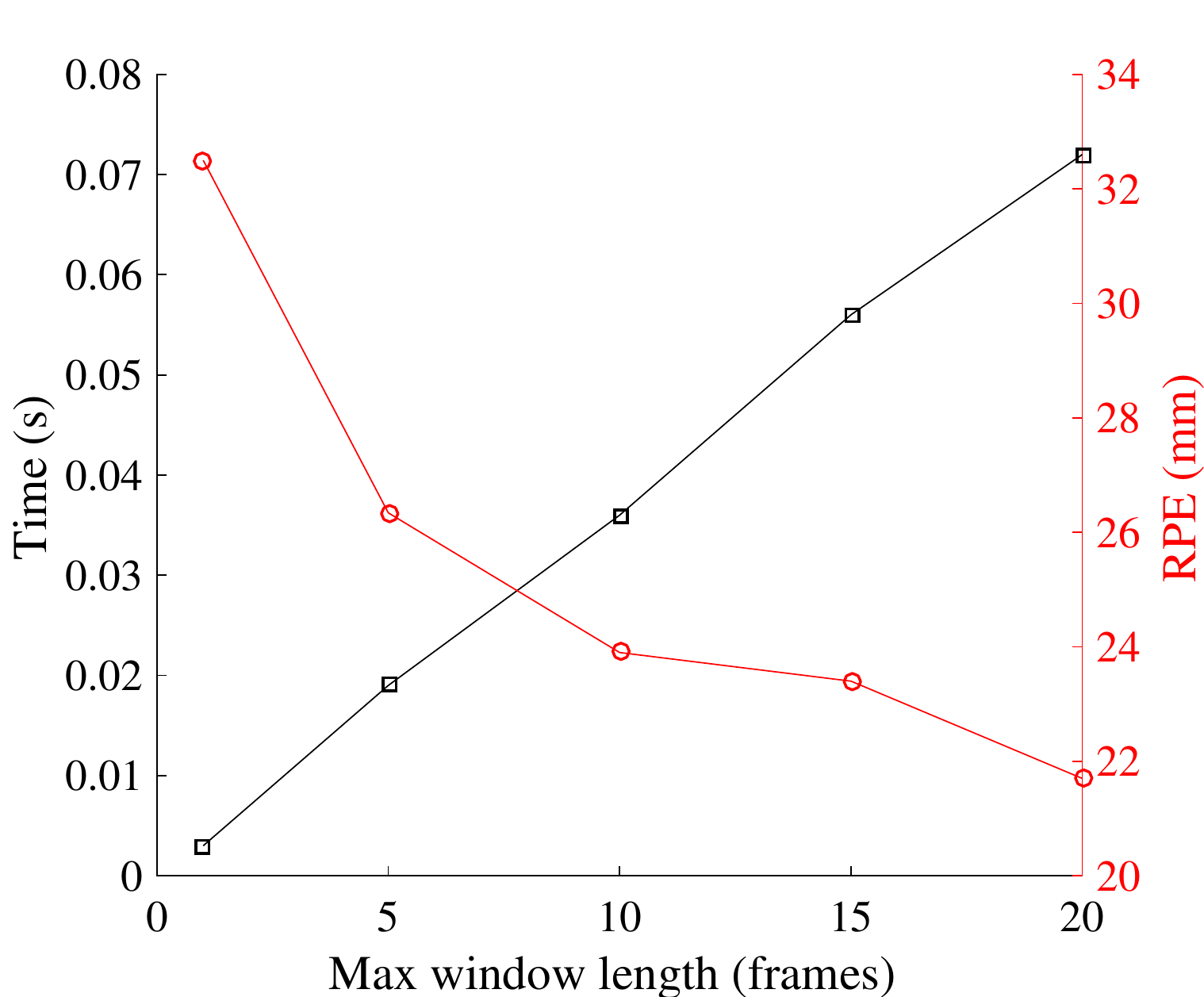}
	\caption{Maximum number of frames used for temporal depth fusion vs. O-GM runtime (black) and RPE (red) on the \textit{fr1/desk} sequence.}
	\label{fig12}
\end{figure}

\subsection{Processing Time}
\label{sec:processing_time}
All sequences were processed offline with a single thread on an Intel Core i5-6500 CPU 3.20 GHz. The method was implemented on MATLAB with C++ mex functions for certain modules as indicated in Table \ref{tab:runtime}. There are many opportunities for optimization: As shown in Fig. \ref{fig6}, point, line and plane processing (i.e. detection, extraction and matching) can be parallelized in three threads. However the plane extraction must be called once more after the depth fusion. As shown in \cite{feng2014fast}, the plane extraction can be speeded-up significantly by avoiding the per-pixel refinement and using a coarser graph. In terms of feature points, faster alternatives to SURF are well known \cite{calonder2010brief}, whereas for detection of line segments, unfortunately to our knowledge, there is a lack of good alternatives, thus one can either do line detection at half resolution (QVGA) or adopt a line tracking approach as in \cite{PLSVO}. \par
As shown in  Fig. \ref{fig12}, increasing the size of the sliding window for the temporal fusion beyond 10 frames can 
improve even further the RPE performance, however the
the cost of depth fusion grows linearly with the number of frames.

\begin{table}[tb]
\centering
\scriptsize{
\begin{tabular}{|l|l|}
\hline
                           & Time   \\ \hline
O-GM$_{10}$                 & 35 ms  \\ \hline
2D point processing        & 31 ms  \\ \hline
2D line processing         & 40 ms  \\ \hline
3D line RANSAC sampling   & 2.3 ms \\ \hline
3D line Fitting (per line) $\dagger$ & 0.1 ms \\ \hline
Plane extraction           & 29 ms  \\ \hline
Plane fitting (per plane) $\dagger$ & 1.5 ms \\ \hline
Plane matching $\dagger$   & 3 ms   \\ \hline
Pose estimation $\dagger$  & 33 ms  \\ \hline
\end{tabular}}
\caption{Timing average results on \textit{fr1/desk}. The three stages of the depth filter were timed both for a sliding window of 5 frames, as O-GM$_5$, and 10 frames, as O-GM$_{10}$, used in our experiments. Both the 2D point and line processing include feature detection, extraction and matching. The processes marked with a $\dagger$ are implemented on MATLAB, while the rest is implemented on C++ and integrated through mex functions.}
\label{tab:runtime}
\end{table}

\begin{figure}[h]
	\centering
	\begin{tabular}{@{}c@{\hspace{-2pt}}c@{\hspace{-2pt}}c@{\hspace{-2pt}}c@{}}
		\includegraphics[scale=0.15]{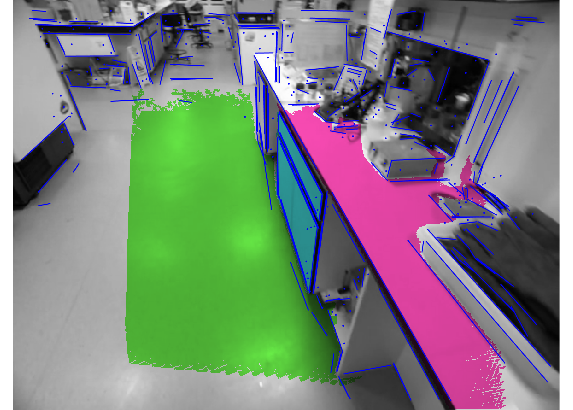} & 
		\includegraphics[scale=0.15]{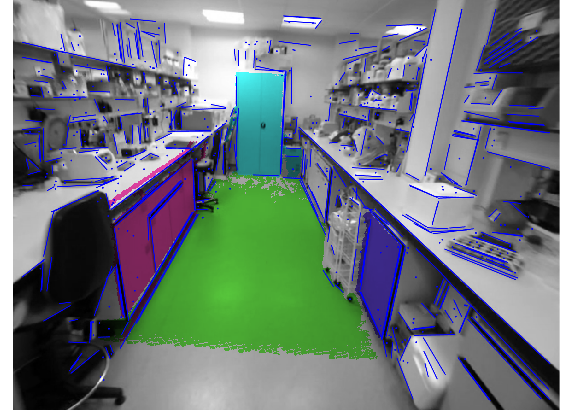} &
		\includegraphics[scale=0.15]{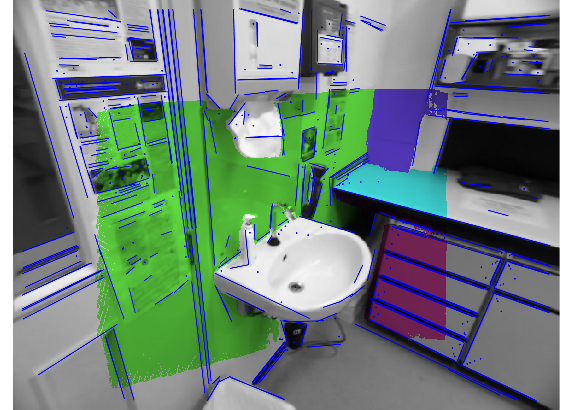} & 
		\includegraphics[scale=0.15]{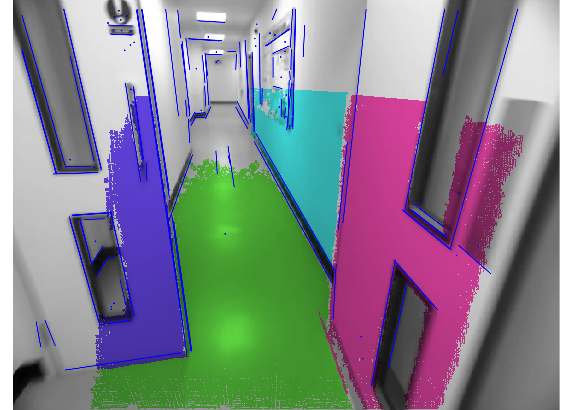} \\[-3pt]
		\includegraphics[scale=0.15]{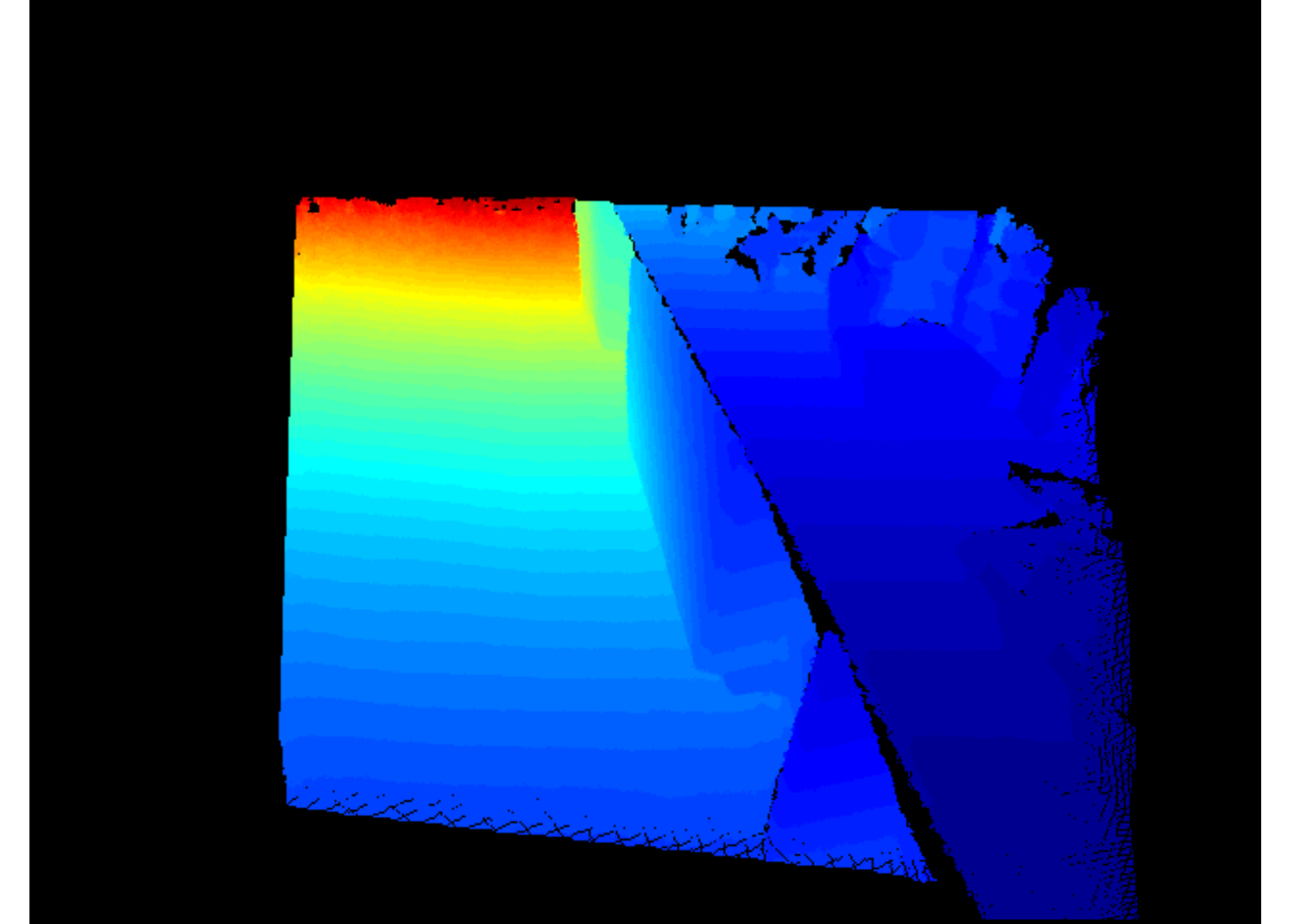} & 
		\includegraphics[scale=0.15]{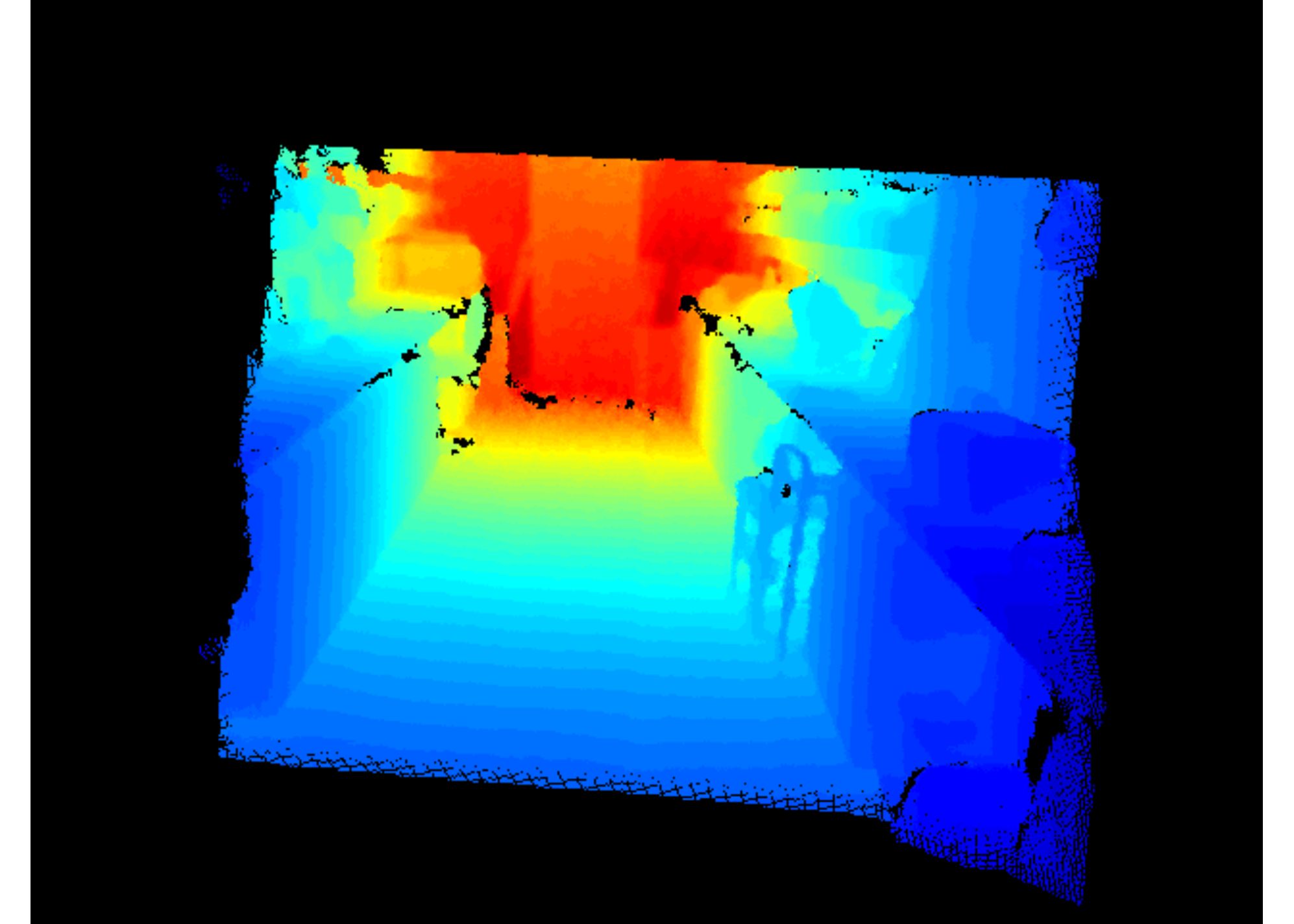} &
		\includegraphics[scale=0.15]{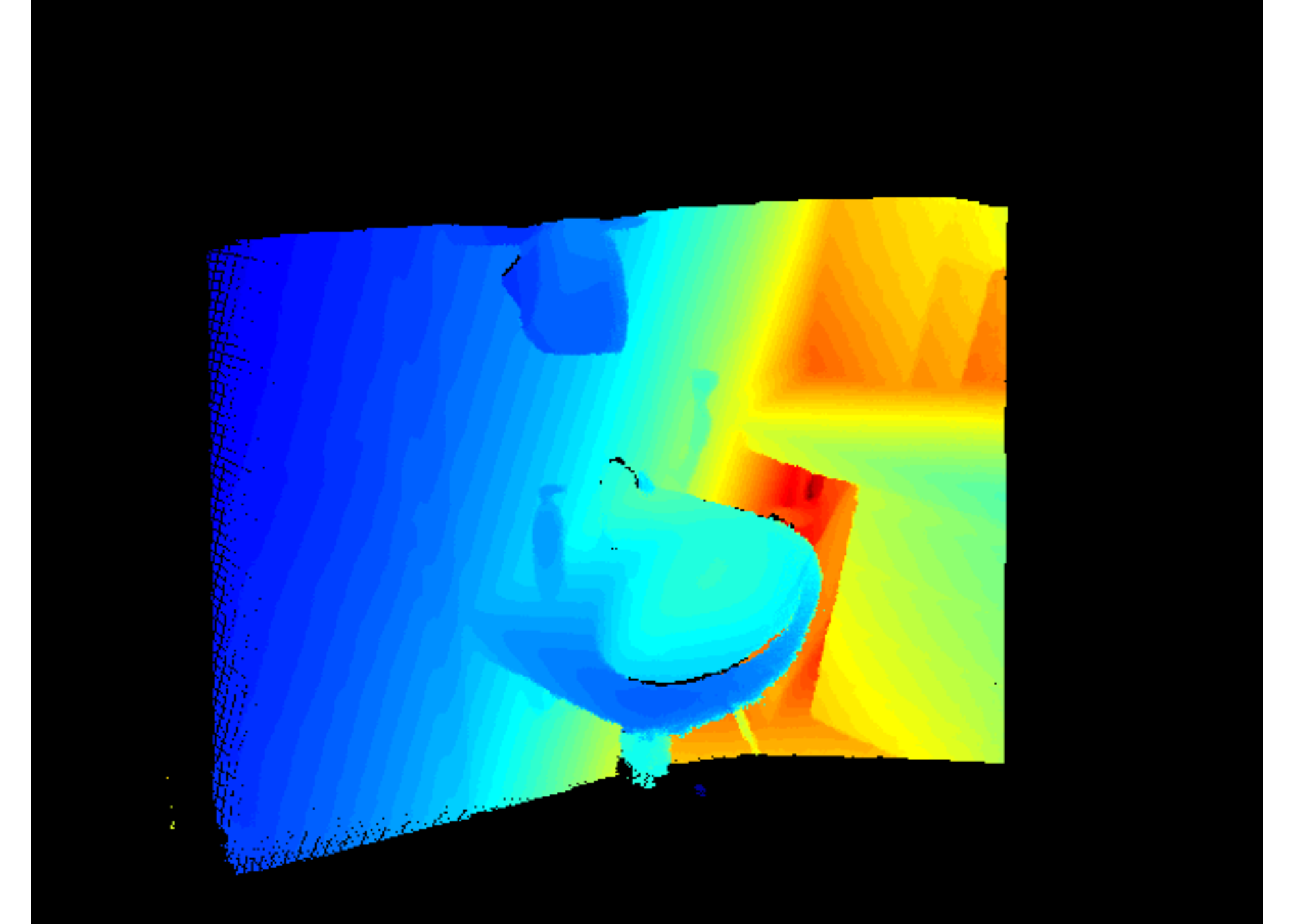} & 
		\includegraphics[scale=0.15]{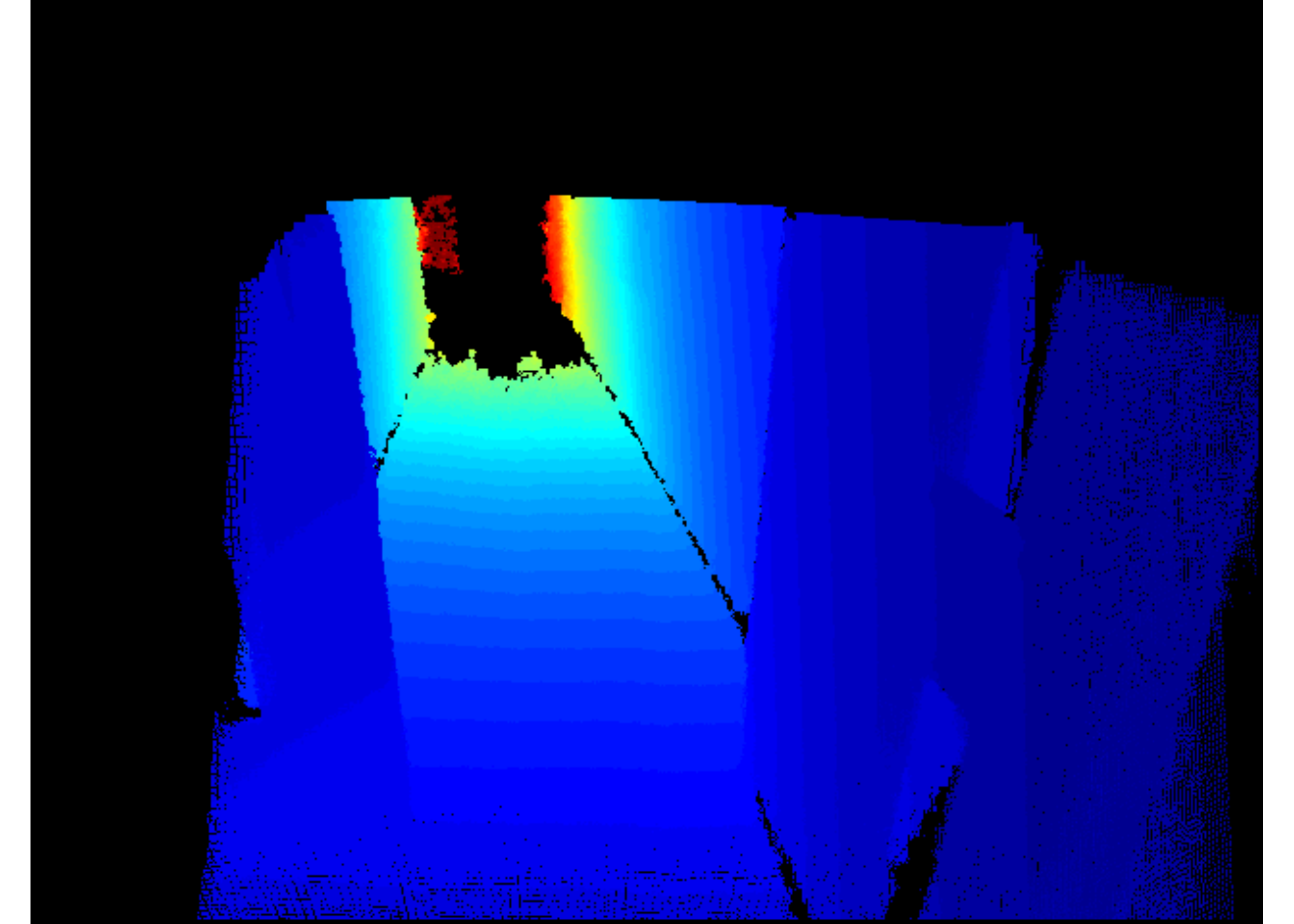} \\[-5pt]
		\multicolumn{4}{c}{ {\scriptsize Lab}} \\
		\includegraphics[scale=0.15]{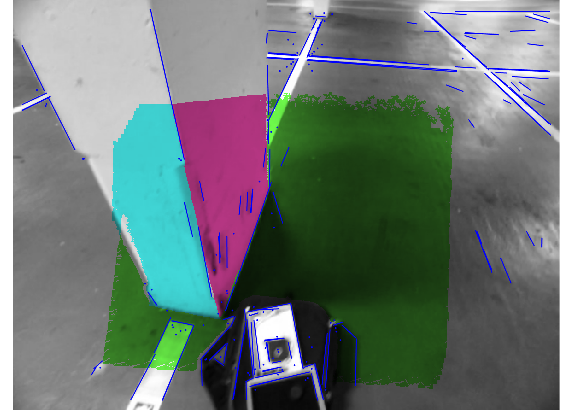} & 
		\includegraphics[scale=0.15]{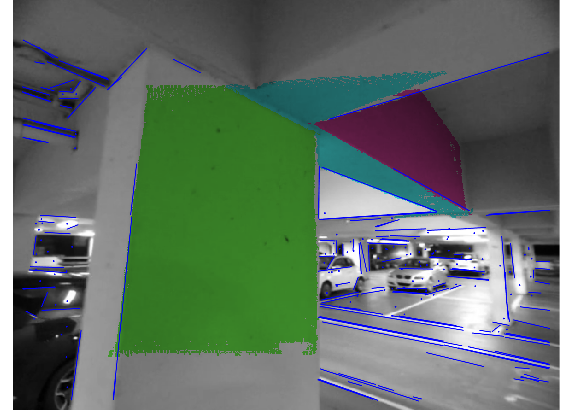} &
		\includegraphics[scale=0.15]{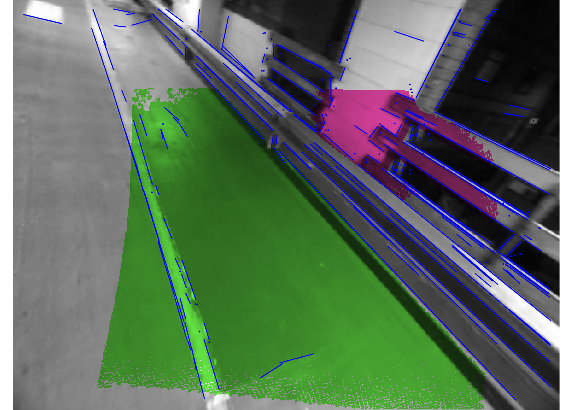} & 
		\includegraphics[scale=0.15]{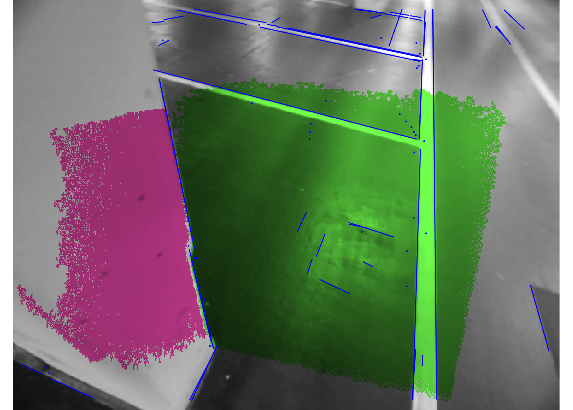} \\[-3pt]
		\includegraphics[scale=0.15]{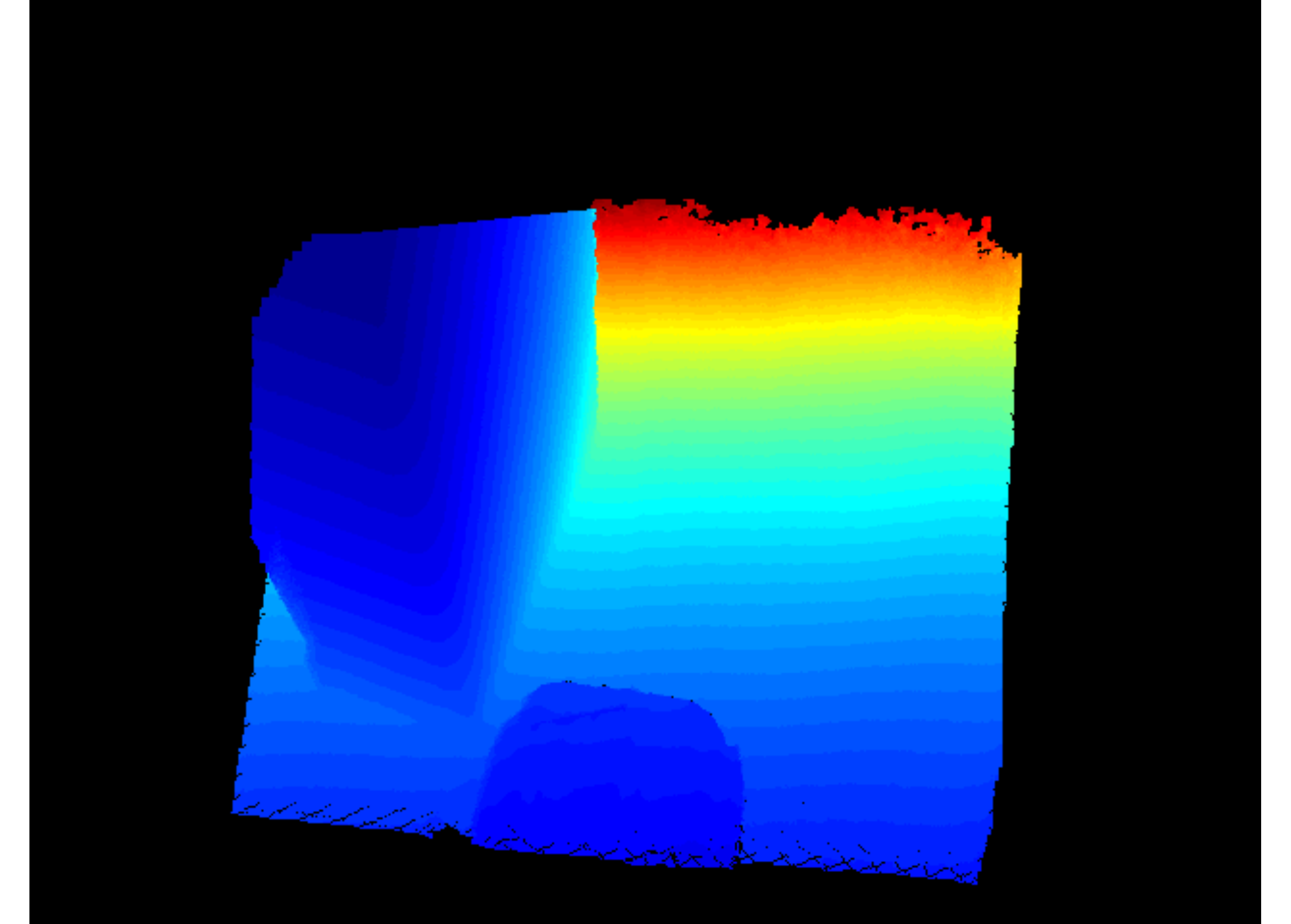} & 
		\includegraphics[scale=0.15]{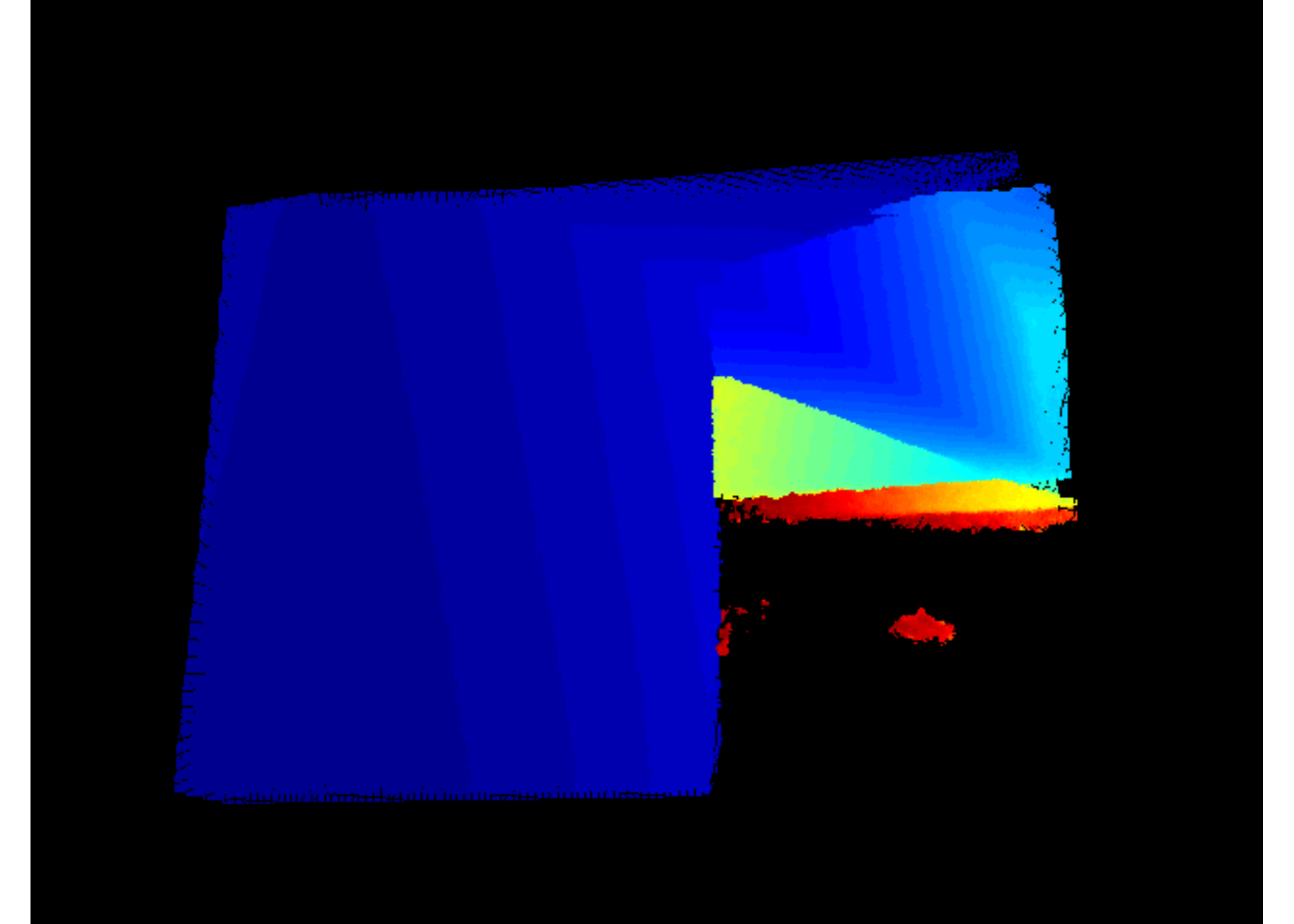} &
		\includegraphics[scale=0.15]{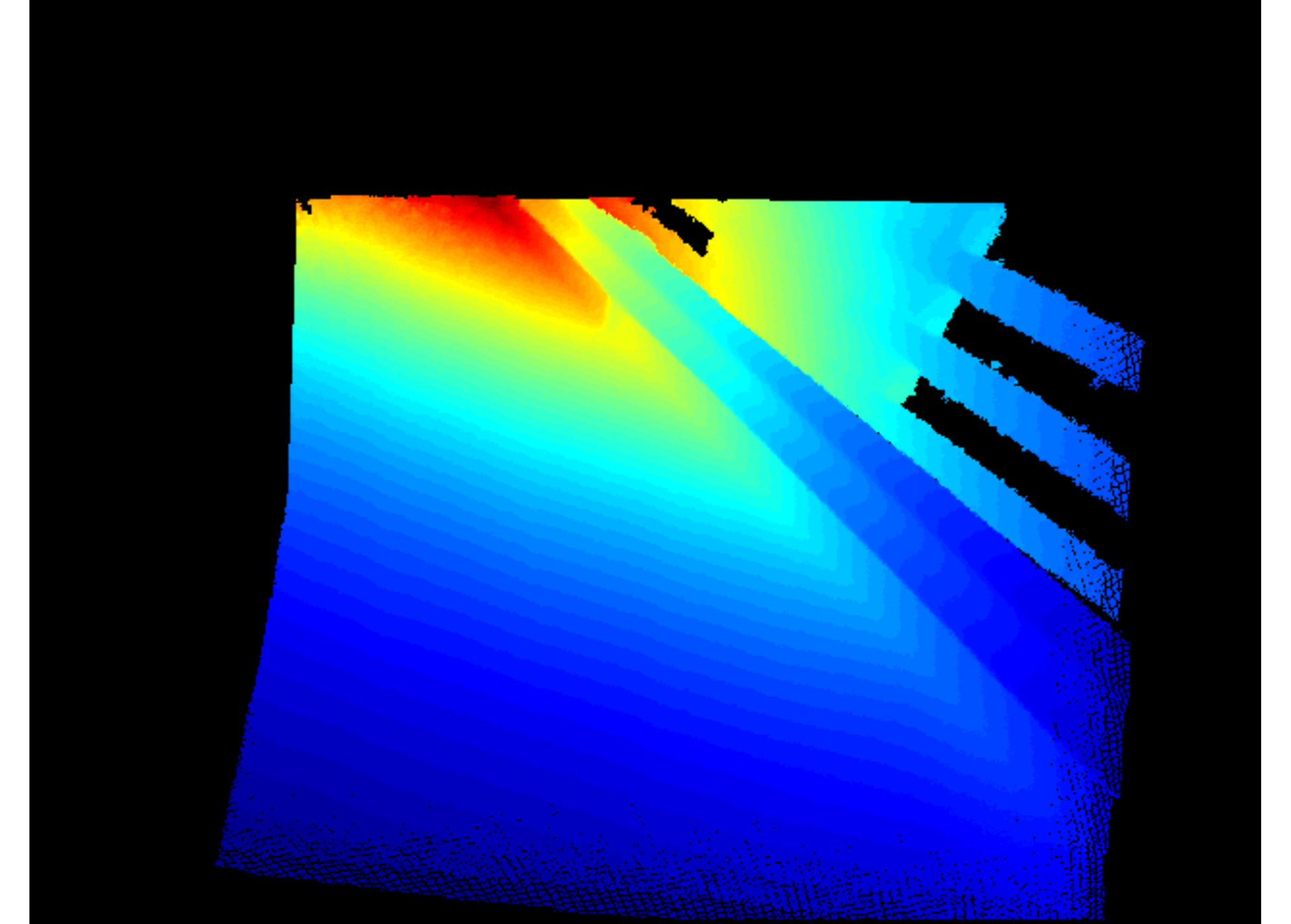} & 
		\includegraphics[scale=0.15]{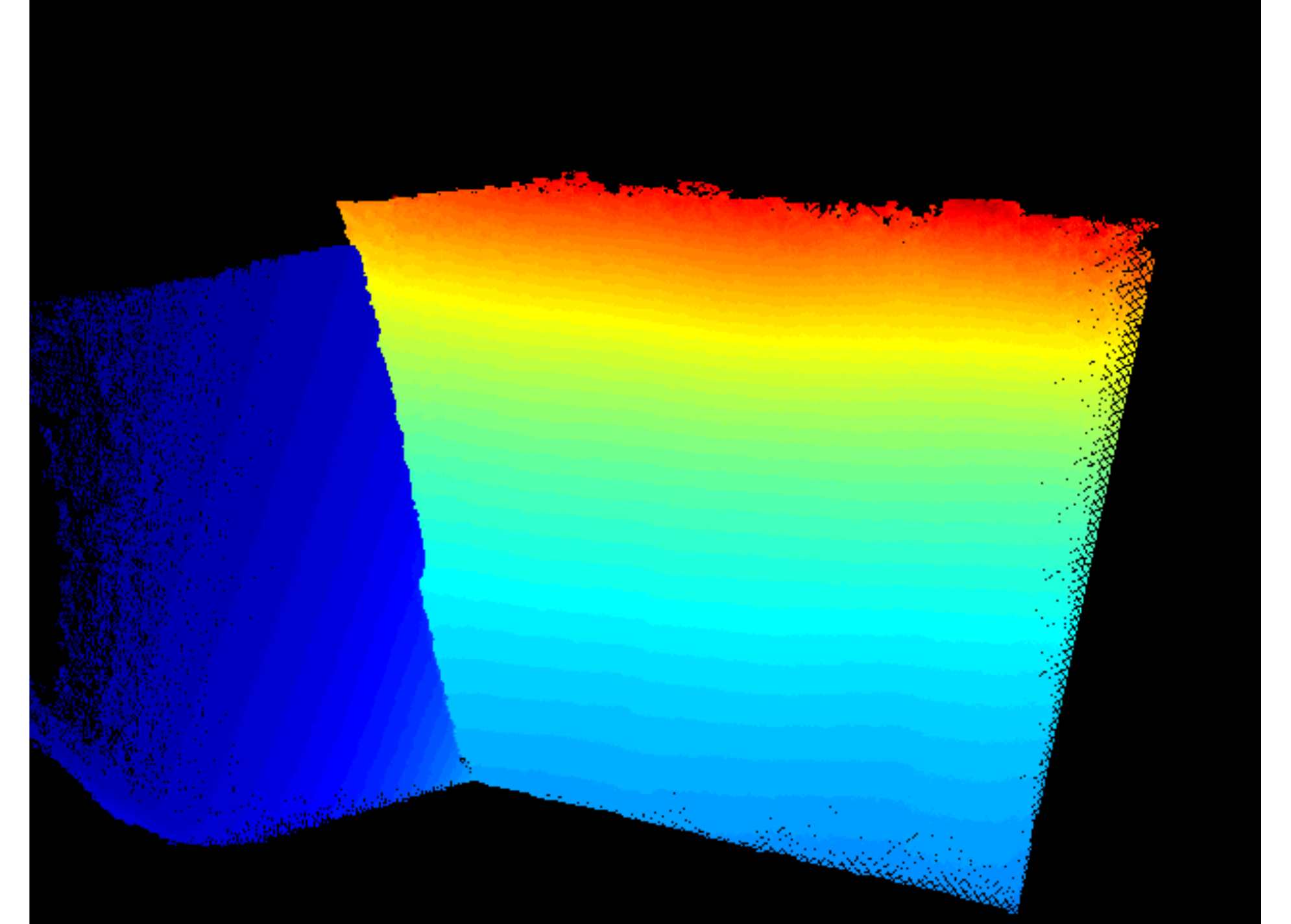} \\[-5pt]
		\multicolumn{4}{c}{ {\scriptsize Parking garage}} \\
		\includegraphics[scale=0.15]{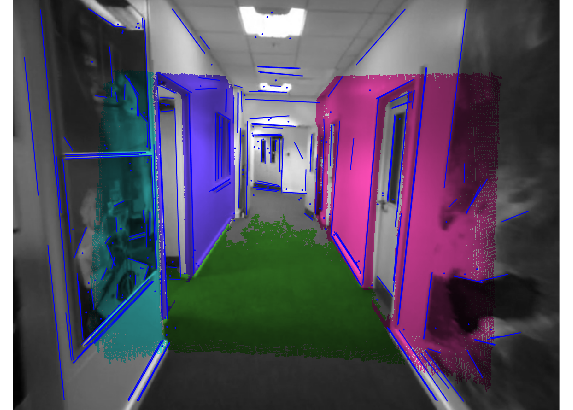} & 
		\includegraphics[scale=0.15]{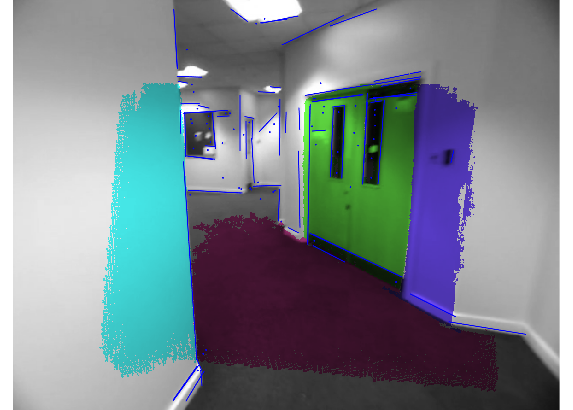} &
		\includegraphics[scale=0.15]{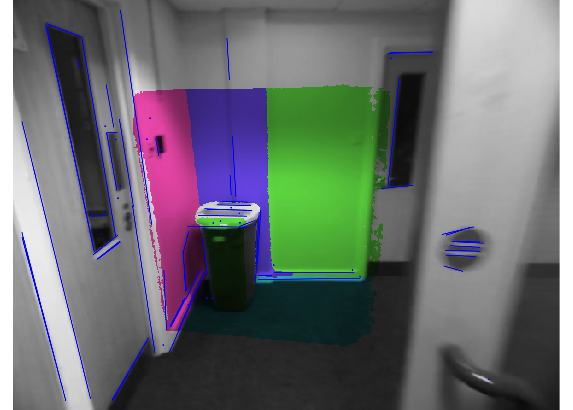} & 
		\includegraphics[scale=0.15]{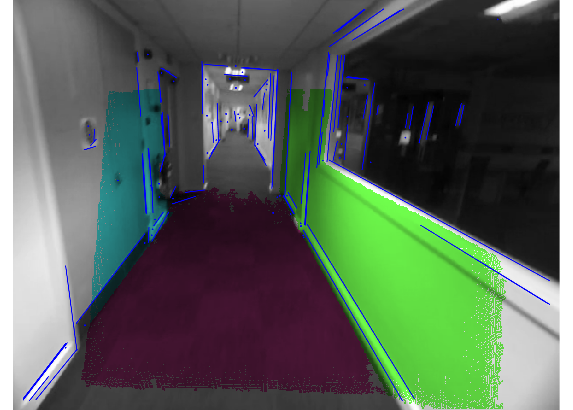} \\[-3pt]
		\includegraphics[scale=0.15]{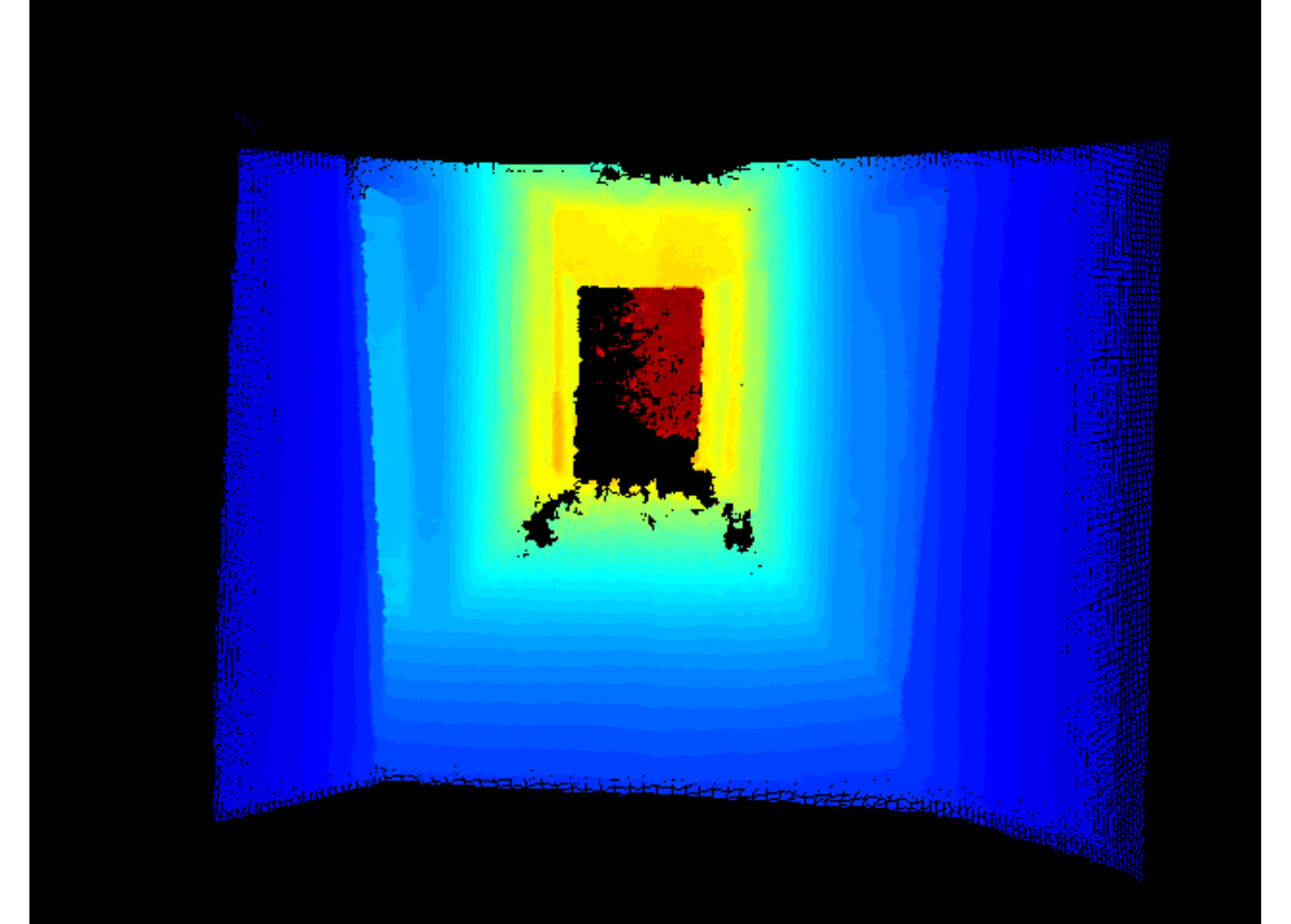} & 
		\includegraphics[scale=0.15]{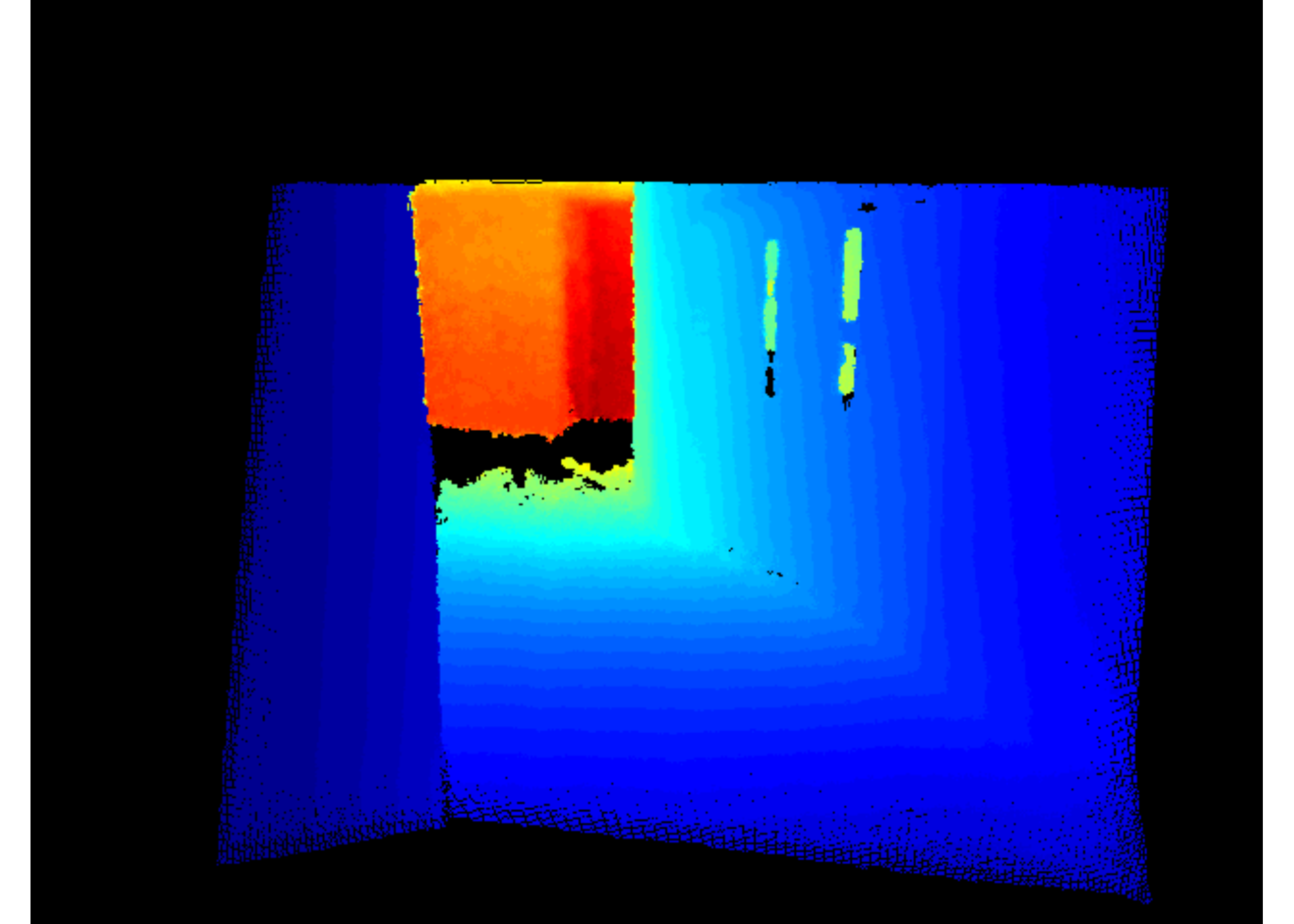} &
		\includegraphics[scale=0.15]{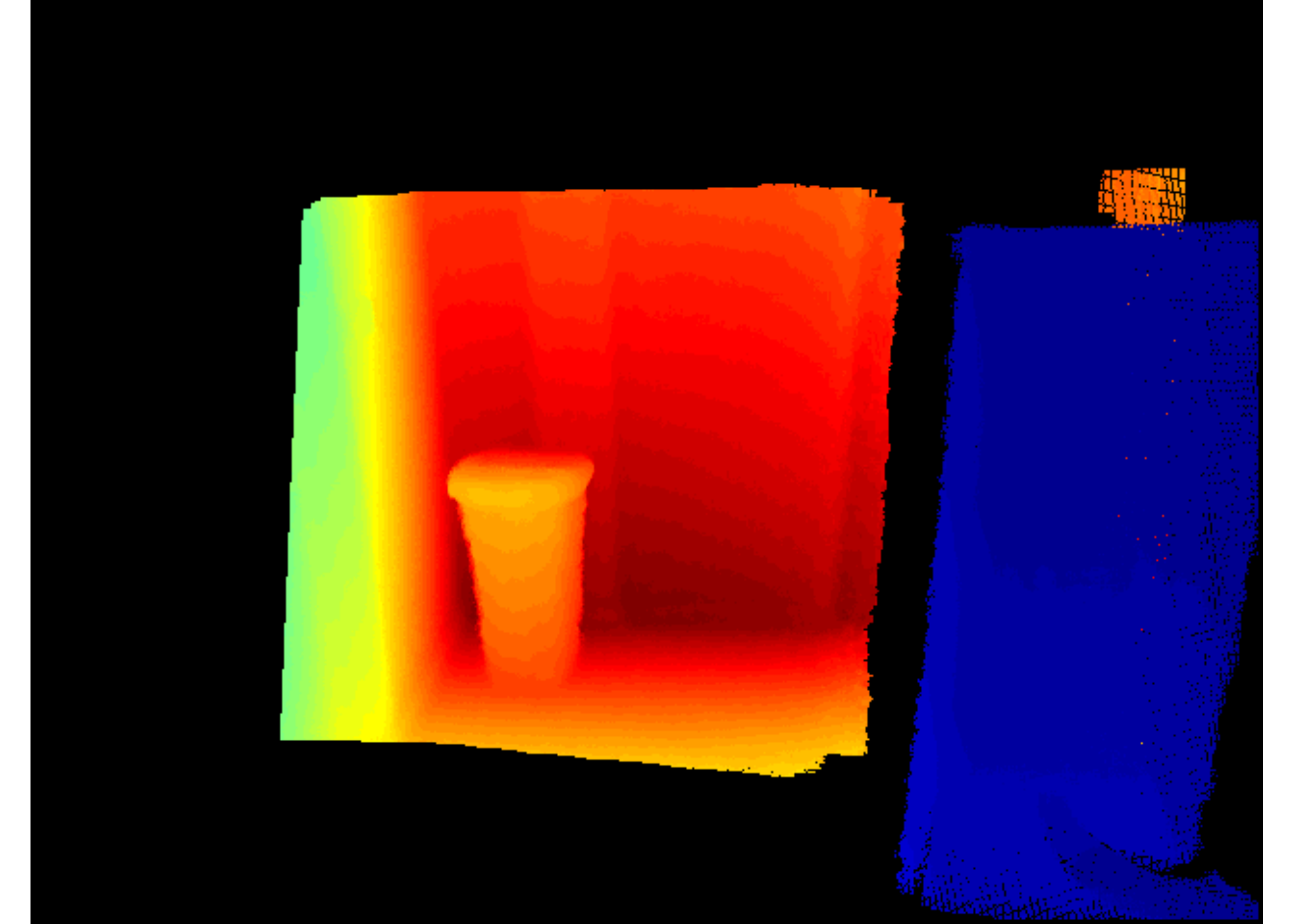} & 
		\includegraphics[scale=0.15]{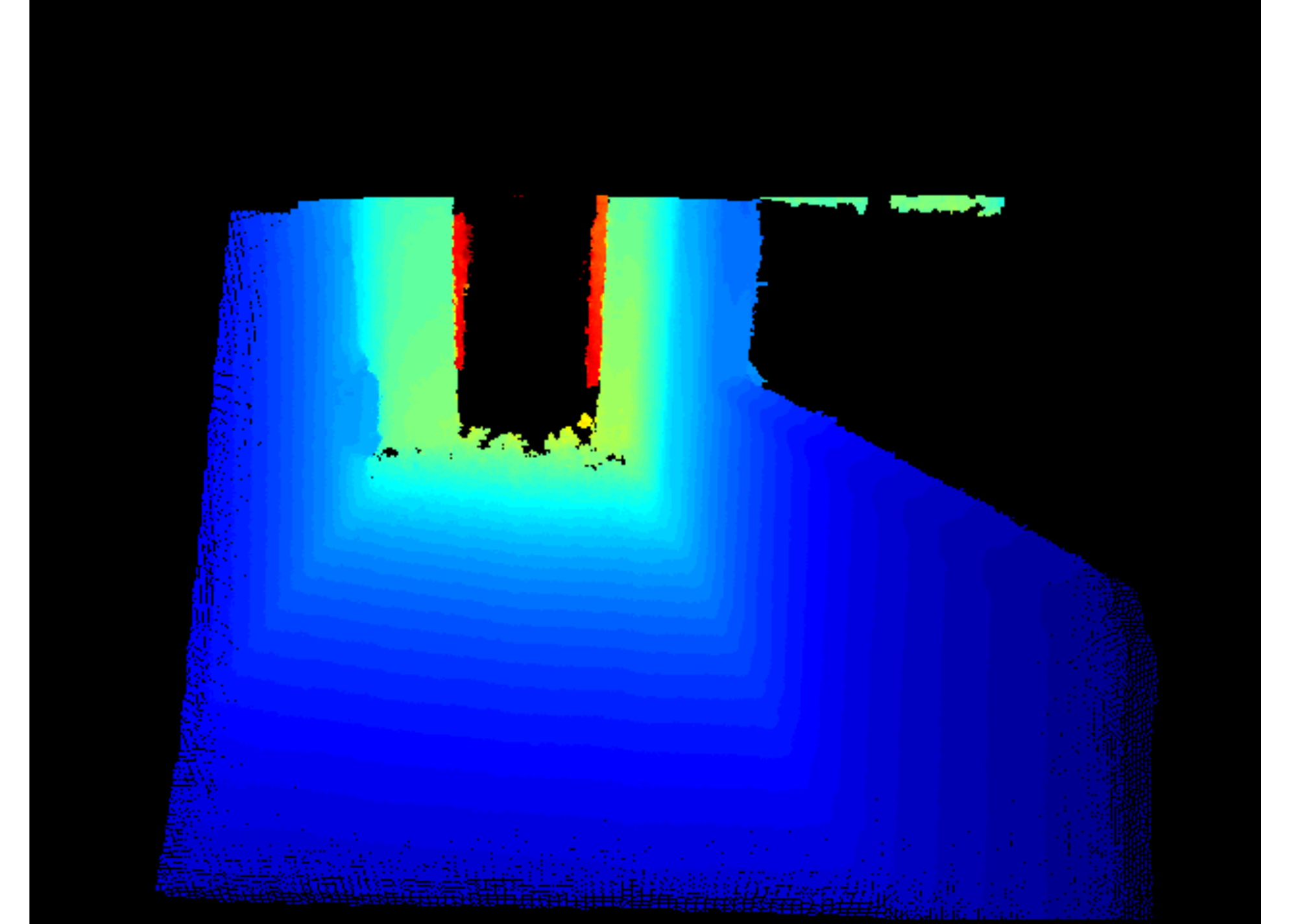} \\[-5pt]
		\multicolumn{4}{c}{ {\scriptsize Corridor}} 
	\end{tabular}
	\caption{Frames from the dataset sequences collected and evaluated in this work. Intensity images with overlaid detected features are shown on the top, while the respective fused and aligned depth maps are shown on the bottom.} 
	\label{fig11}
\end{figure}

\section{Conclusion and Future Work}
\label{sec:conclusions}
Combining points with lines and planes proves to improve the robustness of visual odometry. Our results show no redundancy between the different primitives.
The proposed method achieves state-of-the-art results, in terms of RPE, between frame-to-frame odometry methods. However, in the ICL-NUIM dataset, the ATE yield by our visual odometry is still inferior to some model-to-frame \cite{newcombe2011kinectfusion} and keyframe-to-frame \cite{gutierrez2016dense} based approaches, thus, extending our method to a SLAM version, such as in \cite{ma2016cpa}, is promising research direction in order to reduce the pose drift. \par
Furthermore, the visual odometry performance is improved significantly by using the proposed depth fusion framework. While this framework shows its capability to denoise the raw depth maps, errors may be still introduced and propagated due to flying pixels, occlusions and pose errors, therefore modelling the depth uncertainty is necessary to capture these errors. The full system was additionally evaluated on RGB-D video sequences, captured with a wide-angle RGB camera, where the depth fusion framework plays also the role of recovering old depth measurements that are no longer inside the current FOV of the depth camera. These past measurements can only be maintained for a small number of frames, specified by the depth fusion framework. A more flexible strategy could extend this duration by switching old pixels from depth fusion to a simple hole-filling mode, where points would only be maintained and used if they had an unique pixel projection.

\section*{Acknowledgement}
This work was supported by Sellafield Ltd.

%\section*{References}

\bibliography{mybibfile}

\end{document}